\documentclass{article}

\PassOptionsToPackage{numbers, compress}{natbib}
\usepackage[final]{neurips_2024}

\usepackage[utf8]{inputenc} 
\usepackage[T1]{fontenc}    
\usepackage{graphicx}
\usepackage{booktabs}
\usepackage{amssymb}
\usepackage{pifont}
\usepackage[colorlinks,linkcolor=red,citecolor=blue,urlcolor=magenta,bookmarks=true]{hyperref}        
\usepackage{amsmath}
\usepackage{cleveref}
\usepackage{url}            
\usepackage{booktabs}       
\usepackage{amsfonts}       
\usepackage{nicefrac}       
\usepackage{microtype}      
\usepackage{xcolor}         
\usepackage{indentfirst}
\usepackage[inline, shortlabels]{enumitem}
\usepackage{lipsum}
\usepackage{bm}
\usepackage{algpseudocode}
\usepackage{algorithm}
\usepackage{multirow}
\newcommand{\cmark}{\ding{51}}%
\newcommand{\xmark}{\ding{55}}%
\usepackage{colortbl}
\usepackage{graphicx}
\usepackage{overpic}

\definecolor{gray}{gray}{0.95}
\usepackage{listings}
\usepackage{color}

\definecolor{dkgreen}{rgb}{0,0.6,0}
\definecolor{mauve}{rgb}{0.58,0,0.82}

\title{GaussianCube: A Structured and Explicit \\ Radiance Representation for 3D Generative Modeling}

\author{%
	Bowen Zhang$^{1*}$ \quad\qquad Yiji Cheng$^{2*}$ \qquad Jiaolong Yang$^{3\dagger}$  \qquad Chunyu Wang$^{3\dagger}$ \vspace{8pt}\\
	\textbf{Feng Zhao$^{1\ddagger}$} \quad\qquad \textbf{Yansong Tang$^{2}$} \quad\qquad \textbf{Dong Chen$^{3\ddagger}$}  \quad\qquad \textbf{Baining Guo$^{3}$} \vspace{8pt}\\
	\small{\textsuperscript{1}University of Science and Technology of China \quad \textsuperscript{2}Tsinghua University \quad \textsuperscript{3}Microsoft Research Asia } \vspace{3pt}
}

\begin{document}

\maketitle
{
	\renewcommand{\thefootnote}%
	{\fnsymbol{footnote}}
	\footnotetext[1]{Interns at Microsoft Research Asia. $^{\dagger}$Equal advising. $^{\ddagger}$Corresponding authors.}
}

\begin{abstract}
   We introduce a radiance representation that is both structured and fully explicit and thus greatly facilitates 3D generative modeling. Existing radiance representations either require an implicit feature decoder, which significantly degrades the modeling power of the representation, or are spatially unstructured, making them difficult to integrate with mainstream 3D diffusion methods. We derive GaussianCube by first using a novel densification-constrained Gaussian fitting algorithm, which yields high-accuracy fitting using a fixed number of free Gaussians, and then rearranging these Gaussians into a predefined voxel grid via Optimal Transport. Since GaussianCube is a structured grid representation, it allows us to use standard 3D U-Net as our backbone in diffusion modeling without elaborate designs. More importantly, the high-accuracy fitting of the Gaussians allows us to achieve a high-quality representation with orders of magnitude fewer parameters than previous structured representations for comparable quality, ranging from one to two orders of magnitude. The compactness of GaussianCube greatly eases the difficulty of 3D generative modeling. Extensive experiments conducted on unconditional and class-conditioned object generation, digital avatar creation, and text-to-3D synthesis all show that our model achieves state-of-the-art generation results both qualitatively and quantitatively, underscoring the potential of GaussianCube as a highly accurate and versatile radiance representation for 3D generative modeling. Project page: \href{https://gaussiancube.github.io/}{https://gaussiancube.github.io/}.
\end{abstract}

\section{Introduction}

The field of 3D generation~\cite{wang2023rodin,muller2023diffrf,cao2023large,tang2023volumediffusion,shue20233d,chan2022efficient,gao2022get3d,chen2023single,wang2024phidias,DiffGS} has witnessed remarkable growth, driven by advancements in generative modeling~\cite{ho2020denoising,goodfellow2020generative,nichol2021improved,dhariwal2021diffusion,zhang2022styleswin,karras2019style}. 
Most of the prior works in this domain leverage variants of Neural Radiance Field (NeRF)~\cite{mildenhall2021nerf,chan2022efficient,tang2023volumediffusion,muller2022instant} as their underlying 3D representations, which typically consist of an explicit structured proxy representation and an implicit feature decoder. 
However, such hybrid NeRF variants exhibit degraded representation power, particularly when used for generative modeling where a single implicit feature decoder is shared across all objects. Additionally, the high computational complexity of volumetric rendering leads to both slow rendering speed and extensive memory costs.

Recently, the emergence of 3D Gaussian Splatting (GS)~\cite{kerbl20233d} has enabled improved reconstruction quality and real-time rendering capabilities~\cite{xu2023gaussian,luiten2023dynamic,wu20234d,lu2024manigaussian}. The fully explicit nature of 3DGS eliminates the need for a shared implicit decoder, providing another key advantage over NeRFs. Although 3DGS has been widely studied in scene reconstruction tasks, its spatially unstructured nature presents a significant challenge when applied to mainstream generative modeling frameworks.

To overcome these barriers, we introduce GaussianCube -- an innovative radiance representation that is both structured and fully explicit, with strong fitting capabilities (see~\Cref{tab:comparison_prior_works} for comparisons with prior works). The proposed approach first ensures high-accuracy fitting with a predefined number of free Gaussians, and subsequently organizes these Gaussians into a structured voxel grid. Such an explicit grid-based structure permits the seamless application of standard 3D convolutional architectures, such as U-Net, thereby eliminating the need for complex, specialized network designs~\cite{zhou20213d,wang2023rodin} that are often necessary with unstructured or implicitly decoded representations.

Structuring 3D Gaussians without sacrificing fitting quality is not a trivial task. A naive starting point would be obtaining a fixed number of Gaussians by omitting the densification and pruning steps in GS. However, such simplification fails to lead the Gaussians close to the object surfaces and results in significant quality degradation. In contrast, we propose a \emph{densification-constrained fitting} strategy, which retains the original pruning process yet constrains the number of Gaussians that perform densification, ensuring the total does not exceed a predefined maximum $N^3$. For the subsequent structuralization, we allocate the Gaussians across an $N\times N\times N$ voxel grid using \emph{Optimal Transport (OT)}. Consequently, our fitted Gaussians are systematically arranged within the voxel grid, with each voxel containing the features of a Gaussian. The proposed OT-based structuralization achieves maximal spatial correspondence, characterized by minimal total transport distances, while preserving the expressive power of 3DGS.

\begin{table*}[t]  
	\centering 
	\scriptsize
	\begin{tabular}{ccccc}  
		\hline  
		\textbf{Representation} & \textbf{Spatially-structured} & \textbf{Fully-explicit}  & \textbf{Real-time Rendering} & \textbf{Rel. Parameters$\downarrow$}\\
		\hline
        Instant-NGP~\cite{mildenhall2021nerf} & \xmark & \xmark & \xmark & $26.63 \times$ \\
		Neural Voxels~\cite{tang2023volumediffusion} & \cmark & \xmark & \xmark & $145.9\times$ \\
		Triplane~\cite{chan2022efficient}  & \cmark & \xmark & \xmark & $13.7 \times$ \\  
		Gaussian Splatting~\cite{kerbl20233d} & \xmark & \cmark & \cmark & $4.0 \times$ \\ 
		\cellcolor{gray}\textbf{Our GaussianCube} & \cellcolor{gray}\cmark &
		\cellcolor{gray}\cmark & \cellcolor{gray}\cmark & \cellcolor{gray}$\mathbf{1.0\times}$ \\   
		\hline  
	\end{tabular}  
	\caption{Comparison with previous 3D representations with respect to spatial structure, explicitness, real-time rendering capability, and relative parameter count (Rel. Parameters) for representations of comparable quality.}  
	\vspace{-5mm}
	\label{tab:comparison_prior_works}  
\end{table*}  

The structured nature of GaussianCube enables us to perform efficient 3D diffusion~\cite{ho2020denoising} modeling for the following three reasons: 1) It allows the use of standard 3D U-Net as our backbone for diffusion modeling without elaborate designs. 2) The spatial coherence of GaussianCube permits the use of standard 3D convolutions to capture the correlations among neighboring Gaussians, facilitating efficient feature extraction. 3) GaussianCube enables high-quality fitting with orders of magnitude fewer parameters than prior  structured representations of similar quality. Since recent  works~\cite{li2023generative,blattmann2023align} have demonstrated diffusion models' struggle in handling high-dimensional distributions, the compactness of GaussianCube significantly reduces the modeling difficulty of the generative framework.  

\begin{figure*}[t]
	\small
	\centering
        \renewcommand{\arraystretch}{0.5}
        \vspace{-5mm}
	\begin{tabular}{c}
		\includegraphics[width=0.997\linewidth]{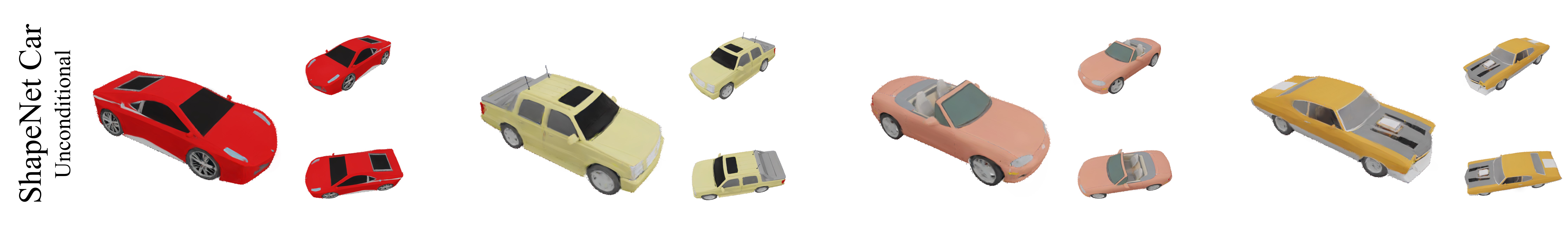}\\
		\includegraphics[width=0.997\linewidth]{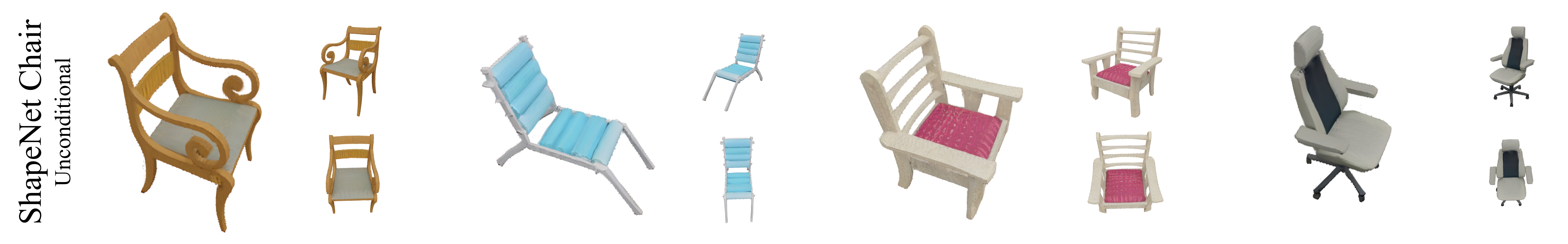}\\
		\includegraphics[width=0.997\linewidth]{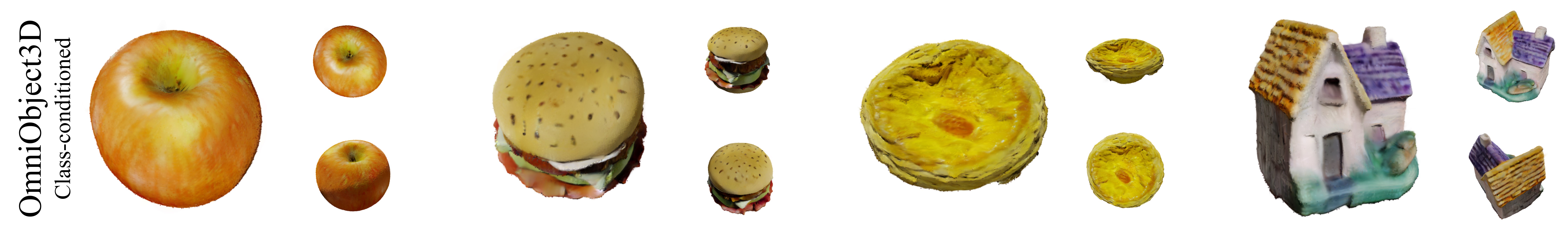}\\
        \includegraphics[width=0.997\linewidth]{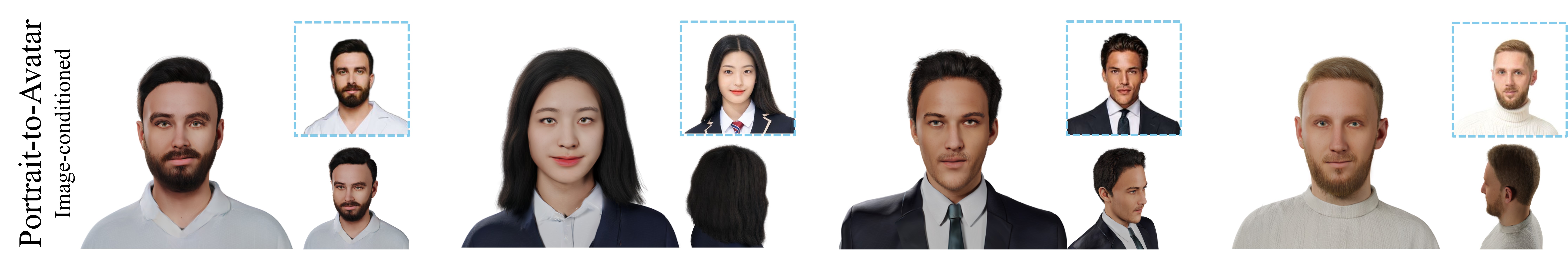}\\
        \includegraphics[width=0.997\linewidth]{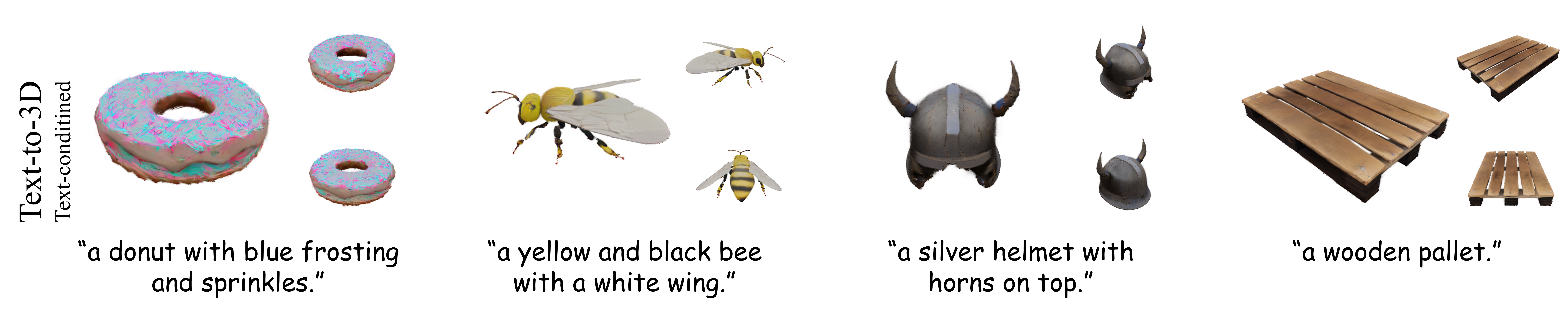}
	\end{tabular}
        \vspace{-3mm}
	\caption{Our diffusion model is able to create diverse objects with complex geometry and rich texture details (top three rows). Our method also supports creating high-fidelity digital avatars (the forth row) conditioned on single portrait images (visualized in dashed boxes) and high-quality 3D assets given text prompts (the fifth row).}
	\vspace{-5mm}
	\label{fig:teaser}
\end{figure*}

We conduct comprehensive experiments to verify the efficacy of our approach. The model's capability for unconditional and class-conditioned generation is evaluated on the ShapeNet~\cite{chang2015shapenet} and OmniObject3D~\cite{wu2023omniobject3d} datasets. Both the quantitative and qualitative comparisons indicate that our model surpasses all previous methods. We also perform digital avatar generation on a synthetic avatar dataset~\cite{wood2021fake}. Our model is capable of producing high-fidelity 3D avatars conditioned on single portrait images, excelling beyond prior art in both identity preservation and detail creation.
Additionally, we assess our model's capacity for the challenging text-to-3D creation task on Objaverse~\cite{deitke2023objaverse}. Our model demonstrates competitive performance both quantitatively and qualitatively, producing results consistent with the given text prompts in just $2.3$ seconds. All experiments show the strong capabilities of our GaussianCube and suggest its potential as a powerful and versatile 3D representation for a variety of applications. Some generated samples of our method are presented in~\Cref{fig:teaser}.

\vspace{-3mm}
\section{Related Work}
\noindent \textbf{Radiance field representation.} Radiance fields model ray interactions with scene surfaces and can be in either implicit or explicit forms. Early works of neural radiance fields (NeRFs)~\cite{mildenhall2021nerf,zhang2020nerf++,park2021nerfies,barron2022mip,pumarola2021d} are often in an implicit form, which represents scenes without defining geometry. These works optimize a continuous scene representation using volumetric ray-marching that leads to extremely high computational costs. Recent works introduce the use of explicit proxy representation~\cite{chan2022efficient,hu2023tri,fridovich2022plenoxels,sun2022direct,muller2022instant,xu2022point} followed by an implicit feature decoder to enable faster rendering. Recently, the 3D Gaussian Splatting methods~\cite{kerbl20233d,xu2023gaussian,wu20234d,cotton2024dynamic,li2024gaussianbody,chen2024survey,zhan2024interactive} utilize 3D Gaussians as their underlying representation and offer impressive reconstruction quality. The fully explicit representation also provides real-time rendering speed. However, the 3D Gaussians are unstructured representation, and require per-scene optimization to achieve photo-realistic quality. In contrast, our work proposes a structured representation termed GaussianCube for 3D generative tasks.

\noindent \textbf{3D generation.}
Previous works of SDS-based optimization~\cite{poole2022dreamfusion,tang2023make,xu2022dream3d,sun2023dreamcraft3d,cheng2023efficient,tang2023dreamgaussian,yi2023gaussiandreamer,tang2024make} distill 2D diffusion priors~\cite{rombach2022high} to a 3D representation with the score functions, but these works are notably time-intensive, often taking several minutes to hours. While 3D-aware GANs~\cite{chan2022efficient,gao2022get3d,chan2021pi,gu2021stylenerf,niemeyer2021giraffe,deng2022gram,xiang2022gram} facilitate view-dependent image generation from single-image collections, they struggle to capture the complexity of diverse objects with intricate geometric variations~\cite{xia2023survey}. Although recent works~\cite{wang2023rodin,muller2023diffrf,gupta20233dgen,tang2023volumediffusion,shue20233d,zhang2024rodinhd} have utilized diffusion models with structured proxy representations for 3D generation, the use of a shared implicit feature decoder across different assets restricts expressiveness and the computational demands of NeRF hinder efficient training. In contrast, we introduce a structured and fully explicit radiance representation for 3D generative modeling, building upon 3DGS~\cite{kerbl20233d}. A concurrent work of \cite{he2024gvgen} includes elaborate designs to form the Gaussians into volumetric representation during fitting, yet does not thoroughly address global correspondence. In contrast, our approach only restricts the total count of Gaussians while allowing freedom in their spatial distribution during the fitting. We then organize these Gaussians into a voxel grid using Optimal Transport, which yields a spatially coherent arrangement with minimal global offset cost, effectively easing the difficulty of generative modeling.

\section{Method}

\begin{figure*}[t]
	\small
	\centering
	\includegraphics[width=0.9\linewidth]{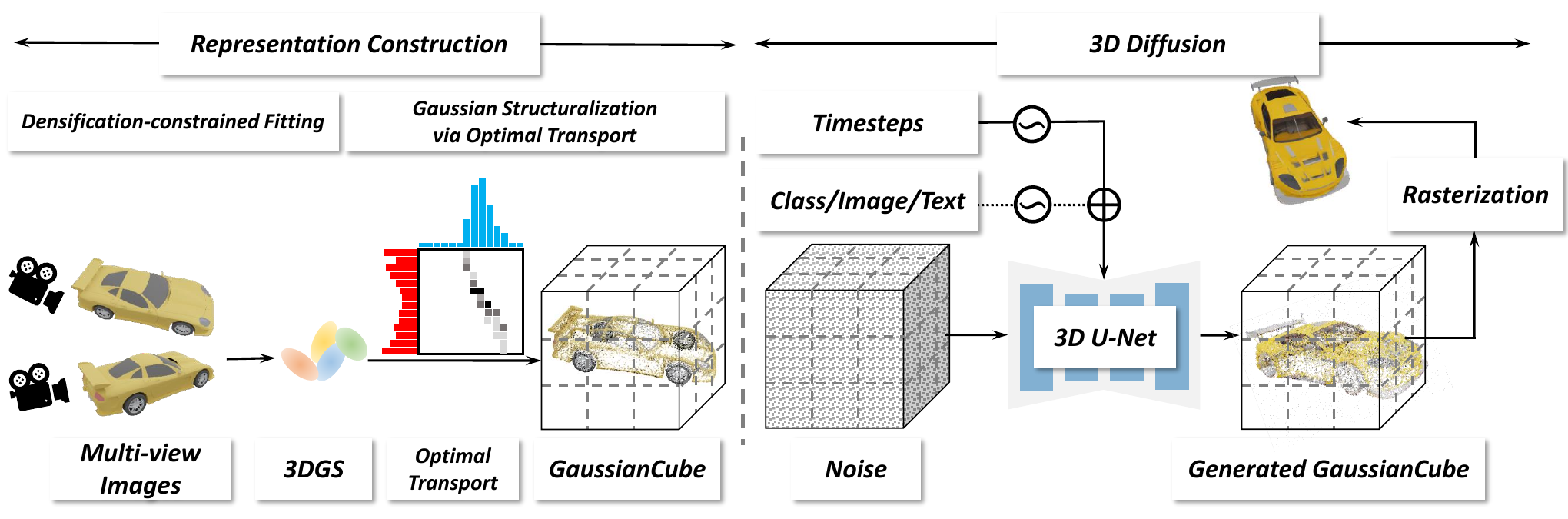}
	\caption{\textbf{Overall framework.} Our framework comprises two main stages of representation construction and 3D diffusion. In the representation construction stage, given multi-view renderings of a 3D asset, we perform \emph{densification-constrained fitting} to obtain 3D Gaussians with constant numbers. Subsequently, the Gaussians are structured into GaussianCube via \emph{Optimal Transport}. In the 3D diffusion stage, our \emph{3D diffusion model} is trained to generate GaussianCube from Gaussian noise.}
	\label{fig:overall_framework}
\end{figure*}

Following prior works, our framework comprises two primary stages as shown in~\Cref{fig:overall_framework}: representation construction and diffusion modeling. In representation construction phase, we first apply a densification-constrained 3DGS fitting algorithm for each object to obtain a constant number of Gaussians. These Gaussians are then organized into the proposed spatially structured GaussianCube via Optimal Transport between the positions of Gaussians and centers of a predefined voxel grid. For diffusion modeling, we train a 3D diffusion model to learn the distribution of GaussianCubes. We will detail our designs for each stage subsequently.

\subsection{Representation Construction}
\label{sec:representation_construction}
We build our GaussianCube upon 3DGS, an explicit representation that offers impressive fitting quality and real-time rendering speed. However, it fails to yield Gaussians of fixed length since the adaptive density control during GS fitting can lead to a varying number of Gaussians for different objects. Furthermore, the lack of a predetermined spatial ordering for Gaussians leads to a disorganized spatial structure. These aspects pose significant challenges to 3D generative modeling. To overcome these obstacles, we first introduce our densification-constrained fitting strategy to obtain a fixed number of free Gaussians. Then, we systematically arrange the resulting Gaussians within a predefined voxel grid via Optimal Transport, thereby achieving a structured and explicit radiance representation.

Formally, a 3D asset is represented by a collection of 3D Gaussians as introduced in Gaussian Splatting~\cite{kerbl20233d}. The geometry of the $i$-th 3D Gaussian $\bm{g}_i$ is given by
\begin{equation}
	\bm{g}_i(\bm{x})=\exp \left(-\frac{1}{2}\left(\bm{x}-\bm{\mu}_i\right)^{\top} \bm{\Sigma}_i^{-1}\left(\bm{x}-\bm{\mu}_i\right)\right),
\end{equation}
where $\bm{\mu}_i \in \mathbb{R}^3$ is the center of the Gaussian and $\bm{\Sigma}_i \in \mathbb{R}^{3\times3}$ is the covariance matrix defining the shape and size, which can be decomposed into a quaternion $\bm{q}_i \in \mathbb{R}^4$ and a vector $\bm{s}_i \in \mathbb{R}^3$ for rotation and scaling, respectively. Moreover, each Gaussian $\bm{g}_i$ have an opacity value $\alpha_i \in \mathbb{R}$ and a color feature $\bm{c}_i \in \mathbb{R}^3$ for rendering. Combining them together, the $C$-channel feature vector $\bm{\theta}_i = \{\bm{\mu}_i, \bm{s}_i, \bm{q}_i, \alpha_i, \bm{c}_i \} \in \mathbb{R}^C$ fully characterizes the Gaussian $\bm{g}_i$. 

\begin{figure*}[t]
	\small
	\centering
	\begin{overpic}[width=0.8\linewidth]{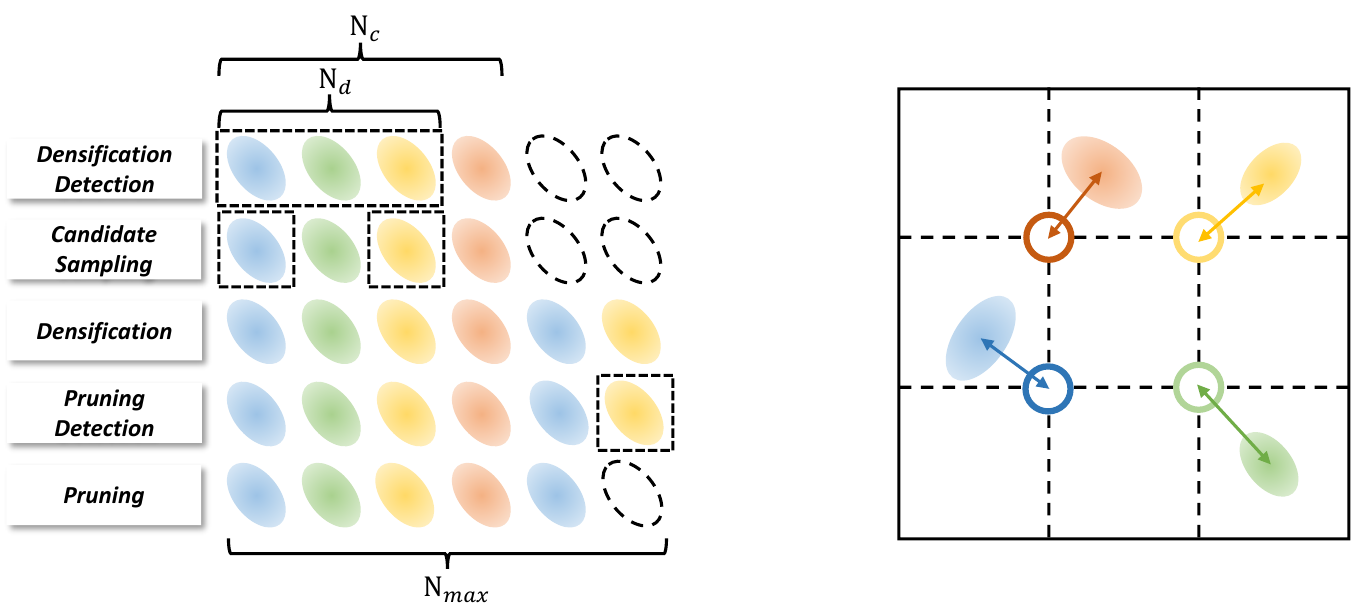}
		\put(32,-2){(a)}
		\put(81,-2){(b)}
	\end{overpic}
	\vspace{1mm}
	\caption{\textbf{Illustration of representation construction.} First, we perform densification-constrained fitting to yield a fixed number of Gaussians, as shown in (a). We then employ Optimal Transport to organize the resultant Gaussians into a voxel grid. A 2D illustration of this process is presented in (b).}
	\label{fig:representation_building}
	\vspace{-3mm}
\end{figure*}

\noindent\textbf{Densification-constrained fitting}. Our approach begins with the aim of maintaining a constant number of Gaussians $\bm{g}\in \mathbb{R}^{N_{\text{max}} \times C}$ across different objects during the fitting. A simplistic approach might involve omitting the densification and pruning in the original GS. However, we argue that such simplifications significantly harm the fitting quality, with empirical evidence shown in~\Cref{tab:ablation_fitting_generation}. Instead, we propose to retain the pruning process while imposing a new constraint on the densification phase as shown in~\Cref{fig:representation_building} (a). The fitting process encompasses several distinct stages: 1) Densification Detection: Assuming the current iteration includes $N_{\text{c}}$ Gaussians, we identify densification candidates by selecting those with view-space position gradient magnitudes exceeding a predefined threshold $\tau$. We denote the number of candidates as $N_d$. 2) Candidate sampling: To prevent exceeding the predefined maximum of $N_{\text{max}}$ Gaussians, we select $\min{(N_{\text{max}} - N_{\text{c}}, N_d)}$ Gaussians with the largest view-space positional gradients from the candidates for densification. 3) Densification: We modify the densification approach by alternating between cloning and splitting actions into separate steps. 4) Pruning Detection and Pruning: We identify and remove the Gaussians with $\alpha$ less than a small threshold $\epsilon$. After completing the fitting process, we pad Gaussians with $\alpha=0$ to reach the target count of $N_{\text{max}}$ without affecting the rendering results. Benefiting from our proposed strategy, we attain a high-quality representation with orders of magnitude fewer parameters compared to existing works of similar quality, which significantly reduces the modeling difficulty for the diffusion models.

\noindent\textbf{Gaussian structuralization via Optimal Transport}. To further organize the obtained Gaussians into a spatially structured representation for 3D generative modeling, we propose to map the Gaussians to a predefined structured voxel grid $\bm{v} \in \mathbb{R}^{N_v \times N_v \times N_v \times C}$ where $N_v=\sqrt[3]{N_{\text{max}}}$. Intuitively, we aim to ``move'' each Gaussian into a voxel while preserving their geometric relations as much as possible. While naive approaches such as nearest neighbor transport fall short in conserving these relations due to disregard for global arrangement with evidence shown in~\Cref{fig:ablation_mapping_generation}, we formulate this as an Optimal Transport (OT) problem~\cite{villani2009optimal,burkard1999linear} between the Gaussians' spatial positions $\{\bm{\mu}_i, i=1,\ldots,N_{\text{max}}\}$ and the voxel grid centers $\{\bm{x}_j, j=1,\ldots, N_{\text{max}}\}$. Let $\mathbf{D}$ be a distance matrix with  $\mathbf{D}_{ij}$ being the moving distance between $\bm{\mu}_i$ and $\bm{x}_j$, i.e., $\mathbf{D}_{ij}=\| \bm{\mu}_i - \bm{x}_j \|^2$. The transport plan is represented by a binary matrix $\mathbf{T}\in\mathbb{R}^{N_{\text{max}}\times N_{\text{max}}}$, and the optimal transport plan is given by:
\begin{equation}
	\begin{array}{ll}
		\underset{\mathbf{T}}{\operatorname{minimize}} & \sum_{i=1}^{N_{\text{max}} }\sum_{j=1}^{N_{\text{max}}} \mathbf{T}_{i j} \mathbf{D}_{i j} \\
		\text { subject to } & \sum_{j=1}^{N_{\text{max}}} \mathbf{T}_{i j}=1 \quad \forall i \in\{1, \ldots, N_{\text{max}}\} \\
		& \sum_{i=1}^{N_{\text{max}}} \mathbf{T}_{i j}=1 \quad \forall j \in\{1, \ldots, N_{\text{max}}\} \\
		& \mathbf{T}_{i j} \in \{0, 1\} \qquad \forall(i, j) \in\{1, \ldots, N_{\text{max}}\} \times\{1, \ldots, N_{\text{max}}\}.
	\end{array}
\end{equation}
The solution is a bijective transport plan $\mathbf{T}^*$ that minimizes the total transport distances.
We employ the Jonker-Volgenant algorithm~\cite{jonker1988shortest} to solve the OT problem. We provide a 2D illustration in~\Cref{fig:representation_building} (b). We organize the Gaussians according to the solutions, with the $j$-th voxel encapsulating the feature vector of the corresponding Gaussian $\bm{\theta}_k = \{\bm{\mu}_k - \bm{x}_j, \bm{s}_k, \bm{q}_k, \alpha_k, \bm{c}_k \} \in \mathbb{R}^C$, where $k$ is determined by the optimal transport plan (\textit{i}.\textit{e}., $\mathbf{T}^*_{k j}=1$). Note that we replace the original Gaussian positions with offsets of the current voxel center to reduce the solution space for diffusion models. As a result, our fitted Gaussians are systematically arranged within a voxel grid $\bm{v}$ and preserve the spatial correspondence of neighboring Gaussians, which further facilitates generative modeling.

\subsection{3D Diffusion on GaussianCube}

We now introduce our 3D diffusion model incorporated with the proposed expressive, efficient and spatially structured representation. After organizing the fitted Gaussians $\bm{g}$ into GaussianCube $\bm{y}$ for each object, we aim to model the distribution of GaussianCube, \textit{i}.\textit{e}., $p(\bm{y})$. 

Formally, the generation procedure can be formulated into the inversion of a discrete-time Markov forward process. During the forward phase, we gradually add noise to $\bm{y}_0 \sim p(\bm{y})$ and obtain a sequence of increasingly noisy samples $\{\bm{y}_t | t \in [0, T]\}$ according to $\bm{y}_t:=\alpha_t \bm{y}_0+\sigma_t \bm{\epsilon}$, where $\bm{\epsilon} \in \mathcal{N}(\mathbf{0}, \boldsymbol{I})$ represents the added Gaussian noise, and $\alpha_t, \sigma_t$ constitute the noise schedule. As a result, $\bm{y}_T$ will finally reach isotropic Gaussian noise after sufficient destruction steps. By reversing the above process, we are able to perform the generation process by gradually denoise the sample starting from pure Gaussian noise $\bm{y}_T \sim \mathcal{N}(\mathbf{0}, \boldsymbol{I})$ until reaching $\bm{y}_0$. Our diffusion model is trained to denoise $\bm{y}_t$ into $\bm{y}_0$ for each timestep $t$, facilitating both unconditional and conditional generation.

\noindent\textbf{Model architecture.} Thanks to the spatially structured organization of the proposed GaussianCube, standard 3D convolution is sufficient to effectively extract and aggregate the features of neighboring Gaussians without elaborate designs. We leverage the standard U-Net network for diffusion~\cite{nichol2021improved,dhariwal2021diffusion} and simply replace the original 2D operators including convolution, attention, upsampling and downsampling with their 3D counterparts.  

\noindent\textbf{Conditioning mechanism.} Our model supports a variety of condition signals to control the generation process. When performing class-conditioned diffusion modeling, we employ adaptive group normalization (AdaGN)~\cite{dhariwal2021diffusion} to inject the class labels into our model. For image-conditioned digital avatar creation, we leverage a pretrained vision transformer~\cite{caron2021emerging} to encode the conditional image into a sequence of feature tokens. We subsequently adopt cross-attention to make the model learn the correspondence between 3D activations and 2D image feature tokens following~\cite{cao2023large}. We also leverage cross-attention as our condition mechanism when creating 3D objects from text, similar to previous text-to-image diffusion models~\cite{rombach2022high}. 

\noindent\textbf{Training objective.} In our 3D diffusion training, we parameterize our model $\hat{\bm{y}}_\theta$ to predict the noise-free input $\bm{y}_0$ using:
\begin{equation}
	\mathcal{L}_{\text {simple }}=\mathbb{E}_{t, \bm{y}_0, \bm{\epsilon}}\left[\left\|\hat{\bm{y}}_\theta\left(\alpha_t \bm{y}_0+\sigma_t \bm{\epsilon}, t,  \bm{c}_{\text{cls}}\right)-\bm{y}_0\right\|_2^2\right],
\end{equation}
where the condition signal $\bm{c}_{\text{cls}}$ is only needed when training conditional diffusion models. We additionally impose image-level supervision to improve the rendering quality of generated GaussianCube, which has been demonstrated to effectively enhance the perceptual details in previous works~\cite{wang2023rodin,muller2023diffrf}. Specifically, we penalize the discrepancy between the rasterized images $I^{t}_{\text{pred}}$ of the model prediction at timestep $t$ and the ground-truth images $I_{\text{gt}}$ using:
\begin{equation}
	\begin{aligned}
		\mathcal{L}_{\text {image }} = \mathbb{E}_{I^{t}_{\text {pred }}}\left(\sum_l\left\|\Psi^l\left(I^{t}_{\text {pred}}\right)-\Psi^l\left(I_{\text {gt}}\right)\right\|_2^2\right) +\mathbb{E}_{I^{t}_{\text {pred}}}\left(\left\|I^{t}_{\text {pred}}-I_{\text {gt }}\right\|_2\right),
	\end{aligned}
\end{equation}
where $\Psi^l$ is the multi-resolution feature extracted using the pre-trained VGG~\cite{simonyan2014very}. Benefiting from the efficiency of both rendering speed and memory costs from GS~\cite{kerbl20233d}, we are able to perform fast training with high-resolution renderings. Our overall training loss can be formulated as:
\begin{equation}
	\mathcal{L} =\mathcal{L}_{\text {simple}}+\lambda\mathcal{L}_{\text{image}},
\end{equation}
where $\lambda$ is a balancing weight.

\begin{table}[t]  
	\centering  
	\scriptsize
    \caption{Comparison with prior 3D representations of spatial structure, fitting quality, relative fitting speed (Rel. Speed) and parameter sizes on ShapeNet Car. $^*$ denotes that the implicit feature decoder is shared across different objects. All methods are evaluated at 30K iterations.}  
	\begin{tabular}{cccccccc}  
		\hline  
		\textbf{Representation} & \textbf{Spatially-structured} & \textbf{PSNR$\uparrow$} & \textbf{LPIPS$\downarrow$} & \textbf{SSIM$\uparrow$} & \textbf{Rel. Speed$\uparrow$} & \textbf{Params (M)$\downarrow$} \\  
		\hline   
		Instant-NGP & \xmark & 33.98  & 0.0386 & 0.9809 & $1\times$ & 12.25 \\
		Gaussian Splatting & \xmark  & \textbf{35.32}  & \textbf{0.0303} & \textbf{0.9874} & \underline{$2.58\times$} & \underline{1.84} \\
		\hline
        Voxels & \cmark & 31.78 & 0.0676 & 0.9664 & $0.15\times$ & 67.12 \\ 
		Voxels$^*$ & \cmark & 30.25 & 0.0926 & 0.9541 & $0.15\times$  & 67.12 \\ 
		Triplane & \cmark & 32.61  & 0.0611 & 0.9709 & $1.05\times$ & 6.30 \\ 
		Triplane$^*$ & \cmark & 31.39 & 0.0759 & 0.9635 & $1.05\times$ & 6.30 \\ 
		\cellcolor{gray}\textbf{Our GaussianCube} & \cellcolor{gray}\cmark & \cellcolor{gray}\underline{34.94} & \cellcolor{gray}\underline{0.0347} & \cellcolor{gray}\underline{0.9863} & \cellcolor{gray}$\mathbf{3.33\times}$ & \cellcolor{gray}\textbf{0.46} \\  
		\hline  
	\end{tabular}  
	\label{tab:exp_fitting}  
\end{table} 

\begin{figure*}[t]
	\centering
	\small
	\setlength\tabcolsep{1pt}
	\begin{overpic}[width=1.0\linewidth]{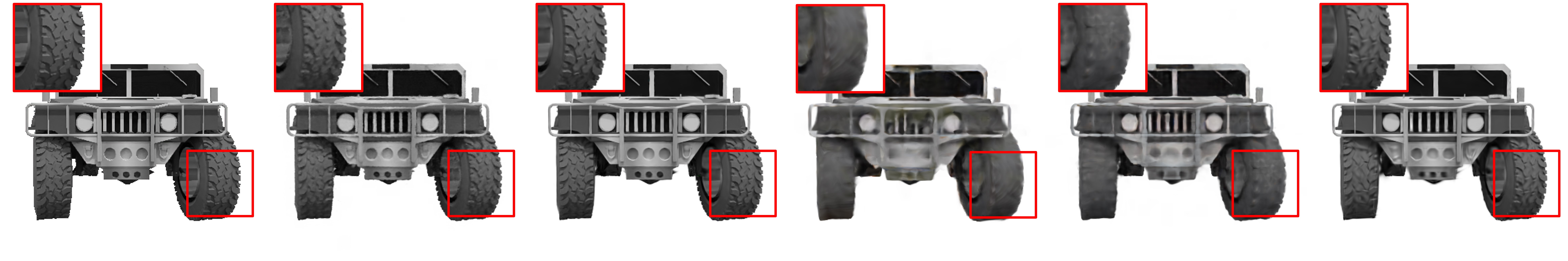}
		\put(3,-2){Ground-truth}
		\put(20,-2){Instant-NGP}
		\put(34,-2){Gaussian Splatting}
		\put(56,-2){Voxel$^*$}
		\put(71,-2){Triplane$^*$}
		\put(83,-2){\textbf{Our GaussianCube}}
	\end{overpic}
	\caption{Qualitative results of object fitting.}
	\vspace{-3mm}
	\label{fig:fitting}
\end{figure*}
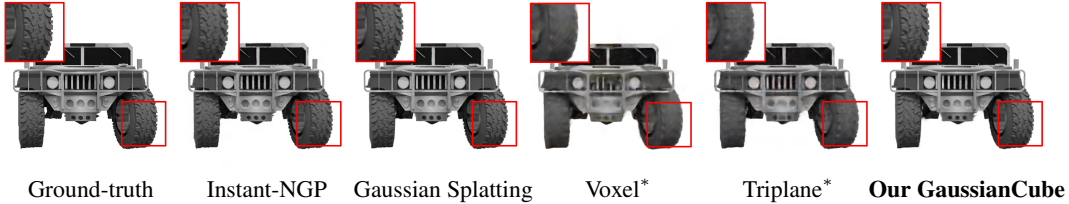

\begin{table*}[t]  
	\centering 
	\small
    \caption{Quantitative results of unconditional generation on ShapeNet Car and Chair~\cite{chang2015shapenet} and class-conditioned generation on OmniObject3D~\cite{wu2023omniobject3d}.} 
    \vspace{1mm}
	\begin{tabular}{ccccccccc}  
		\hline  
		\multirow{2}{*}{\textbf{Method}} & \multicolumn{2}{c}{\textbf{ShapeNet Car}} & \multicolumn{2}{c}{\textbf{ShapeNet Chair}} & \multicolumn{2}{c}{\textbf{OmniObject3D}} \\
		& \textbf{FID-50K$\downarrow$} & \textbf{KID-50K(\textperthousand)$\downarrow$} & \textbf{FID-50K$\downarrow$} & \textbf{KID-50K(\textperthousand)$\downarrow$} & \textbf{FID-50K$\downarrow$} & \textbf{KID-50K(\textperthousand)$\downarrow$}\\  
		\hline  
		EG3D  & 30.48 & 20.42 & 27.98 & 16.01 & - & - \\  
		GET3D  & 17.15 & 9.58 & 19.24 & 10.95 & - & - \\  
		DiffTF & 51.88 & 41.10 & 47.08 & 31.29 & 46.06 & 22.86\\  
		\cellcolor{gray}\textbf{Ours} & \cellcolor{gray}\textbf{13.01} & \cellcolor{gray}\textbf{8.46} & \cellcolor{gray}\textbf{15.99} & \cellcolor{gray}\textbf{9.95} & \cellcolor{gray}\textbf{11.62} & \cellcolor{gray}\textbf{2.78}\\   
		\hline  
	\end{tabular}   
	\label{tab:uncond_clscond_gen}  
\end{table*}  

\begin{figure*}[t]
	\small
	\centering
	\begin{overpic}[width=0.9\linewidth]{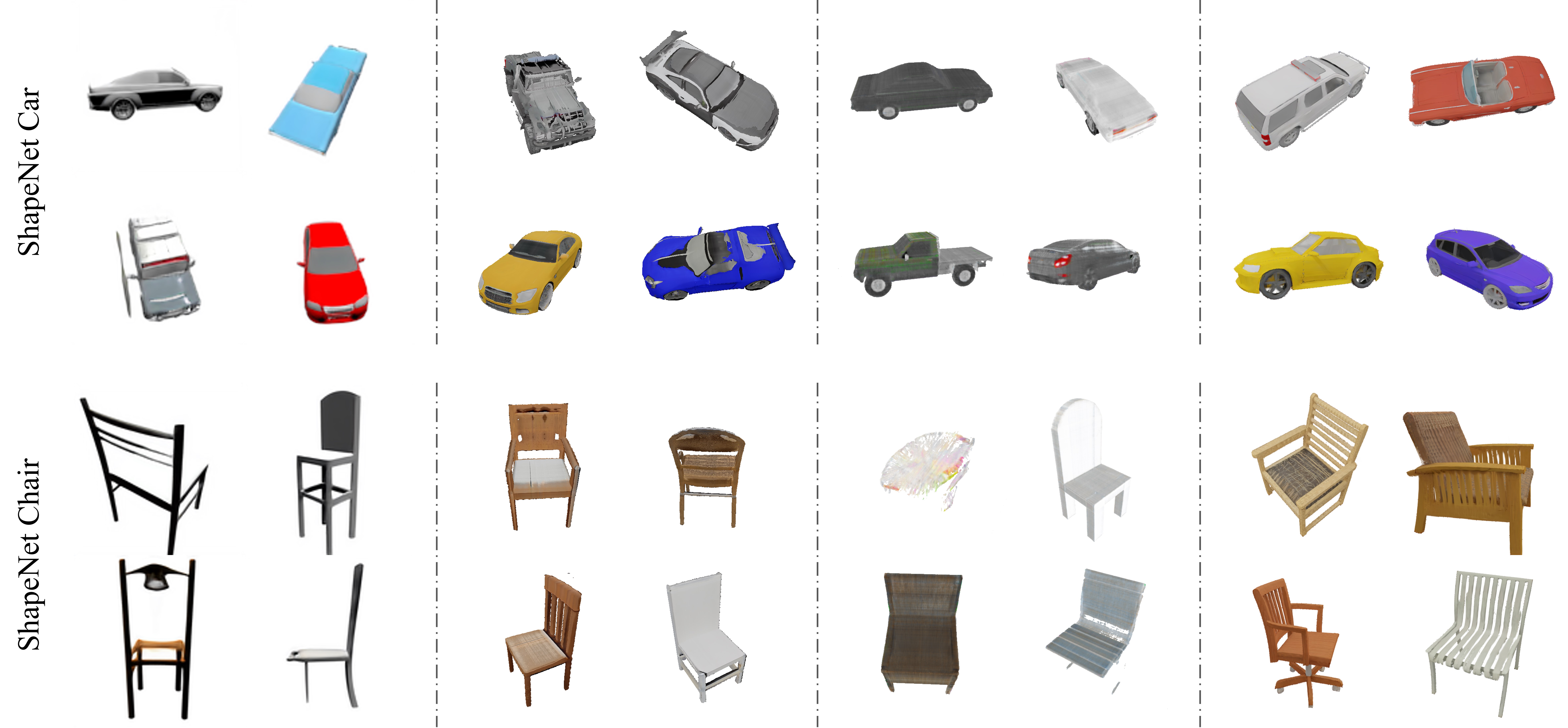}
		\put(12,-3){EG3D~\cite{chan2022efficient}}
		\put(35,-3){GET3D~\cite{gao2022get3d}}
		\put(60,-3){DiffTF~\cite{cao2023large}}
		\put(86,-3){\textbf{Ours}}
	\end{overpic}
	\vspace{3mm}
	\caption{Qualitative comparison of unconditional 3D generation on ShapeNet Car and Chair datasets. Our model is capable of generating results of complex geometry with rich details.}
	\label{fig:uncond_gen}
\end{figure*}

\begin{figure*}[h!]
	\centering
	\small
	\begin{overpic}[width=0.9\linewidth]{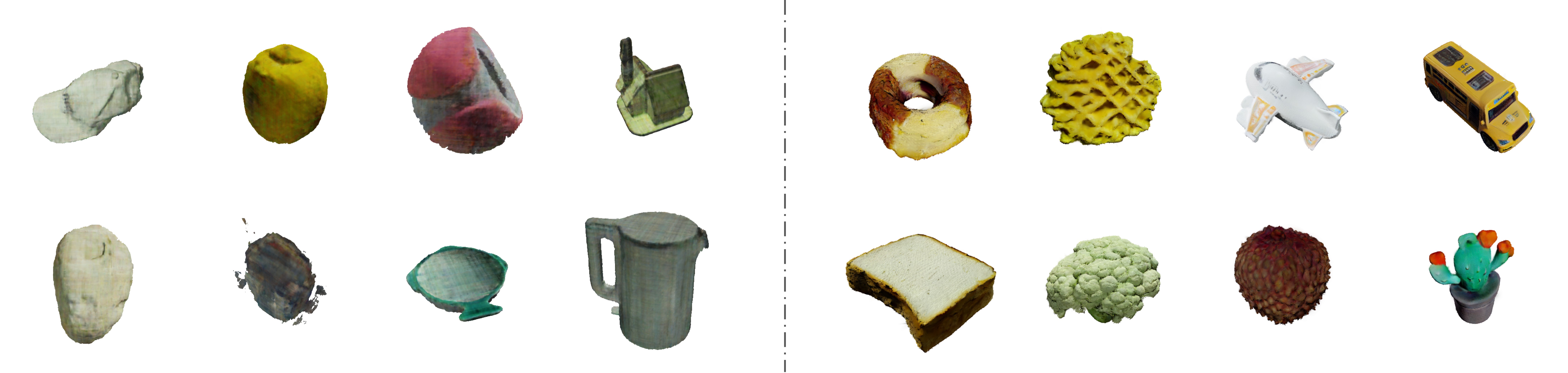}
		\put(18,-3){DiffTF~\cite{cao2023large}}
		\put(75,-3){\textbf{Ours}}
	\end{overpic}
	\vspace{3mm}
	\caption{Qualitative comparison of class-conditioned 3D generation on large-vocabulary OmniObject3D~\cite{wu2023omniobject3d}. Our model is able to handle diverse distribution with semantically accurate results.}
	\label{fig:clscond_gen}
\end{figure*}

\section{Experiments}

\subsection{Dataset and Metrics}
To measure the expressiveness and efficiency of various 3D representations, we fit 100 objects in ShapeNet Car~\cite{chang2015shapenet} using each representation and report the PSNR, LPIPS~\cite{zhang2018unreasonable} and Structural Similarity Index Measure (SSIM) metrics when synthesizing novel views. Furthermore, we conduct experiments of single-category unconditional generation on ShapeNet~\cite{chang2015shapenet} Car and Chair, and class-conditioned generation on real-world scanned dataset OmniObject3D~\cite{wu2023omniobject3d}. We compute the FID~\cite{heusel2017gans} and KID~\cite{binkowski2018demystifying} scores between 50K generated renderings and 50K ground-truth renderings. For image-conditioned digital avatar generation, we utilize the synthetic avatar dataset~\cite{wood2021fake}, which comprises highly-detailed 3D avatars created by synthetic pipeline. We assess the generation quality of 5K rendering from 500 test avatars and additionally include cosine similarity of identity embedding~\cite{deng2019arcface} (CSIM) to measure the ID preservation. The experiments of text-to-3D generation are performed on the large-scale challenging Objaverse dataset~\cite{deitke2023objaverse}. We numerically evaluate the text alignment quality using CLIP score~\cite{radford2021learning} of 300 test prompts.
All images are rendered with $512 \times 512$ resolution.
For more details of data, please refer to~\Cref{supp:implementation_details}.

\subsection{Implementation Details}
For GaussianCube construction, we set $N_{\text{max}}$ to 32,768 and $C$ to 14 across all datasets. We perform the proposed densification-constrained fitting for 30K iterations, which requires approximately 2.67 minutes on a single V100 GPU for each object. After OT-based structuralization, we obtain $32\times32\times32\times14$ GaussianCube for each object. The OT-based structuralization takes around 2 minutes per object on an AMD EPYC 7763v CPU.
For the 3D diffusion model, we adopt the ADM U-Net network~\cite{nichol2021improved,dhariwal2021diffusion}. We perform full attention at the resolution of $8^3$ and $4^3$ within the network. The timesteps of diffusion models are set to $1,000$ and we train the models using the cosine noise schedule~\cite{nichol2021improved} with loss weight $\lambda$ set to $10$. We deploy 16 Tesla V100 GPUs for the ShapeNet Car, ShapeNet Chair, OmniObject3D, and Synthetic Avatar datasets, whereas 32 Tesla V100 GPUs are used for training on the Objaverse dataset. It takes about one week to train our model on ShapeNet Car, ShapeNet Chair, and OmniObject3D, and approximately two weeks for the Synthetic Avatar and Objaverse datasets. For more training details, please refer to~\Cref{supp:implementation_details}.

\begin{table*}[t]  
	\centering 
	\small
    \caption{Quantitative results of digital avatar creation conditioned on single portrait image.} 
    \vspace{1mm}
	\begin{tabular}{ccccccc}  
		\hline  
		\textbf{Method} & \textbf{PSNR$\uparrow$} & \textbf{LPIPS$\downarrow$} & \textbf{SSIM$\uparrow$} & \textbf{CSIM$\uparrow$} & \textbf{FID-5K$\downarrow$} & \textbf{KID-5K(\textperthousand)$\downarrow$}\\  
		\hline  
	Rodin w/o 2D SR & 18.80 & 0.2842 & 0.7439 & 0.6594 & 32.07 & 24.78\\
        Rodin & 18.59 & 0.2821 & 0.7373 & 0.6466 & 20.02 & 9.24\\
        \cellcolor{gray}\textbf{Ours} & \cellcolor{gray}\textbf{21.87} & \cellcolor{gray}\textbf{0.1768} & \cellcolor{gray}\textbf{0.7703} & \cellcolor{gray}\textbf{0.7821} & \cellcolor{gray}\textbf{8.32} & \cellcolor{gray}\textbf{2.67}\\
		\hline  
	\end{tabular}   
	\label{tab:avatar_gen}  
\end{table*}  

\begin{figure*}[t]
    \small
    \centering
    \begin{overpic}[width=0.8\linewidth]{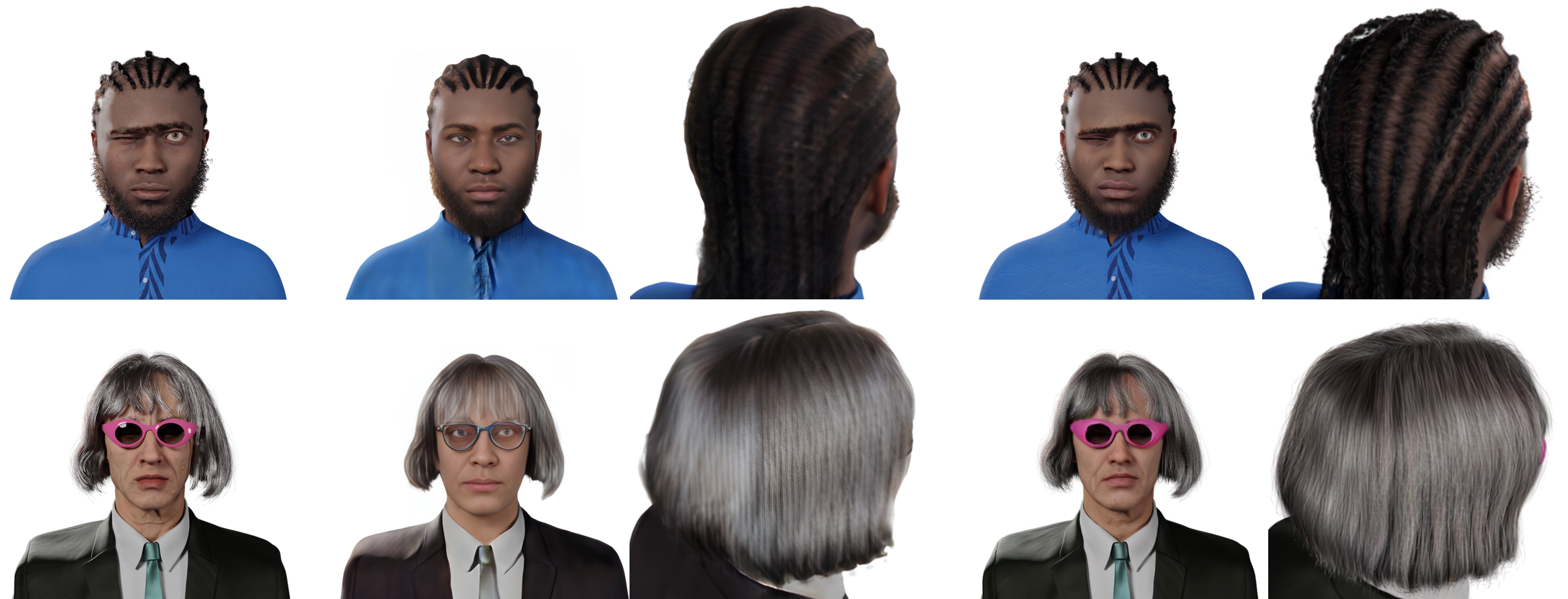}
		\put(5, -3){Reference}
        \put(35, -3){Rodin~\cite{wang2023rodin}}
		\put(78, -3){\textbf{Ours}}
    \end{overpic}
  \vspace{3mm}
  \caption{Qualitative comparison of 3D avatar creation conditioned on single frontal portraits.}
  \label{fig:real_world_avatars}
  \vspace{-3mm}
\end{figure*}

\begin{table*}[t]  
	\centering 
	\small
    \caption{Quantitative results of text-to-3D creation. Inference time is measured on a single A100 GPU. While Shape-E, LGM achieve comparable CLIP scores as ours, they either utilize millions of training data or leverage 2D diffusion prior.} 
    \vspace{1mm}
	\begin{tabular}{cccccc}  
		\hline  
		 & \textbf{DreamGaussian} & \textbf{VolumeDiffusion} & \textbf{Shap-E} & \textbf{LGM} & \textbf{Ours}\\  
		\hline  
	\textbf{CLIP Score$\uparrow$} & 26.38 & 24.41 & 30.52 & 30.06 & \textbf{30.56}\\
        \textbf{Inference Time (s)$\downarrow$} & $\sim120$ & 4.95 & 4.42 & \textbf{1.55} & 2.30 \\
		\hline  
	\end{tabular}  
	\label{tab:textcond_gen} 
\end{table*}  

\begin{figure*}[t]
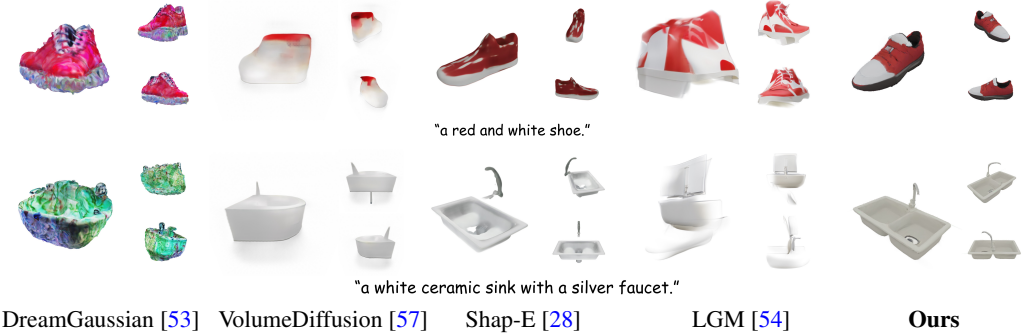

	\centering
	\small
	\begin{overpic}[width=0.98\linewidth]{imgs/results/text_cond_all.jpg}
		\put(0,-2){DreamGaussian~\cite{tang2023dreamgaussian}}
        \put(21, -2){VolumeDiffusion~\cite{tang2023volumediffusion}}
        \put(45, -2){Shap-E~\cite{jun2023shap}}
        \put(67, -2){LGM~\cite{tang2024lgm}}
		\put(88, -2){\textbf{Ours}}
	\end{overpic}
	\vspace{2mm}
	\caption{Qualitative comparison of text-to-3D generation on Objaverse~\cite{deitke2023objaverse}. Our model is able to generate high-quality samples according to the given text prompts.}
	\label{fig:textcond_gen}
\end{figure*}

\subsection{Main Results}
\noindent\textbf{3D fitting.} We first evaluate our representation capability of object fitting against previous NeRF-based representations including Voxels~\cite{tang2023volumediffusion} and Triplane~\cite{chan2022efficient}, which are widely adopted in previous 3D generation works~\cite{chan2022efficient,wang2023rodin,cao2023large,muller2023diffrf,tang2023volumediffusion}. We set the representation size of Voxels and Triplane to $128\times128\times128\times32$ and $256\times256\times32$ respectively for comparable fitting quality. We also include Instant-NGP~\cite{muller2022instant} and original Gaussian Splatting~\cite{kerbl20233d} for reference despite their unsuitability for generative modeling due to their unstructured spatial nature. As shown in~\Cref{tab:exp_fitting}, our GaussianCube outperforms all NeRF-based representations among all metrics.~\Cref{fig:representation_building} illustrates that GaussianCube can faithfully reconstruct geometry details and intricate textures. Moreover, we achieve such high-quality fitting with orders of magnitude fewer parameters than previous structured representation due to the densification-constrained fitting, showcasing our compactness. Notably, the shared implicit feature decoder in the multi-object fitting of NeRF-based methods leads to significant decreases in quality compared to single-object fitting as evidenced in \Cref{tab:exp_fitting}. While the fully explicit nature of GS results in no quality gap between single and multiple object fitting.

\noindent\textbf{Single-category unconditional generation.} For unconditional generation, we compare our method with the state-of-the-art 3D generation works including 3D-aware GANs~\cite{chan2022efficient,gao2022get3d} and Triplane diffusion models~\cite{cao2023large}. As shown in~\Cref{tab:uncond_clscond_gen}, our method surpasses all prior works in terms of both FID and KID scores and sets new records. We provide visual comparisons in~\Cref{fig:uncond_gen}, where EG3D and DiffTF tend to generate blurry results with poor geometry, and GET3D fails to provide satisfactory textures. In contrast, our method yields high-fidelity results with authentic geometry and sharp textures.

\begin{table}[t]  
	\centering
    \scriptsize
    \caption{Quantitative ablation of both representation fitting and generation quality on ShapeNet Car.} 
	\begin{tabular}{lccccc@{\hspace{3mm}}cc}  
		\hline  
		\multirow{2}{*}{\textbf{Method}} & \multirow{2}{*}{\textbf{Densify \& Prune}} & \multicolumn{3}{c}{\textbf{Representation Fitting}} & \multicolumn{2}{c}{\textbf{Generation}}\\
		& & \textbf{PSNR$\uparrow$} & \textbf{LPIPS$\downarrow$} & \textbf{SSIM$\uparrow$} & \textbf{FID-50K$\downarrow$} &
		\textbf{KID-50K(\textperthousand)$\downarrow$}\\  
		\hline  
		A. Voxel grid w/o offset & \xmark & 25.87  & 0.1228 & 0.9217 & - & -\\
		B. Voxel grid w/ offset & \xmark & 30.18  & 0.0780 & 0.9628 & 40.52 & 24.35 \\  
		\hline
		C. Ours w/o OT & \cmark & \textbf{34.94} & \textbf{0.0346} & \textbf{0.9863} & 21.41 & 14.37 \\
		\cellcolor{gray}\textbf{D. Ours} & \cellcolor{gray}\cmark & \cellcolor{gray}\textbf{34.94} & \cellcolor{gray}\textbf{0.0346} & \cellcolor{gray}\textbf{0.9863} & \cellcolor{gray}\textbf{13.01} & \cellcolor{gray}\textbf{8.46} \\  
		\hline  
	\end{tabular}  
	\label{tab:ablation_fitting_generation}  
\end{table} 

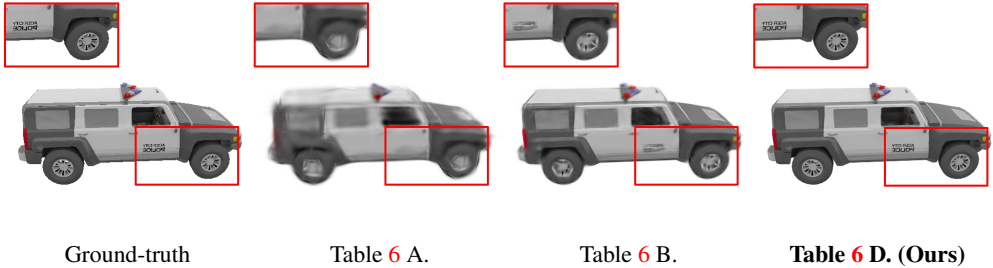
\begin{figure*}[t]
	\centering
	\small
	\begin{overpic}[width=0.95\linewidth]{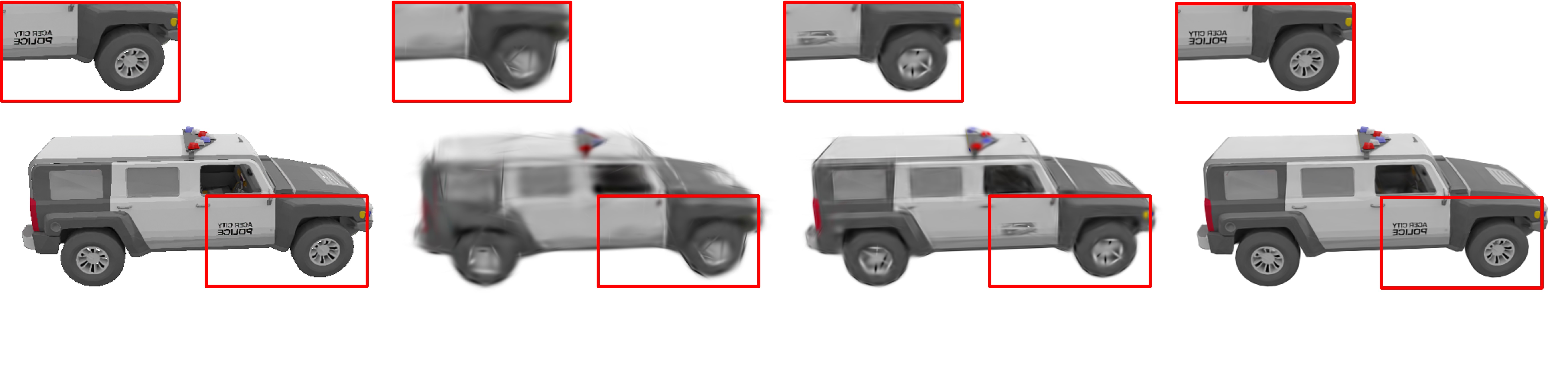}
		\put(6,-1){Ground-truth}
		\put(32,-1){~\Cref{tab:ablation_fitting_generation} A.}
		\put(57,-1){~\Cref{tab:ablation_fitting_generation} B.}
		\put(78,-1){\textbf{~\Cref{tab:ablation_fitting_generation} D. (Ours)}}
	\end{overpic}
	\caption{Qualitative ablation of representation fitting. }
	\label{fig:ablation_fitting}
\end{figure*}

\begin{figure*}[h!]
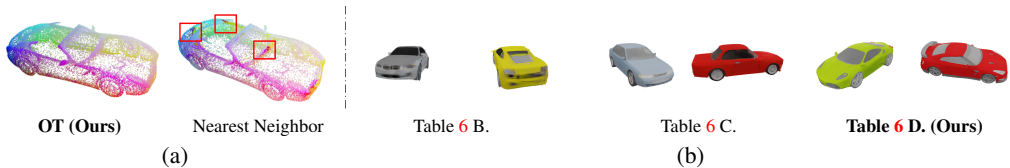

	\centering
	\scriptsize
	\setlength\tabcolsep{1pt}
	\begin{overpic}[width=0.98\linewidth]{imgs/results/ablation/ablation_mapping_results_small.jpg}
		\put(5,-2){\textbf{OT (Ours)}}
		\put(20,-2){Nearest Neighbor}
		\put(41,-2){~\Cref{tab:ablation_fitting_generation} B.}
		\put(65,-2){~\Cref{tab:ablation_fitting_generation} C.}
		\put(83,-2){\textbf{~\Cref{tab:ablation_fitting_generation} D. (Ours)}}
		\put(17,-5){\small{(a)}}
		\put(67,-5){\small{(b)}}
	\end{overpic}
	\vspace{6mm}
	\caption{Visual ablation of the Gaussian organization methods and 3D generation. For visualization of Gaussian structuralization in (a), we map the coordinates of the corresponding voxel of each Gaussians to RGB values to visualize the organization. Our OT-based solution also results in the best generation quality shown in (b).}
	\label{fig:ablation_mapping_generation}
\end{figure*}

\noindent\textbf{Large-vocabulary class-conditioned generation.} We also compare class-conditioned generation with DiffTF~\cite{cao2023large} on more diverse and challenging OmniObject3D~\cite{wu2023omniobject3d} dataset. We achieve significantly better FID and KID scores than DiffTF as shown in~\Cref{tab:uncond_clscond_gen}.
Visual comparisons in~\Cref{fig:clscond_gen} reveal that DiffTF often struggles to create intricate geometry and detailed textures, whereas our method is able to generate objects with complex geometry and realistic textures.

\noindent\textbf{Image-conditioned avatar generation.} For 3D avatar generation conditioned on a single reference image, we compare our method with state-of-the-art Triplane diffusion models, Rodin~\cite{rombach2022high}. Our model surpasses Rodin among all evaluated metrics as shown in~\Cref{tab:avatar_gen}. Although Rodin utilizes a 2D refiner~\cite{wang2021towards} to boost the visual quality of facial areas, which significantly compromises 3D consistency. Our model still outperforms it by direct real-3D generation. Results in~\Cref{fig:real_world_avatars} demonstrate that our model faithfully preserves the identity, expression and accessories of the references with rich details, while Rodin struggles to provide satisfactory results even using 2D refinement.

\noindent\textbf{Text-to-3D generation.} We compare text-to-3D generation with prior arts including diffusion models~\cite{jun2023shap,tang2023volumediffusion}, optimization-based method~\cite{tang2023dreamgaussian} and feed-forward method~\cite{tang2024lgm}. Our model achieves competitive text-3D alignment results as shown in~\Cref{tab:textcond_gen}. The visual comparison in~\Cref{fig:textcond_gen} shows that our model is able to create high-quality samples aligning with text prompts in just $2.3$ seconds. DreamGaussian tends to create over-saturated results and suffers from Janus problem. VolumeDiffusion produces unsatisfactory textures with poor text alignment. Shap-E can produce semantically accurate results but struggles to generate complex geometry. LGM reconstructs 3D Gaussians from multi-view images generated by text-conditioned multi-view diffusion pipeline~\cite{shi2023mvdream}, whereas the inconsistency~\cite{tang2024lgm} of the generated multi-views often results in inaccurate geometric reconstruction.

\subsection{Ablation Study}

We first examine the key factors in representation construction on ShapeNet Car. To spatially structure the Gaussians, a simplistic approach would be anchoring the positions of Gaussians to a predefined voxel grid while omitting densification and pruning, which leads to severe failure when fitting the objects as shown in~\Cref{fig:ablation_fitting}. Even by introducing learnable offsets to the voxel grid, 
the results still lack details. 
We observe the offsets are typically too small to effectively lead the Gaussians close to the object surfaces, which indicates the importance of densification in the fitting process. Instead, GaussianCube can capture both complex geometry and intricate details as shown in~\Cref{fig:ablation_fitting}. The numerical comparison in~\Cref{tab:ablation_fitting_generation} also demonstrates the superior fitting quality of GaussianCube.

We also evaluate how the representation affects 3D generative modeling on ShapeNet Car as shown in~\Cref{tab:ablation_fitting_generation} and \Cref{fig:ablation_mapping_generation}. Limited by the poor fitting quality, performing diffusion modeling on voxel grid with learnable offsets leads to blurry generation results as shown in~\Cref{fig:ablation_mapping_generation}. To validate the importance of organizing Gaussians via Optimal Transport (OT), we compare with the organization based on nearest neighbor transport. We linearly map each Gaussian’s corresponding coordinates of voxel to RGB color to visualize different organizations. As shown in~\Cref{fig:ablation_mapping_generation} (a), our proposed OT approach yields smooth color transitions, indicating that our method successfully preserves the spatial correspondence. However, nearest neighbor results in abrupt color transitions due to their disregard for global structure. Both the quantitative results in~\Cref{tab:ablation_fitting_generation} and visual comparisons~\Cref{fig:ablation_mapping_generation} indicate that our globally structured arrangement facilitates generative modeling by alleviating its complexity, successfully leading to superior generation quality.

\section{Conclusion}
We have presented GaussianCube, a structured and explicit radiance representation crafted for 3D generative models. We begin by fitting each 3D object with a constant number of Gaussians using our proposed densification-constrained fitting algorithm. We further organize the obtained Gaussians into a spatially structured representation by solving the Optimal Transport between the positions of Gaussians and the predefined voxel grid. The proposed GaussianCube is spatially structured, allowing to use standard 3D U-Net for diffusion modeling without elaborate designs. Moreover, GaussianCube can achieve high-quality fitting using much fewer parameters compared to prior works of similar quality, which further eases the difficulty of generative modeling. Our 3D diffusion models equipped with GaussianCube achieve state-of-the-art generation quality on the evaluated datasets, underscoring its potential of GaussianCube as a versatile and powerful radiance representation for 3D generation.

\noindent\textbf{Acknowledgments:} This work was supported in part by the Anhui Provincial Natural Science Foundation under Grant 2108085UD12. We acknowledge the support of GPU cluster built by MCC Lab of Information Science and Technology Institution, USTC. We also thank anonymous reviewers for their valuable comments.

\clearpage
\setcitestyle{numbers}
\bibliography{main}

\begin{thebibliography}{77}
\providecommand{\natexlab}[1]{#1}
\providecommand{\url}[1]{\texttt{#1}}
\expandafter\ifx\csname urlstyle\endcsname\relax
  \providecommand{\doi}[1]{doi: #1}\else
  \providecommand{\doi}{doi: \begingroup \urlstyle{rm}\Url}\fi

\bibitem[Barron et~al.(2022)Barron, Mildenhall, Verbin, Srinivasan, and Hedman]{barron2022mip}
Jonathan~T Barron, Ben Mildenhall, Dor Verbin, Pratul~P Srinivasan, and Peter Hedman.
\newblock Mip-nerf 360: Unbounded anti-aliased neural radiance fields.
\newblock In \emph{Proceedings of the IEEE/CVF Conference on Computer Vision and Pattern Recognition}, pages 5470--5479, 2022.

\bibitem[Bi{\'n}kowski et~al.(2018)Bi{\'n}kowski, Sutherland, Arbel, and Gretton]{binkowski2018demystifying}
Miko{\l}aj Bi{\'n}kowski, Danica~J Sutherland, Michael Arbel, and Arthur Gretton.
\newblock Demystifying mmd gans.
\newblock \emph{arXiv preprint arXiv:1801.01401}, 2018.

\bibitem[Blattmann et~al.(2023)Blattmann, Rombach, Ling, Dockhorn, Kim, Fidler, and Kreis]{blattmann2023align}
Andreas Blattmann, Robin Rombach, Huan Ling, Tim Dockhorn, Seung~Wook Kim, Sanja Fidler, and Karsten Kreis.
\newblock Align your latents: High-resolution video synthesis with latent diffusion models.
\newblock In \emph{Proceedings of the IEEE/CVF Conference on Computer Vision and Pattern Recognition}, pages 22563--22575, 2023.

\bibitem[Burkard and Cela(1999)]{burkard1999linear}
Rainer~E Burkard and Eranda Cela.
\newblock Linear assignment problems and extensions.
\newblock In \emph{Handbook of combinatorial optimization: Supplement volume A}, pages 75--149. Springer, 1999.

\bibitem[Cao et~al.(2023)Cao, Hong, Wu, Pan, and Liu]{cao2023large}
Ziang Cao, Fangzhou Hong, Tong Wu, Liang Pan, and Ziwei Liu.
\newblock Large-vocabulary 3d diffusion model with transformer.
\newblock \emph{arXiv preprint arXiv:2309.07920}, 2023.

\bibitem[Caron et~al.(2021)Caron, Touvron, Misra, J{\'e}gou, Mairal, Bojanowski, and Joulin]{caron2021emerging}
Mathilde Caron, Hugo Touvron, Ishan Misra, Herv{\'e} J{\'e}gou, Julien Mairal, Piotr Bojanowski, and Armand Joulin.
\newblock Emerging properties in self-supervised vision transformers.
\newblock In \emph{Proceedings of the IEEE/CVF International Conference on Computer Vision}, pages 9650--9660, 2021.

\bibitem[Chan et~al.(2021)Chan, Monteiro, Kellnhofer, Wu, and Wetzstein]{chan2021pi}
Eric~R Chan, Marco Monteiro, Petr Kellnhofer, Jiajun Wu, and Gordon Wetzstein.
\newblock pi-gan: Periodic implicit generative adversarial networks for 3d-aware image synthesis.
\newblock In \emph{Proceedings of the IEEE/CVF Conference on Computer Vision and Pattern Recognition}, pages 5799--5809, 2021.

\bibitem[Chan et~al.(2022)Chan, Lin, Chan, Nagano, Pan, De~Mello, Gallo, Guibas, Tremblay, Khamis, et~al.]{chan2022efficient}
Eric~R Chan, Connor~Z Lin, Matthew~A Chan, Koki Nagano, Boxiao Pan, Shalini De~Mello, Orazio Gallo, Leonidas~J Guibas, Jonathan Tremblay, Sameh Khamis, et~al.
\newblock Efficient geometry-aware 3d generative adversarial networks.
\newblock In \emph{Proceedings of the IEEE/CVF Conference on Computer Vision and Pattern Recognition}, pages 16123--16133, 2022.

\bibitem[Chang et~al.(2015)Chang, Funkhouser, Guibas, Hanrahan, Huang, Li, Savarese, Savva, Song, Su, et~al.]{chang2015shapenet}
Angel~X Chang, Thomas Funkhouser, Leonidas Guibas, Pat Hanrahan, Qixing Huang, Zimo Li, Silvio Savarese, Manolis Savva, Shuran Song, Hao Su, et~al.
\newblock Shapenet: An information-rich 3d model repository.
\newblock \emph{arXiv preprint arXiv:1512.03012}, 2015.

\bibitem[Chen and Wang(2024)]{chen2024survey}
Guikun Chen and Wenguan Wang.
\newblock A survey on 3d gaussian splatting.
\newblock \emph{arXiv preprint arXiv:2401.03890}, 2024.

\bibitem[Chen et~al.(2023)Chen, Gu, Chen, Tian, Tu, Liu, and Su]{chen2023single}
Hansheng Chen, Jiatao Gu, Anpei Chen, Wei Tian, Zhuowen Tu, Lingjie Liu, and Hao Su.
\newblock Single-stage diffusion nerf: A unified approach to 3d generation and reconstruction.
\newblock \emph{arXiv preprint arXiv:2304.06714}, 2023.

\bibitem[Cheng et~al.(2023)Cheng, Yin, Huang, Yu, Liu, Feng, Yang, and Tang]{cheng2023efficient}
Yiji Cheng, Fei Yin, Xiaoke Huang, Xintong Yu, Jiaxiang Liu, Shikun Feng, Yujiu Yang, and Yansong Tang.
\newblock Efficient text-guided 3d-aware portrait generation with score distillation sampling on distribution.
\newblock \emph{arXiv preprint arXiv:2306.02083}, 2023.

\bibitem[Cotton and Peyton(2024)]{cotton2024dynamic}
R~James Cotton and Colleen Peyton.
\newblock Dynamic gaussian splatting from markerless motion capture reconstruct infants movements.
\newblock In \emph{Proceedings of the IEEE/CVF Winter Conference on Applications of Computer Vision}, pages 60--68, 2024.

\bibitem[Deitke et~al.(2023)Deitke, Schwenk, Salvador, Weihs, Michel, VanderBilt, Schmidt, Ehsani, Kembhavi, and Farhadi]{deitke2023objaverse}
Matt Deitke, Dustin Schwenk, Jordi Salvador, Luca Weihs, Oscar Michel, Eli VanderBilt, Ludwig Schmidt, Kiana Ehsani, Aniruddha Kembhavi, and Ali Farhadi.
\newblock Objaverse: A universe of annotated 3d objects.
\newblock In \emph{Proceedings of the IEEE/CVF Conference on Computer Vision and Pattern Recognition}, pages 13142--13153, 2023.

\bibitem[Deng et~al.(2019)Deng, Guo, Xue, and Zafeiriou]{deng2019arcface}
Jiankang Deng, Jia Guo, Niannan Xue, and Stefanos Zafeiriou.
\newblock Arcface: Additive angular margin loss for deep face recognition.
\newblock In \emph{Proceedings of the IEEE/CVF Conference on Computer Vision and Pattern Recognition}, pages 4690--4699, 2019.

\bibitem[Deng et~al.(2022)Deng, Yang, Xiang, and Tong]{deng2022gram}
Yu~Deng, Jiaolong Yang, Jianfeng Xiang, and Xin Tong.
\newblock Gram: Generative radiance manifolds for 3d-aware image generation.
\newblock In \emph{Proceedings of the IEEE/CVF Conference on Computer Vision and Pattern Recognition}, pages 10673--10683, 2022.

\bibitem[Dhariwal and Nichol(2021)]{dhariwal2021diffusion}
Prafulla Dhariwal and Alexander Nichol.
\newblock Diffusion models beat gans on image synthesis.
\newblock \emph{Advances in Neural Information Processing Systems}, 34:\penalty0 8780--8794, 2021.

\bibitem[Fridovich-Keil et~al.(2022)Fridovich-Keil, Yu, Tancik, Chen, Recht, and Kanazawa]{fridovich2022plenoxels}
Sara Fridovich-Keil, Alex Yu, Matthew Tancik, Qinhong Chen, Benjamin Recht, and Angjoo Kanazawa.
\newblock Plenoxels: Radiance fields without neural networks.
\newblock In \emph{Proceedings of the IEEE/CVF Conference on Computer Vision and Pattern Recognition}, pages 5501--5510, 2022.

\bibitem[Gao et~al.(2022)Gao, Shen, Wang, Chen, Yin, Li, Litany, Gojcic, and Fidler]{gao2022get3d}
Jun Gao, Tianchang Shen, Zian Wang, Wenzheng Chen, Kangxue Yin, Daiqing Li, Or~Litany, Zan Gojcic, and Sanja Fidler.
\newblock Get3d: A generative model of high quality 3d textured shapes learned from images.
\newblock \emph{arXiv preprint arXiv:2209.11163}, 2022.

\bibitem[Goodfellow et~al.(2020)Goodfellow, Pouget-Abadie, Mirza, Xu, Warde-Farley, Ozair, Courville, and Bengio]{goodfellow2020generative}
Ian Goodfellow, Jean Pouget-Abadie, Mehdi Mirza, Bing Xu, David Warde-Farley, Sherjil Ozair, Aaron Courville, and Yoshua Bengio.
\newblock Generative adversarial networks.
\newblock \emph{Communications of the ACM}, 63\penalty0 (11):\penalty0 139--144, 2020.

\bibitem[Gu et~al.(2021)Gu, Liu, Wang, and Theobalt]{gu2021stylenerf}
Jiatao Gu, Lingjie Liu, Peng Wang, and Christian Theobalt.
\newblock Stylenerf: A style-based 3d-aware generator for high-resolution image synthesis.
\newblock \emph{arXiv preprint arXiv:2110.08985}, 2021.

\bibitem[Gupta et~al.(2023)Gupta, Xiong, Nie, Jones, and O{\u{g}}uz]{gupta20233dgen}
Anchit Gupta, Wenhan Xiong, Yixin Nie, Ian Jones, and Barlas O{\u{g}}uz.
\newblock 3dgen: Triplane latent diffusion for textured mesh generation.
\newblock \emph{arXiv preprint arXiv:2303.05371}, 2023.

\bibitem[He et~al.(2024)He, Chen, Peng, Huang, Li, Huang, Yuan, Ouyang, and He]{he2024gvgen}
Xianglong He, Junyi Chen, Sida Peng, Di~Huang, Yangguang Li, Xiaoshui Huang, Chun Yuan, Wanli Ouyang, and Tong He.
\newblock Gvgen: Text-to-3d generation with volumetric representation.
\newblock \emph{arXiv preprint arXiv:2403.12957}, 2024.

\bibitem[Heusel et~al.(2017)Heusel, Ramsauer, Unterthiner, Nessler, and Hochreiter]{heusel2017gans}
Martin Heusel, Hubert Ramsauer, Thomas Unterthiner, Bernhard Nessler, and Sepp Hochreiter.
\newblock Gans trained by a two time-scale update rule converge to a local nash equilibrium.
\newblock \emph{Advances in Neural Information Processing Systems}, 30, 2017.

\bibitem[Ho et~al.(2020)Ho, Jain, and Abbeel]{ho2020denoising}
Jonathan Ho, Ajay Jain, and Pieter Abbeel.
\newblock Denoising diffusion probabilistic models.
\newblock \emph{Advances in Neural Information Processing Systems}, 33:\penalty0 6840--6851, 2020.

\bibitem[Hu et~al.(2023)Hu, Wang, Ma, Yang, Gao, Liu, and Ma]{hu2023tri}
Wenbo Hu, Yuling Wang, Lin Ma, Bangbang Yang, Lin Gao, Xiao Liu, and Yuewen Ma.
\newblock Tri-miprf: Tri-mip representation for efficient anti-aliasing neural radiance fields.
\newblock In \emph{Proceedings of the IEEE/CVF International Conference on Computer Vision}, pages 19774--19783, 2023.

\bibitem[Jonker and Volgenant(1988)]{jonker1988shortest}
Roy Jonker and Ton Volgenant.
\newblock A shortest augmenting path algorithm for dense and sparse linear assignment problems.
\newblock In \emph{DGOR/NSOR: Papers of the 16th Annual Meeting of DGOR in Cooperation with NSOR/Vortr{\"a}ge der 16. Jahrestagung der DGOR zusammen mit der NSOR}, pages 622--622. Springer, 1988.

\bibitem[Jun and Nichol(2023)]{jun2023shap}
Heewoo Jun and Alex Nichol.
\newblock Shap-e: Generating conditional 3d implicit functions.
\newblock \emph{arXiv preprint arXiv:2305.02463}, 2023.

\bibitem[Karras et~al.(2019)Karras, Laine, and Aila]{karras2019style}
Tero Karras, Samuli Laine, and Timo Aila.
\newblock A style-based generator architecture for generative adversarial networks.
\newblock In \emph{Proceedings of the IEEE/CVF Conference on Computer Vision and Pattern Recognition}, pages 4401--4410, 2019.

\bibitem[Kerbl et~al.(2023)Kerbl, Kopanas, Leimk{\"u}hler, and Drettakis]{kerbl20233d}
Bernhard Kerbl, Georgios Kopanas, Thomas Leimk{\"u}hler, and George Drettakis.
\newblock 3d gaussian splatting for real-time radiance field rendering.
\newblock \emph{ACM Transactions on Graphics}, 42\penalty0 (4), 2023.

\bibitem[Li et~al.(2024)Li, Yao, Xie, Chen, and Jiang]{li2024gaussianbody}
Mengtian Li, Shengxiang Yao, Zhifeng Xie, Keyu Chen, and Yu-Gang Jiang.
\newblock Gaussianbody: Clothed human reconstruction via 3d gaussian splatting.
\newblock \emph{arXiv preprint arXiv:2401.09720}, 2024.

\bibitem[Li et~al.(2023)Li, Tucker, Snavely, and Holynski]{li2023generative}
Zhengqi Li, Richard Tucker, Noah Snavely, and Aleksander Holynski.
\newblock Generative image dynamics.
\newblock \emph{arXiv preprint arXiv:2309.07906}, 2023.

\bibitem[Loshchilov and Hutter(2019)]{adamw}
Ilya Loshchilov and Frank Hutter.
\newblock Decoupled weight decay regularization.
\newblock In \emph{International Conference on Learning Representations, {ICLR}}, 2019.

\bibitem[Lu et~al.(2022)Lu, Zhou, Bao, Chen, Li, and Zhu]{lu2022dpm}
Cheng Lu, Yuhao Zhou, Fan Bao, Jianfei Chen, Chongxuan Li, and Jun Zhu.
\newblock Dpm-solver: A fast ode solver for diffusion probabilistic model sampling in around 10 steps.
\newblock \emph{Advances in Neural Information Processing Systems}, 35:\penalty0 5775--5787, 2022.

\bibitem[Lu et~al.(2024)Lu, Zhang, Wang, Liu, Lu, and Tang]{lu2024manigaussian}
Guanxing Lu, Shiyi Zhang, Ziwei Wang, Changliu Liu, Jiwen Lu, and Yansong Tang.
\newblock Manigaussian: Dynamic gaussian splatting for multi-task robotic manipulation.
\newblock \emph{arXiv preprint arXiv:2403.08321}, 2024.

\bibitem[Luiten et~al.(2023)Luiten, Kopanas, Leibe, and Ramanan]{luiten2023dynamic}
Jonathon Luiten, Georgios Kopanas, Bastian Leibe, and Deva Ramanan.
\newblock Dynamic 3d gaussians: Tracking by persistent dynamic view synthesis.
\newblock \emph{arXiv preprint arXiv:2308.09713}, 2023.

\bibitem[Meng et~al.(2021)Meng, He, Song, Song, Wu, Zhu, and Ermon]{meng2021sdedit}
Chenlin Meng, Yutong He, Yang Song, Jiaming Song, Jiajun Wu, Jun-Yan Zhu, and Stefano Ermon.
\newblock Sdedit: Guided image synthesis and editing with stochastic differential equations.
\newblock \emph{arXiv preprint arXiv:2108.01073}, 2021.

\bibitem[Mildenhall et~al.(2021)Mildenhall, Srinivasan, Tancik, Barron, Ramamoorthi, and Ng]{mildenhall2021nerf}
Ben Mildenhall, Pratul~P Srinivasan, Matthew Tancik, Jonathan~T Barron, Ravi Ramamoorthi, and Ren Ng.
\newblock Nerf: Representing scenes as neural radiance fields for view synthesis.
\newblock \emph{Communications of the ACM}, 65\penalty0 (1):\penalty0 99--106, 2021.

\bibitem[M{\"u}ller et~al.(2023)M{\"u}ller, Siddiqui, Porzi, Bulo, Kontschieder, and Nie{\ss}ner]{muller2023diffrf}
Norman M{\"u}ller, Yawar Siddiqui, Lorenzo Porzi, Samuel~Rota Bulo, Peter Kontschieder, and Matthias Nie{\ss}ner.
\newblock Diffrf: Rendering-guided 3d radiance field diffusion.
\newblock In \emph{Proceedings of the IEEE/CVF Conference on Computer Vision and Pattern Recognition}, pages 4328--4338, 2023.

\bibitem[M{\"u}ller et~al.(2022)M{\"u}ller, Evans, Schied, and Keller]{muller2022instant}
Thomas M{\"u}ller, Alex Evans, Christoph Schied, and Alexander Keller.
\newblock Instant neural graphics primitives with a multiresolution hash encoding.
\newblock \emph{ACM Transactions on Graphics (ToG)}, 41\penalty0 (4):\penalty0 1--15, 2022.

\bibitem[Nichol and Dhariwal(2021)]{nichol2021improved}
Alexander~Quinn Nichol and Prafulla Dhariwal.
\newblock Improved denoising diffusion probabilistic models.
\newblock In \emph{International Conference on Machine Learning}, pages 8162--8171. PMLR, 2021.

\bibitem[Niemeyer and Geiger(2021)]{niemeyer2021giraffe}
Michael Niemeyer and Andreas Geiger.
\newblock Giraffe: Representing scenes as compositional generative neural feature fields.
\newblock In \emph{Proceedings of the IEEE/CVF Conference on Computer Vision and Pattern Recognition}, pages 11453--11464, 2021.

\bibitem[Park et~al.(2021)Park, Sinha, Barron, Bouaziz, Goldman, Seitz, and Martin-Brualla]{park2021nerfies}
Keunhong Park, Utkarsh Sinha, Jonathan~T Barron, Sofien Bouaziz, Dan~B Goldman, Steven~M Seitz, and Ricardo Martin-Brualla.
\newblock Nerfies: Deformable neural radiance fields.
\newblock In \emph{Proceedings of the IEEE/CVF International Conference on Computer Vision}, pages 5865--5874, 2021.

\bibitem[Poole et~al.(2022)Poole, Jain, Barron, and Mildenhall]{poole2022dreamfusion}
Ben Poole, Ajay Jain, Jonathan~T Barron, and Ben Mildenhall.
\newblock Dreamfusion: Text-to-3d using 2d diffusion.
\newblock \emph{arXiv preprint arXiv:2209.14988}, 2022.

\bibitem[Pumarola et~al.(2021)Pumarola, Corona, Pons-Moll, and Moreno-Noguer]{pumarola2021d}
Albert Pumarola, Enric Corona, Gerard Pons-Moll, and Francesc Moreno-Noguer.
\newblock D-nerf: Neural radiance fields for dynamic scenes.
\newblock In \emph{Proceedings of the IEEE/CVF Conference on Computer Vision and Pattern Recognition}, pages 10318--10327, 2021.

\bibitem[Radford et~al.(2021)Radford, Kim, Hallacy, Ramesh, Goh, Agarwal, Sastry, Askell, Mishkin, Clark, et~al.]{radford2021learning}
Alec Radford, Jong~Wook Kim, Chris Hallacy, Aditya Ramesh, Gabriel Goh, Sandhini Agarwal, Girish Sastry, Amanda Askell, Pamela Mishkin, Jack Clark, et~al.
\newblock Learning transferable visual models from natural language supervision.
\newblock In \emph{International Conference on Machine Learning}, pages 8748--8763. PMLR, 2021.

\bibitem[Rombach et~al.(2022)Rombach, Blattmann, Lorenz, Esser, and Ommer]{rombach2022high}
Robin Rombach, Andreas Blattmann, Dominik Lorenz, Patrick Esser, and Bj{\"o}rn Ommer.
\newblock High-resolution image synthesis with latent diffusion models.
\newblock In \emph{Proceedings of the IEEE/CVF Conference on Computer Vision and Pattern Recognition}, pages 10684--10695, 2022.

\bibitem[Shi et~al.(2023)Shi, Wang, Ye, Long, Li, and Yang]{shi2023mvdream}
Yichun Shi, Peng Wang, Jianglong Ye, Mai Long, Kejie Li, and Xiao Yang.
\newblock Mvdream: Multi-view diffusion for 3d generation.
\newblock \emph{arXiv preprint arXiv:2308.16512}, 2023.

\bibitem[Shue et~al.(2023)Shue, Chan, Po, Ankner, Wu, and Wetzstein]{shue20233d}
J~Ryan Shue, Eric~Ryan Chan, Ryan Po, Zachary Ankner, Jiajun Wu, and Gordon Wetzstein.
\newblock 3d neural field generation using triplane diffusion.
\newblock In \emph{Proceedings of the IEEE/CVF Conference on Computer Vision and Pattern Recognition}, pages 20875--20886, 2023.

\bibitem[Simonyan and Zisserman(2014)]{simonyan2014very}
Karen Simonyan and Andrew Zisserman.
\newblock Very deep convolutional networks for large-scale image recognition.
\newblock \emph{arXiv preprint arXiv:1409.1556}, 2014.

\bibitem[Sun et~al.(2022)Sun, Sun, and Chen]{sun2022direct}
Cheng Sun, Min Sun, and Hwann-Tzong Chen.
\newblock Direct voxel grid optimization: Super-fast convergence for radiance fields reconstruction.
\newblock In \emph{Proceedings of the IEEE/CVF Conference on Computer Vision and Pattern Recognition}, pages 5459--5469, 2022.

\bibitem[Sun et~al.(2023)Sun, Zhang, Shao, Wang, Liu, Xie, and Liu]{sun2023dreamcraft3d}
Jingxiang Sun, Bo~Zhang, Ruizhi Shao, Lizhen Wang, Wen Liu, Zhenda Xie, and Yebin Liu.
\newblock Dreamcraft3d: Hierarchical 3d generation with bootstrapped diffusion prior.
\newblock \emph{arXiv preprint arXiv:2310.16818}, 2023.

\bibitem[Tang et~al.(2023{\natexlab{a}})Tang, Ren, Zhou, Liu, and Zeng]{tang2023dreamgaussian}
Jiaxiang Tang, Jiawei Ren, Hang Zhou, Ziwei Liu, and Gang Zeng.
\newblock Dreamgaussian: Generative gaussian splatting for efficient 3d content creation.
\newblock \emph{arXiv preprint arXiv:2309.16653}, 2023{\natexlab{a}}.

\bibitem[Tang et~al.(2024{\natexlab{a}})Tang, Chen, Chen, Wang, Zeng, and Liu]{tang2024lgm}
Jiaxiang Tang, Zhaoxi Chen, Xiaokang Chen, Tengfei Wang, Gang Zeng, and Ziwei Liu.
\newblock Lgm: Large multi-view gaussian model for high-resolution 3d content creation.
\newblock \emph{arXiv preprint arXiv:2402.05054}, 2024{\natexlab{a}}.

\bibitem[Tang et~al.(2023{\natexlab{b}})Tang, Wang, Zhang, Zhang, Yi, Ma, and Chen]{tang2023make}
Junshu Tang, Tengfei Wang, Bo~Zhang, Ting Zhang, Ran Yi, Lizhuang Ma, and Dong Chen.
\newblock Make-it-3d: High-fidelity 3d creation from a single image with diffusion prior.
\newblock In \emph{Proceedings of the IEEE/CVF International Conference on Computer Vision}, pages 22819--22829, 2023{\natexlab{b}}.

\bibitem[Tang et~al.(2024{\natexlab{b}})Tang, Zeng, Fan, Wang, Dai, Chen, and Ma]{tang2024make}
Junshu Tang, Yanhong Zeng, Ke~Fan, Xuheng Wang, Bo~Dai, Kai Chen, and Lizhuang Ma.
\newblock Make-it-vivid: Dressing your animatable biped cartoon characters from text.
\newblock In \emph{Proceedings of the IEEE/CVF Conference on Computer Vision and Pattern Recognition}, pages 6243--6253, 2024{\natexlab{b}}.

\bibitem[Tang et~al.(2023{\natexlab{c}})Tang, Gu, Wang, Zhang, Bao, Chen, and Guo]{tang2023volumediffusion}
Zhicong Tang, Shuyang Gu, Chunyu Wang, Ting Zhang, Jianmin Bao, Dong Chen, and Baining Guo.
\newblock Volumediffusion: Flexible text-to-3d generation with efficient volumetric encoder.
\newblock \emph{arXiv preprint arXiv:2312.11459}, 2023{\natexlab{c}}.

\bibitem[Villani et~al.(2009)]{villani2009optimal}
C{\'e}dric Villani et~al.
\newblock \emph{Optimal transport: old and new}, volume 338.
\newblock Springer, 2009.

\bibitem[Wang et~al.(2023)Wang, Zhang, Zhang, Gu, Bao, Baltrusaitis, Shen, Chen, Wen, Chen, et~al.]{wang2023rodin}
Tengfei Wang, Bo~Zhang, Ting Zhang, Shuyang Gu, Jianmin Bao, Tadas Baltrusaitis, Jingjing Shen, Dong Chen, Fang Wen, Qifeng Chen, et~al.
\newblock Rodin: A generative model for sculpting 3d digital avatars using diffusion.
\newblock In \emph{Proceedings of the IEEE/CVF Conference on Computer Vision and Pattern Recognition}, pages 4563--4573, 2023.

\bibitem[Wang et~al.(2021)Wang, Li, Zhang, and Shan]{wang2021towards}
Xintao Wang, Yu~Li, Honglun Zhang, and Ying Shan.
\newblock Towards real-world blind face restoration with generative facial prior.
\newblock In \emph{Proceedings of the IEEE/CVF Conference on Computer Vision and Pattern Recognition}, pages 9168--9178, 2021.

\bibitem[Wang et~al.(2024)Wang, Wang, He, Hancke, Liu, and Lau]{wang2024phidias}
Zhenwei Wang, Tengfei Wang, Zexin He, Gerhard Hancke, Ziwei Liu, and Rynson~WH Lau.
\newblock Phidias: A generative model for creating 3d content from text, image, and 3d conditions with reference-augmented diffusion.
\newblock \emph{arXiv preprint arXiv:2409.11406}, 2024.

\bibitem[Wood et~al.(2021)Wood, Baltru{\v{s}}aitis, Hewitt, Dziadzio, Cashman, and Shotton]{wood2021fake}
Erroll Wood, Tadas Baltru{\v{s}}aitis, Charlie Hewitt, Sebastian Dziadzio, Thomas~J Cashman, and Jamie Shotton.
\newblock Fake it till you make it: face analysis in the wild using synthetic data alone.
\newblock In \emph{Proceedings of the IEEE/CVF International Conference on Computer Vision}, pages 3681--3691, 2021.

\bibitem[Wu et~al.(2023{\natexlab{a}})Wu, Yi, Fang, Xie, Zhang, Wei, Liu, Tian, and Wang]{wu20234d}
Guanjun Wu, Taoran Yi, Jiemin Fang, Lingxi Xie, Xiaopeng Zhang, Wei Wei, Wenyu Liu, Qi~Tian, and Xinggang Wang.
\newblock 4d gaussian splatting for real-time dynamic scene rendering.
\newblock \emph{arXiv preprint arXiv:2310.08528}, 2023{\natexlab{a}}.

\bibitem[Wu et~al.(2023{\natexlab{b}})Wu, Zhang, Fu, Wang, Ren, Pan, Wu, Yang, Wang, Qian, et~al.]{wu2023omniobject3d}
Tong Wu, Jiarui Zhang, Xiao Fu, Yuxin Wang, Jiawei Ren, Liang Pan, Wayne Wu, Lei Yang, Jiaqi Wang, Chen Qian, et~al.
\newblock Omniobject3d: Large-vocabulary 3d object dataset for realistic perception, reconstruction and generation.
\newblock In \emph{Proceedings of the IEEE/CVF Conference on Computer Vision and Pattern Recognition}, pages 803--814, 2023{\natexlab{b}}.

\bibitem[Xia and Xue(2023)]{xia2023survey}
Weihao Xia and Jing-Hao Xue.
\newblock A survey on deep generative 3d-aware image synthesis.
\newblock \emph{ACM Computing Surveys}, 56\penalty0 (4):\penalty0 1--34, 2023.

\bibitem[Xiang et~al.(2022)Xiang, Yang, Deng, and Tong]{xiang2022gram}
Jianfeng Xiang, Jiaolong Yang, Yu~Deng, and Xin Tong.
\newblock Gram-hd: 3d-consistent image generation at high resolution with generative radiance manifolds.
\newblock \emph{arXiv preprint arXiv:2206.07255}, 2022.

\bibitem[Xu et~al.(2022{\natexlab{a}})Xu, Wang, Cheng, Cao, Shan, Qie, and Gao]{xu2022dream3d}
Jiale Xu, Xintao Wang, Weihao Cheng, Yan-Pei Cao, Ying Shan, Xiaohu Qie, and Shenghua Gao.
\newblock Dream3d: Zero-shot text-to-3d synthesis using 3d shape prior and text-to-image diffusion models.
\newblock \emph{arXiv preprint arXiv:2212.14704}, 2022{\natexlab{a}}.

\bibitem[Xu et~al.(2022{\natexlab{b}})Xu, Xu, Philip, Bi, Shu, Sunkavalli, and Neumann]{xu2022point}
Qiangeng Xu, Zexiang Xu, Julien Philip, Sai Bi, Zhixin Shu, Kalyan Sunkavalli, and Ulrich Neumann.
\newblock Point-nerf: Point-based neural radiance fields.
\newblock In \emph{Proceedings of the IEEE/CVF Conference on Computer Vision and Pattern Recognition}, pages 5438--5448, 2022{\natexlab{b}}.

\bibitem[Xu et~al.(2023)Xu, Chen, Li, Zhang, Wang, Zheng, and Liu]{xu2023gaussian}
Yuelang Xu, Benwang Chen, Zhe Li, Hongwen Zhang, Lizhen Wang, Zerong Zheng, and Yebin Liu.
\newblock Gaussian head avatar: Ultra high-fidelity head avatar via dynamic gaussians.
\newblock \emph{arXiv preprint arXiv:2312.03029}, 2023.

\bibitem[Yi et~al.(2023)Yi, Fang, Wu, Xie, Zhang, Liu, Tian, and Wang]{yi2023gaussiandreamer}
Taoran Yi, Jiemin Fang, Guanjun Wu, Lingxi Xie, Xiaopeng Zhang, Wenyu Liu, Qi~Tian, and Xinggang Wang.
\newblock Gaussiandreamer: Fast generation from text to 3d gaussian splatting with point cloud priors.
\newblock \emph{arXiv preprint arXiv:2310.08529}, 2023.

\bibitem[Zhan et~al.(2024)Zhan, Shao, Wang, Yang, and Zhou]{zhan2024interactive}
Youyi Zhan, Tianjia Shao, He~Wang, Yin Yang, and Kun Zhou.
\newblock Interactive rendering of relightable and animatable gaussian avatars.
\newblock \emph{arXiv preprint arXiv:2407.10707}, 2024.

\bibitem[Zhang et~al.(2022)Zhang, Gu, Zhang, Bao, Chen, Wen, Wang, and Guo]{zhang2022styleswin}
Bowen Zhang, Shuyang Gu, Bo~Zhang, Jianmin Bao, Dong Chen, Fang Wen, Yong Wang, and Baining Guo.
\newblock Styleswin: Transformer-based gan for high-resolution image generation.
\newblock In \emph{Proceedings of the IEEE/CVF Conference on Computer Vision and Pattern Recognition}, pages 11304--11314, 2022.

\bibitem[Zhang et~al.(2024)Zhang, Cheng, Wang, Zhang, Yang, Tang, Zhao, Chen, and Guo]{zhang2024rodinhd}
Bowen Zhang, Yiji Cheng, Chunyu Wang, Ting Zhang, Jiaolong Yang, Yansong Tang, Feng Zhao, Dong Chen, and Baining Guo.
\newblock Rodinhd: High-fidelity 3d avatar generation with diffusion models.
\newblock \emph{arXiv preprint arXiv:2407.06938}, 2024.

\bibitem[Zhang et~al.(2020)Zhang, Riegler, Snavely, and Koltun]{zhang2020nerf++}
Kai Zhang, Gernot Riegler, Noah Snavely, and Vladlen Koltun.
\newblock Nerf++: Analyzing and improving neural radiance fields.
\newblock \emph{arXiv preprint arXiv:2010.07492}, 2020.

\bibitem[Zhang et~al.(2018)Zhang, Isola, Efros, Shechtman, and Wang]{zhang2018unreasonable}
Richard Zhang, Phillip Isola, Alexei~A Efros, Eli Shechtman, and Oliver Wang.
\newblock The unreasonable effectiveness of deep features as a perceptual metric.
\newblock In \emph{Proceedings of the IEEE/CVF Conference on Computer Vision and Pattern Recognition}, pages 586--595, 2018.

\bibitem[Zhou et~al.(2024)Zhou, Zhang, and Liu]{DiffGS}
Junsheng Zhou, Weiqi Zhang, and Yu-Shen Liu.
\newblock Diffgs: Functional gaussian splatting diffusion.
\newblock In \emph{Advances in Neural Information Processing Systems (NeurIPS)}, 2024.

\bibitem[Zhou et~al.(2021)Zhou, Du, and Wu]{zhou20213d}
Linqi Zhou, Yilun Du, and Jiajun Wu.
\newblock 3d shape generation and completion through point-voxel diffusion.
\newblock In \emph{Proceedings of the IEEE/CVF International Conference on Computer Vision}, pages 5826--5835, 2021.

\end{thebibliography}
\bibliographystyle{plainnat}

\newpage

\appendix
\section{Appendix}

\subsection{Additional Implementation Details}
\label{supp:implementation_details}

\noindent\textbf{Dataset preparation.} We conduct experiments on ShapeNet Car~\cite{chang2015shapenet}, ShapeNet Chair~\cite{chang2015shapenet}, OmniObject3D~\cite{wu2023omniobject3d}, Synthetic Avatar~\cite{wood2021fake} and Objaverse~\cite{deitke2023objaverse} datasets. For each dataset, we report the total number of objects used for training, the number of views rendered per object for GaussianCube fitting and the distribution of camera poses used for rendering in~\Cref{supp/tab:dataset_details}. For the Objaverse dataset, we excluded low-quality objects, such as those without textures or with defective reconstructions following~\cite{tang2023volumediffusion}. We also report the object bounding box $\bm{b}$ in the world coordinate system of each dataset in~\Cref{supp/tab:dataset_details}, which is used to construct the predefined voxel grid within $[-\bm{b}, \bm{b}]^3$ during OT-based Gaussian structuralization.

\noindent\textbf{Representation construction.} We set $N_{\text{max}}$ to 32768 and $C$ to 14 omitting the view-dependent spherical harmonics. This simplification appears to have a negligible impact on object fitting while concurrently reducing the data dimension, thereby alleviating the difficulty of diffusion modeling. During our densification-constrained fitting procedure, we primarily follow the hyper-parameters in original Gaussian Splatting~\cite{kerbl20233d}. For OT-based Gaussian structuralization, we adopt an approximate solution for the OT problem due to the $O\left(N_{\text{max}}^3\right)$ time complexity of Jonker-Volgenant algorithm~\cite{jonker1988shortest}. This is achieved by dividing the positions of the Gaussians and the voxel grid into four sorted segments and then applying the Jonker-Volgenant solver to each segment individually. We empirically find this approximation successfully strikes a balance between computational efficiency and spatial structure preservation. The proposed densification-constrained fitting takes around $2.67$ minutes for each object of 30K iterations and the OT-based voxelization takes around $2$ minutes which can be run on CPU in parallel.

\noindent\textbf{3D Diffusion.} To train the 3D diffusion model, we initially compute the instance-wise statistics of mean $\bar{\bm{\mu}} \in \mathbb{R}^{N_v \times N_v \times N_v \times C}$ and standard deviation $\bar{\bm{\sigma}} \in \mathbb{R}^{N_v \times N_v \times N_v \times C}$, from the GaussianCubes of each training dataset respectively. These statistical measures are then utilized to normalize the training data. For our 3D diffusion model architecture, we adopt the ADM-UNet from~\cite{dhariwal2021diffusion} and replace the convolution, upsampling, downsampling and attention operations with 3D implementations. We train our model using AdamW optimizer~\cite{adamw}, and apply exponential moving average (EMA) with a rate of 0.9999 during training. We clamp the prediction of opacity $\alpha$ to $[0,1)$ and clamp the minimum value of predicted scaling $\bm{s}$ to $0$ to ensure validity. For unconditional generation on ShapeNet, we train the model with a base learning rate $5e-5$ for 850K iterations and then decay the learning rate to $5e-6$ for another 150K iterations. For 3D digital avatar creation from a single portrait image, we adopt the pretrained DINO ViT-B/16~\cite{caron2021emerging} to encode the $512\times512$ conditional images into $1025\times768$ conditional feature tokens. For text-to-3D creation, we take CLIP-L/14~\cite{radford2021learning} to encode the text prompts into $77\times768$ conditional feature tokens. We provide more detailed configurations of the model architectures, diffusion training and inference for each dataset in~\Cref{supp/tab:training_details}.

\noindent\textbf{Implementation of Gaussian organization visualization in ~\Cref{fig:ablation_mapping_generation} (a).} For the $i$-th Gaussian, we obtain its corresponding voxel grid centers $\bm{x}_k \in \mathbb{R}^3$ according to Optimal Transport plan $\mathbf{T}^*$ (\textit{i}.\textit{e}., $\mathbf{T}^*_{i k}=1$) as illustrated in~\Cref{sec:representation_construction}. To visualize the coordinates of $\bm{x}_k$, we map them to RGB color $\bm{C}_k \in \mathbb{R}^3$ using:
\begin{equation}
    \bm{C}_k = \frac{(\bm{x}_k + \bm{b})}{2\bm{b}} \times \bm{255},
\end{equation}
where $\bm{b}$ is the bounding box in the world coordinate system. The resultant point cloud like visualizations are shown in~\Cref{fig:ablation_mapping_generation} (a), where smooth color transitions indicate coherent spatial correspondence preservation.

\subsection{Additional Ablation Study and Analysis}

\noindent\textbf{Ablation of $N_{\text{max}}$ in densification-constrained fitting.} We conduct experiments to evaluate how $N_{\text{max}}$ affects fitting on ShapeNet Car. The results in~\Cref{supp/tab:ablation_fitting} indicate that there is a clear trend where increasing $N_{\text{max}}$ leads to improved fitting accuracy. However, a larger $N_{\text{max}}$ also incurs higher computational costs during diffusion training. Therefore, we set $N_{\text{max}}$ to 32,768 to strike a balance between high-quality fitting and computational efficiency.

\noindent\textbf{Ablation of classifier-free guidance in class-conditioned generation.} We study how classifier-free guidance (CFG) impacts our generation quality when inference class-conditioned diffusion models. We report the FID and KID metrics in~\Cref{supp/tab:cfg_scale} under different CFG scales.

\begin{figure*}[h]
	\small
	\centering
	\begin{tabular}{ccccc}
		\includegraphics[width=0.18\linewidth]{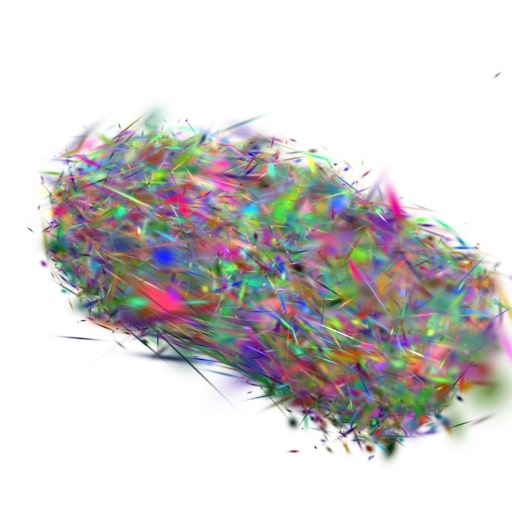} &
		\includegraphics[width=0.18\linewidth]{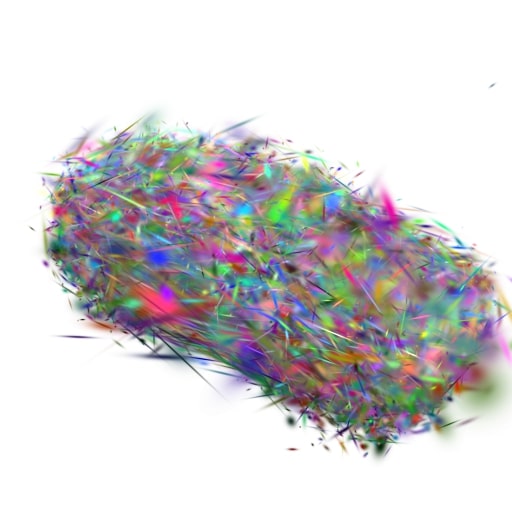} &
		\includegraphics[width=0.18\linewidth]{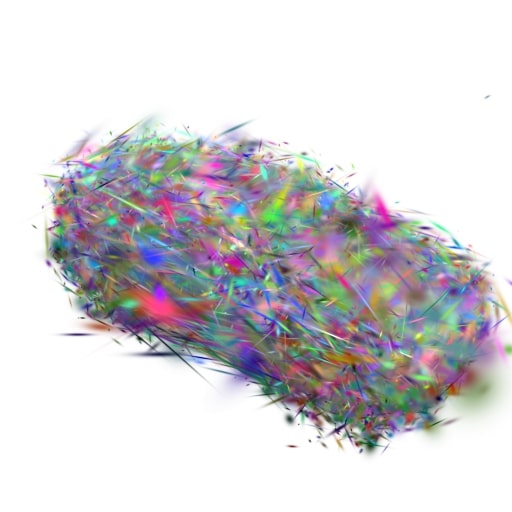} &
        \includegraphics[width=0.18\linewidth]{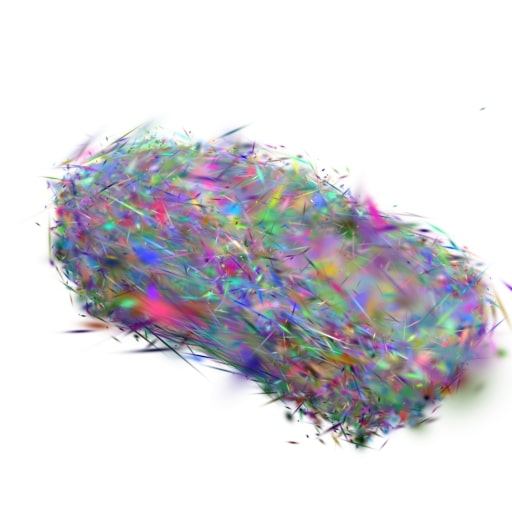} &
		\includegraphics[width=0.18\linewidth]{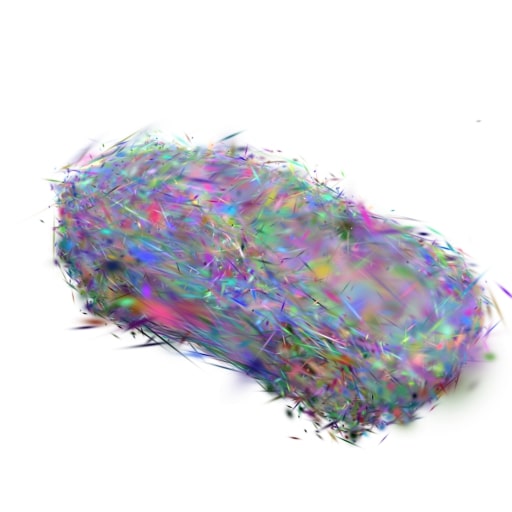} \\
		\includegraphics[width=0.18\linewidth]{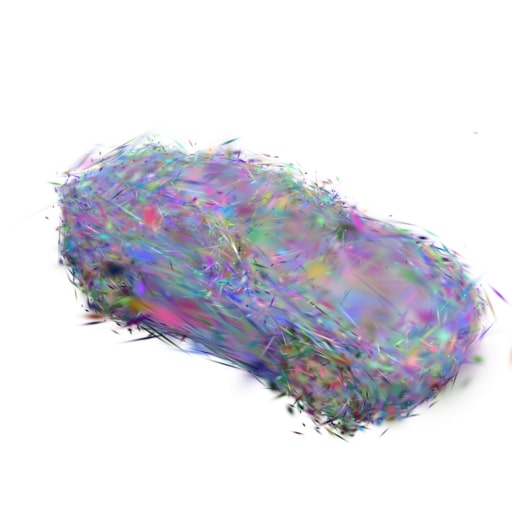} &
        \includegraphics[width=0.18\linewidth]{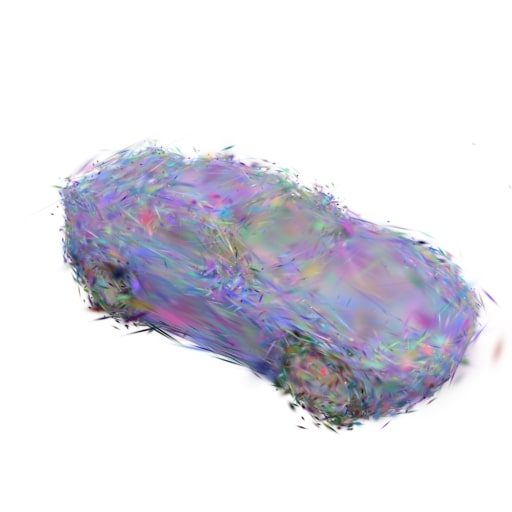} &
        \includegraphics[width=0.18\linewidth]{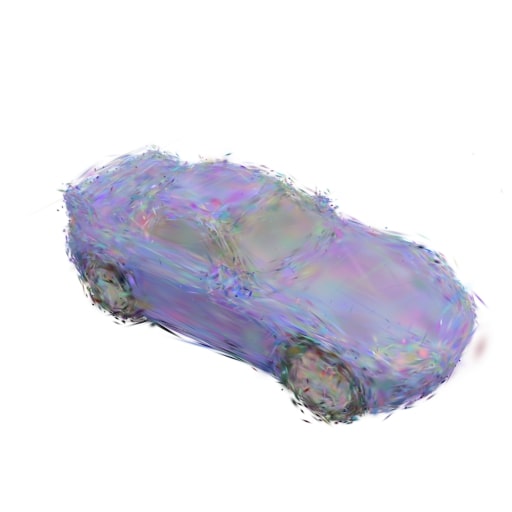} &
        \includegraphics[width=0.18\linewidth]{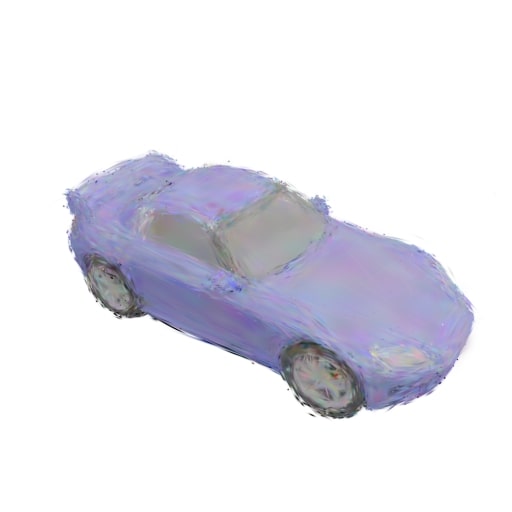} &
        \includegraphics[width=0.18\linewidth]{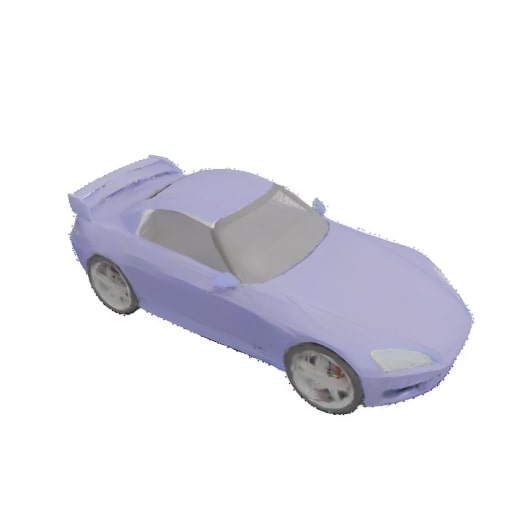}
	\end{tabular}
	\caption{Visualization of generation results in intermediate diffusion timesteps.}
	\vspace{-3mm}
	\label{fig/supp:intermediate}
\end{figure*}

\noindent\textbf{Visualization of intermediate results in the denoising process.} During inference, our model starts from Gaussian noise and progressively denoises to yield the high-quality GaussianCube. We present visualizations of the intermediate renderings $\bm{y}_t$ at various timesteps $t \in [0, T]$ throughout the denoising process, offering a detailed insight into the GaussianCube diffusion procedure. As illustrated in~\Cref{fig/supp:intermediate}, our model first establishes the global structure and then incrementally enhances the details, which is similar to previous 3D diffusion models~\cite{wang2023rodin,shue20233d}.

\begin{table*}[t]  
	\centering 
	\small
        \caption{Details of each dataset.} 
        \vspace{1mm}
	\begin{tabular}{cccccc}  
		\hline  
		\textbf{Dataset} & \textbf{\# Objects} & \textbf{\# Views per object} & \textbf{Rotation Angle} & \textbf{Elevation Angle} & \textbf{Bounding Box} \\  
		\hline  
	  ShapeNet Car & 7,462 & 150 & $[0, 2\pi]$ & $[\frac{1}{6}\pi, \frac{1}{2}\pi]$ & 0.45 \\
        ShapeNet Chair & 6,775 & 150 & $[0, 2\pi]$ & $[\frac{1}{6}\pi, \frac{1}{2}\pi]$ & 0.35\\
        OmniObject3D & 5,795 & 100 & $[0, 2\pi]$ & $[0, \frac{1}{2}\pi]$ & 1.0 \\
        Synthetic Avatar & 98,000 & 300 & $[0, 2\pi]$ & $[\frac{1}{6}\pi, \frac{2}{3}\pi]$ & 40.0 \\
        Objaverse & 125,653 & 150 & $[0, 2\pi]$ & $[0, \frac{2}{3}\pi]$ & 0.5 \\
		\hline  
	\end{tabular}  
	\label{supp/tab:dataset_details}  
        \vspace{-5mm}
\end{table*}  

\begin{table*}[t]  
	\centering 
	\small
        \caption{Detailed configuration of model architecture, diffusion training and inference on each dataset.} 
        \vspace{1mm}
        \setlength\tabcolsep{3pt}
	\begin{tabular}{cccccc}  
		\hline  
		\textbf{} & \textbf{ShapeNet Car} & \textbf{ShapeNet Car} & \textbf{OmniObject3D} & \textbf{Synthetic Avatar} & \textbf{Objaverse} \\  
		\hline  
	  Diffusion Steps & 1,000 & 1,000 & 1,000 & 1,000 & 1,000 \\
        Noise Schedule & Cosine & Cosine & Cosine & Cosine & Cosine \\
        NFEs & 300 & 300 & 300 & 250 & 44 \\
        Inference Time (s) & 10.06 & 10.06 & 10.06 & 13.80 & 2.30 \\
        Inference Sampler & DPM-solver~\cite{lu2022dpm} & DPM-solver~\cite{lu2022dpm} & DPM-solver~\cite{lu2022dpm} & DPM-solver~\cite{lu2022dpm} & DPM-solver~\cite{lu2022dpm} \\
        DPM-solver Order & 3 & 3 & 3 & 2 & 2 \\
        DPM-solver Mode & Multi-step & Multi-step & Multi-step & Multi-step & Adaptive \\
        CFG Scale & - & - & 2.0 & 1.3 & 3.5 \\
        Model Size & 82M & 82M & 82M & 339M & 339M \\
        Channels & 64 & 64 & 64 & 128 & 128 \\
        Channel Mult. & (1,2,3,4) & (1,2,3,4) & (1,2,3,4) & (1,2,3,4) & (1,2,3,4) \\
        Num. Res. Blocks & 3 & 3 & 3 & 3 & 3 \\ 
        Attn Resolutions & (8, 4) & (8, 4) & (8, 4) & (8, 4) & (8, 4) \\
        Num. Head Channels & 64 & 64 & 64 & 64 & 64 \\
        Dropout & 0 & 0 & 0 & 0 & 0 \\
        Scale Shift Norm & True & True & True & True & True \\
        Training Steps & 1,000K & 1,000K & 700K & 1,200K & 1,800K \\
        Training GPUs & 16 & 16 & 16 & 16 & 32 \\
        Batch Size & 128 & 128 & 128 & 128 & 256 \\
        Base Lr & $5e-5$ & $5e-5$ & $5e-5$ & $5e-5$ & $5e-5$ \\
        Lr Decay Steps & 850K & 850K & - & - & - \\
		\hline  
	\end{tabular}  
	\label{supp/tab:training_details}  
        \vspace{-5mm}
\end{table*}  

\begin{table}[t]  
	\centering  
	\small
        \caption{Quantitative ablation of $N_{\text{max}}$ in densification-constrained fitting. We set $N_{\text{max}}$ to 32,768 in this paper.}  
	\begin{tabular}{ccccc}  
		\hline  
		\textbf{$N_{\text{max}}$} & \textbf{$N_v$} & \textbf{PSNR$\uparrow$} & \textbf{LPIPS$\downarrow$} & \textbf{SSIM$\uparrow$} \\  
		\hline   
		4096  & 16 & 32.56 & 0.0547 & 0.9765 \\
		13824 & 24 & 34.32 & 0.0396 &  0.9842\\
		\cellcolor{gray}32768 & \cellcolor{gray}32 & \cellcolor{gray}34.94 & \cellcolor{gray}0.0347 &  \cellcolor{gray}0.9863 \\ 
		110592 & 48 & 35.29 & 0.0307 & 0.9874 \\ 
		262144 & 64 & 35.34 & 0.0301 & 0.9875\\ 
		\hline  
	\end{tabular}  
	\vspace{-3mm}
	\label{supp/tab:ablation_fitting}  
\end{table} 

\begin{table*}[t]  
	\centering 
        \caption{Quantitative ablation of CFG scale in the class-conditioned generation of OmniObject3D~\cite{wu2023omniobject3d}.} 
	\begin{tabular}{ccccccc}  
		\hline  
		\textbf{Scale} & w/o CFG & 1.3 & 1.5 & 2.0 & 3.0 & 6.0\\  
		\hline  
		\textbf{FID-50K$\downarrow$} & 13.39 & 12.07 & 11.72 & \textbf{11.62} & 12.99 & 32.80 \\  
	  \textbf{KID-50K(\textperthousand)$\downarrow$} & 4.01 & 3.12 & 3.00 & \textbf{2.78} & 3.17 & 14.36\\   
		\hline  
	\end{tabular}  
	 
	\label{supp/tab:cfg_scale}  
        \vspace{-3mm}
\end{table*}  

\begin{figure*}[h]
	\small
	\centering
	\begin{tabular}{c}
		\includegraphics[width=0.997\linewidth]{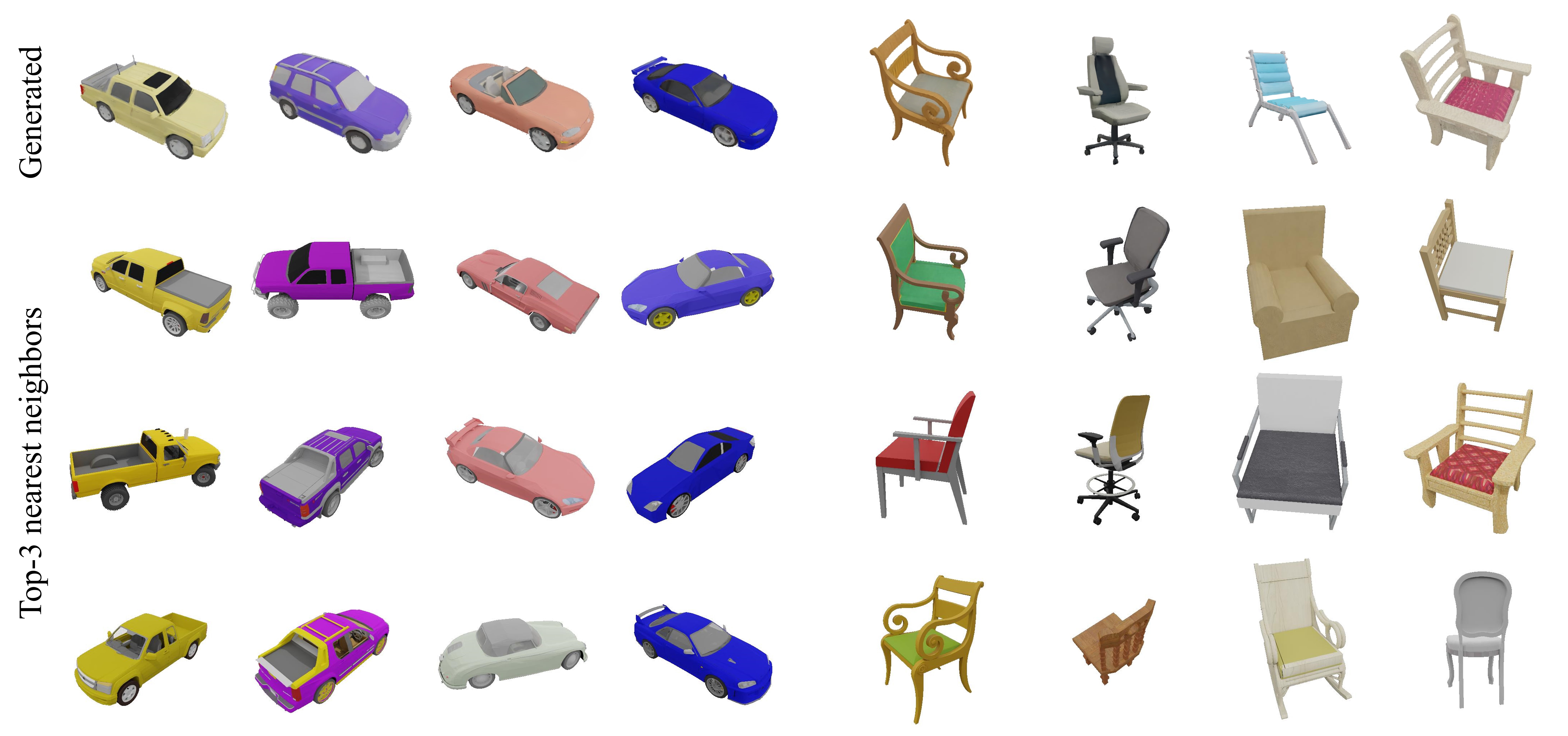}
	\end{tabular}
	\caption{Visualization of nearest neighbor search on ShapeNet Car and Chair.}
	\vspace{-3mm}
	\label{fig/supp:nearest_neighbors}
\end{figure*}

\noindent\textbf{Nearest neighbors analysis.} We perform nearest neighbor search of some unconditionally generated samples in the paper according to the similarity of pretrained CLIP~\cite{radford2021learning} features. The results in~\Cref{fig/supp:nearest_neighbors} demonstrate that our model is capable of generating novel geometry and textures rather than simply memorizing the training data.

\begin{figure*}[h!]
	\small
	\centering
    \renewcommand{\arraystretch}{0.5}
	\setlength\tabcolsep{1pt}
	\begin{tabular}{ccccc}
		\includegraphics[width=0.2\linewidth]{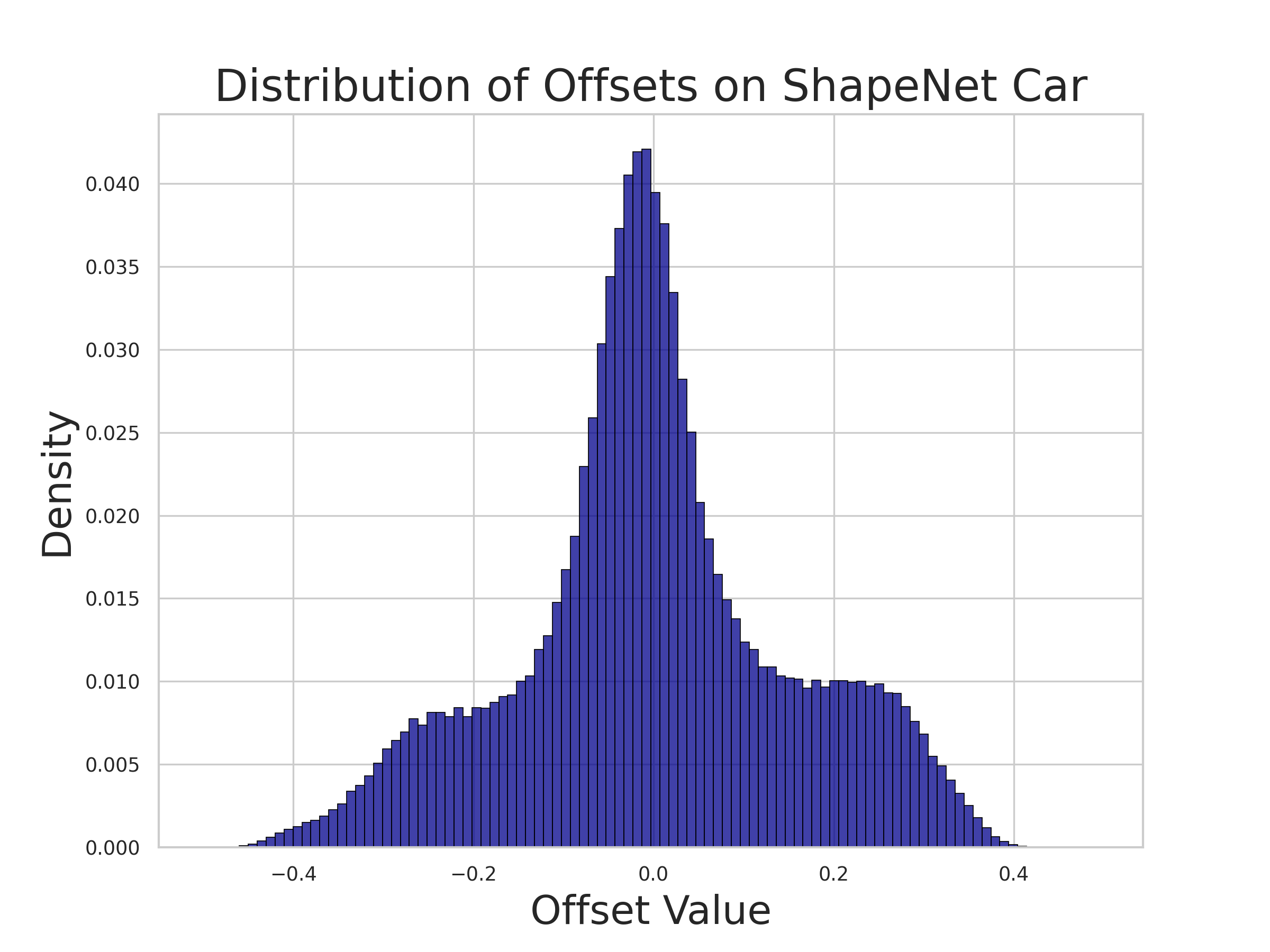} &
		\includegraphics[width=0.2\linewidth]{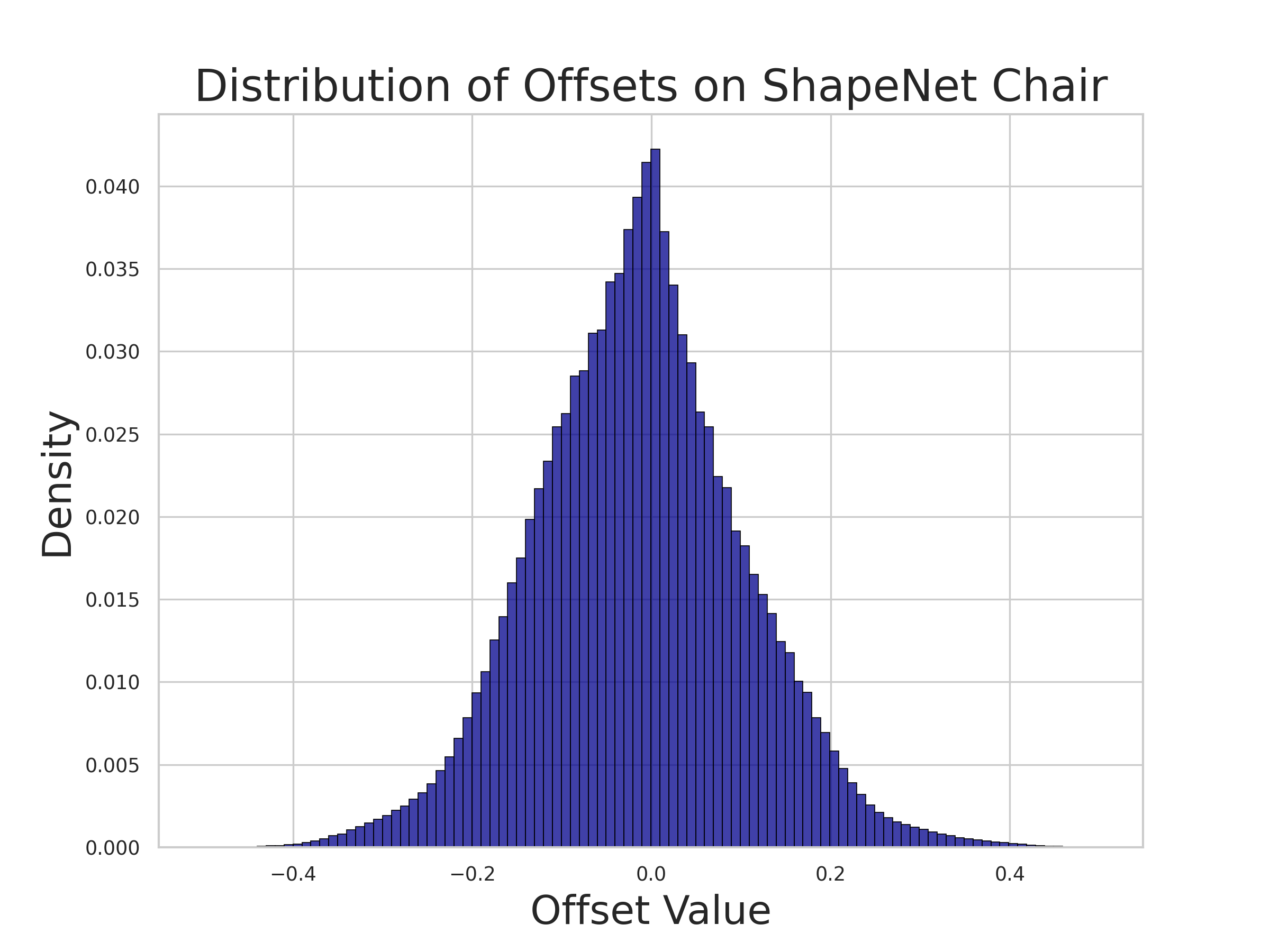} &
		\includegraphics[width=0.2\linewidth]{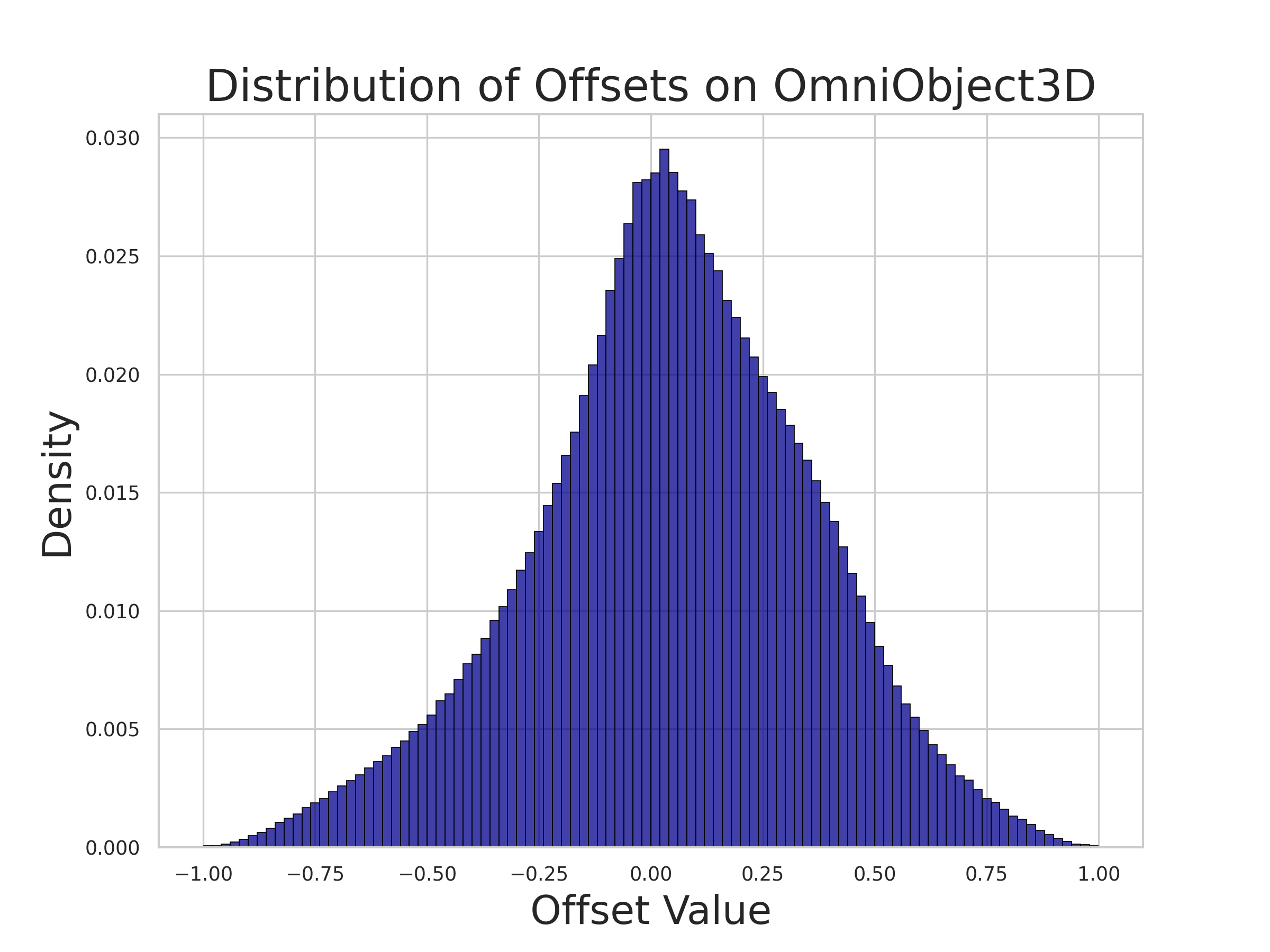} &
        \includegraphics[width=0.2\linewidth]{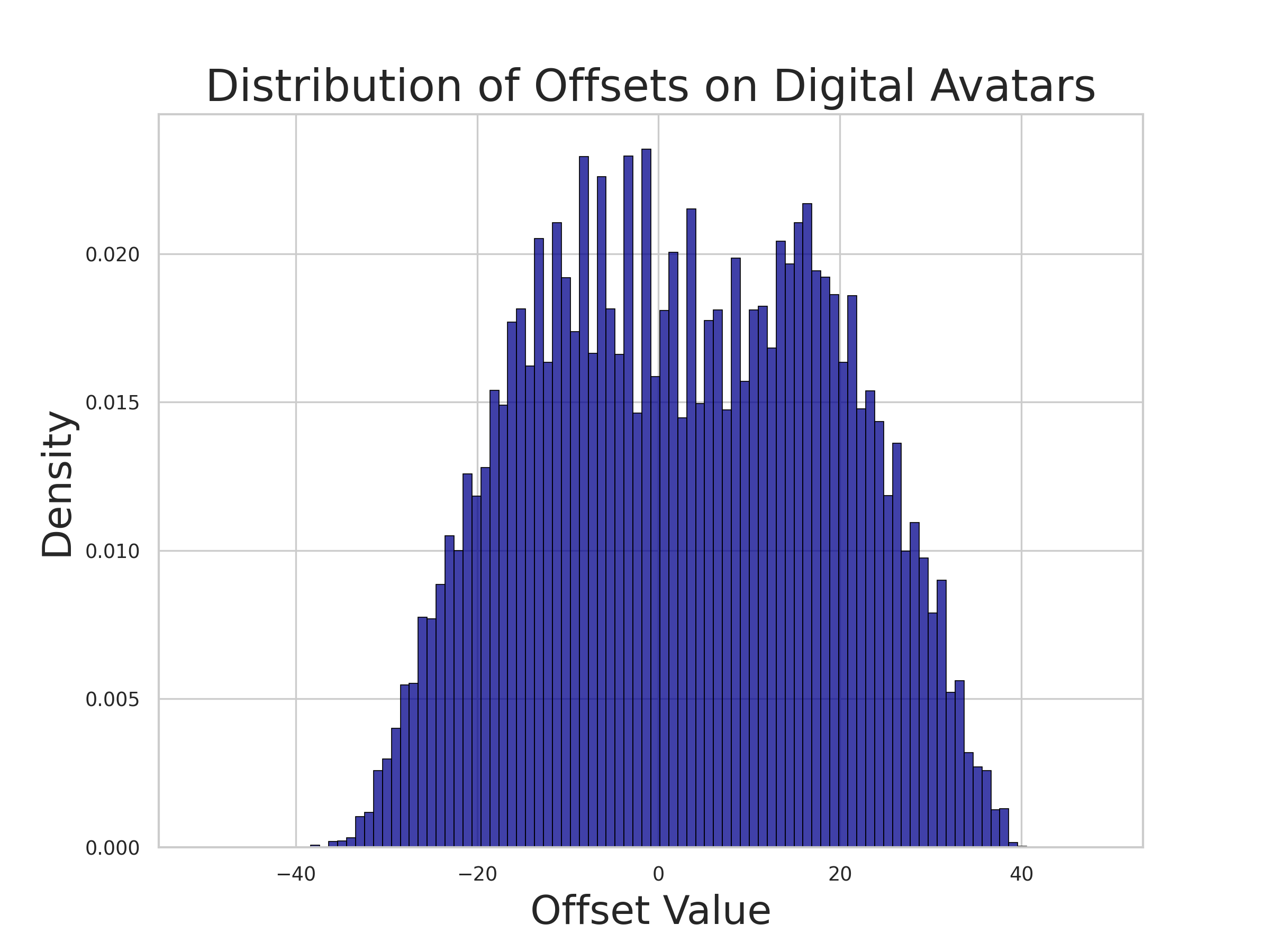} &
        \includegraphics[width=0.2\linewidth]{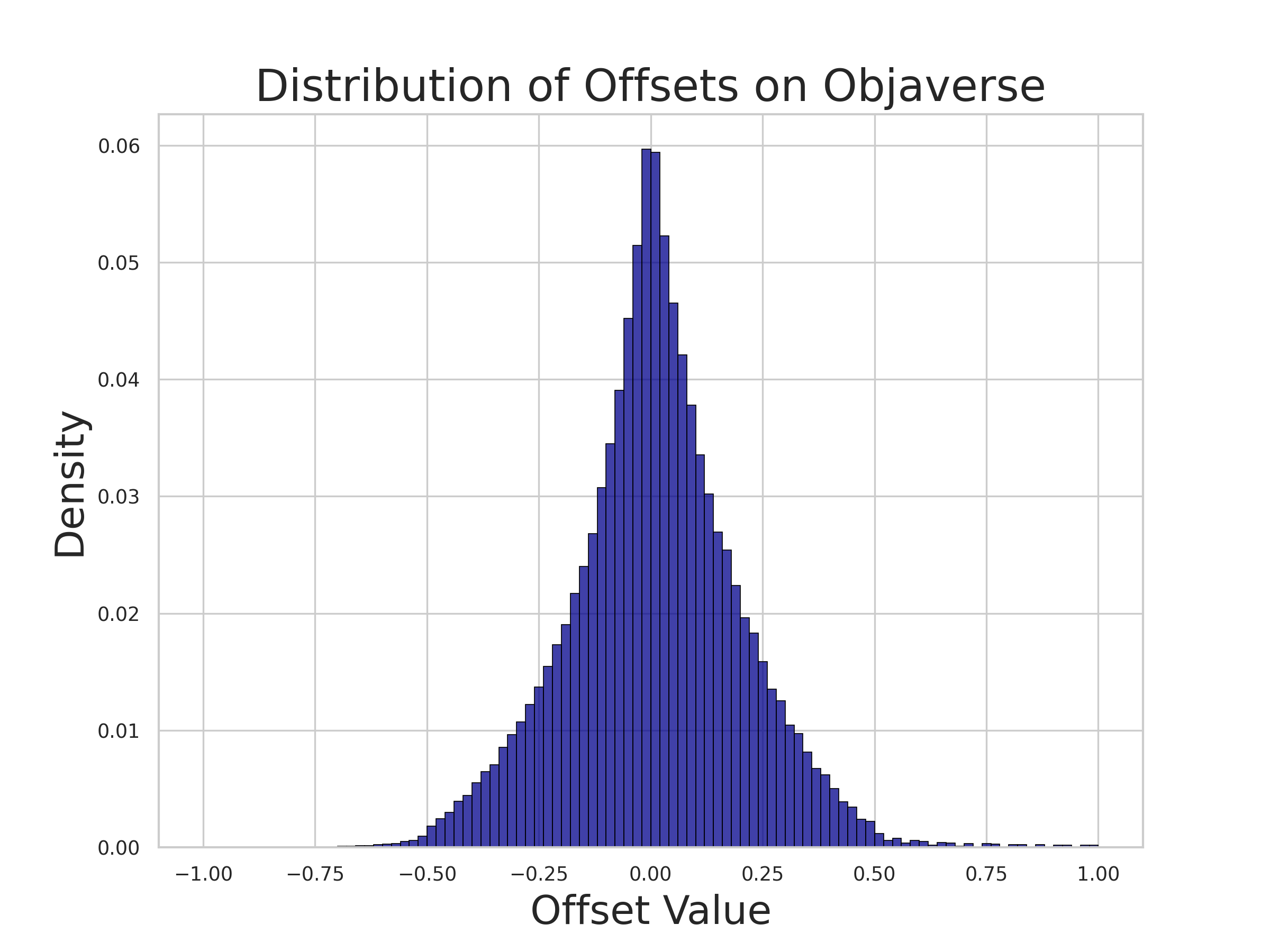}
	\end{tabular}
        \vspace{-3mm}
	\caption{Distribution of offsets from voxel centers in a random selection of 1K GaussianCubes on each experimental dataset.}
	\vspace{-3mm}
	\label{supp/fig:offset_vis}
\end{figure*}

\noindent\textbf{Distribution visualization of offset from voxel grids of fitted GaussianCubes.} We visualize the offset distribution of 1K randomly selected GaussianCubes from each experimental dataset in~\Cref{supp/fig:offset_vis}. We observe that most distributions exhibit a bell curve, similar to a normal distribution. However, the Digital Synthetic Avatar dataset presents a more uniform distribution with multiple peaks. We believe these distributions offer valuable insights into how well the fitted 3D Gaussians align with voxel grid centers. Bell-shaped distributions akin to a normal distribution, such as in the ShapeNet Car and Chair datasets, suggest a strong initial alignment and lower complexity. On the other hand, broader distributions (e.g., the Digital Synthetic Avatar dataset) indicate a higher level of detail (for instance, hair) and a greater need for adjustments during organization.

\subsection{Additional Visual Results}

\begin{figure*}[t]
    \small
    \centering
    \begin{tabular}{c@{\hspace{3mm}}c@{\hspace{1mm}}c@{\hspace{3mm}}c@{\hspace{1mm}}c} 
    \includegraphics[width=0.18\linewidth]{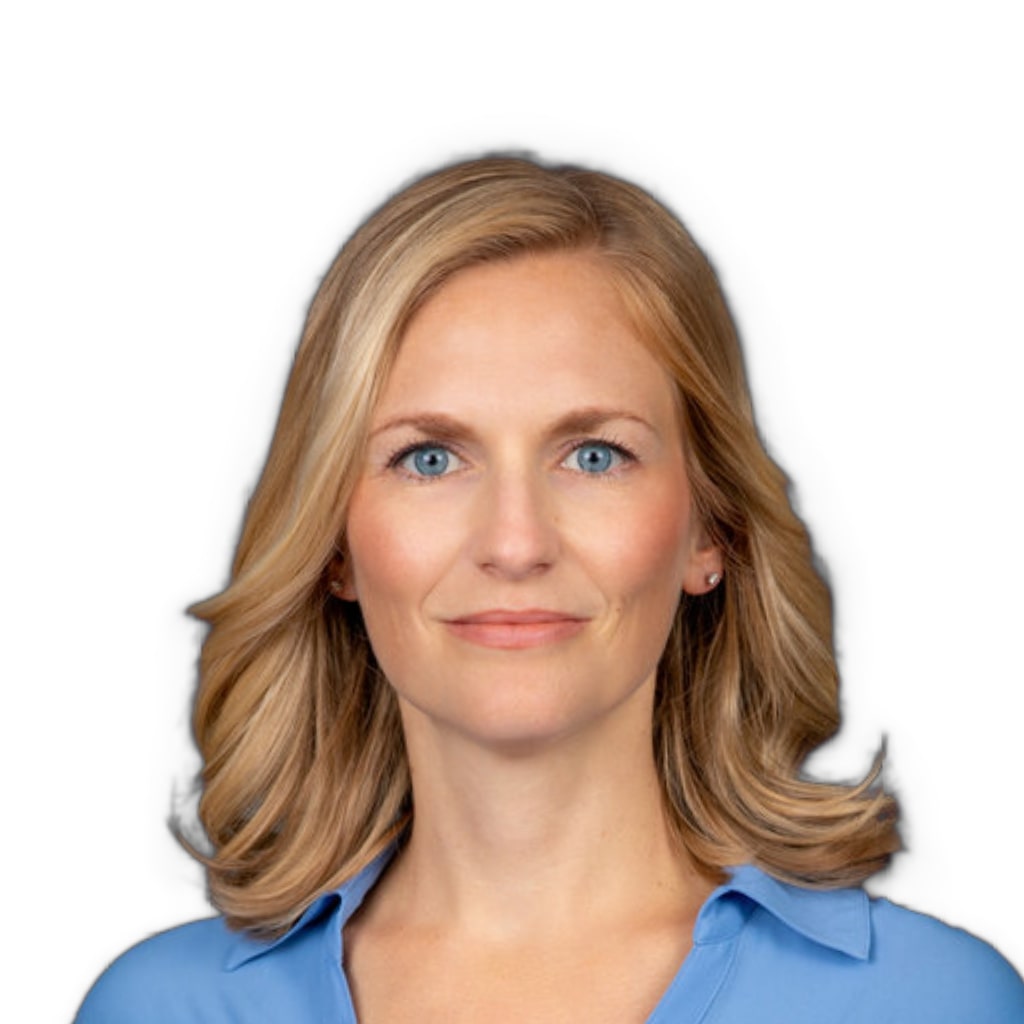}  &
    \includegraphics[width=0.18\linewidth]{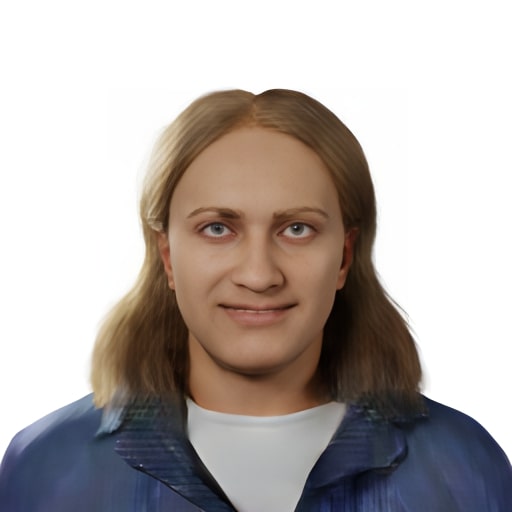} &
    \includegraphics[width=0.18\linewidth]{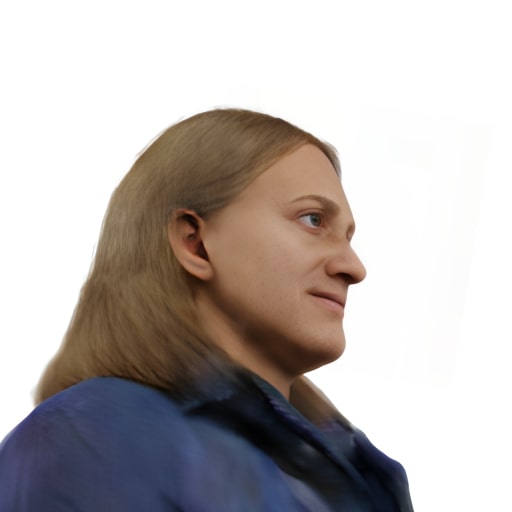} &
    \includegraphics[width=0.18\linewidth]{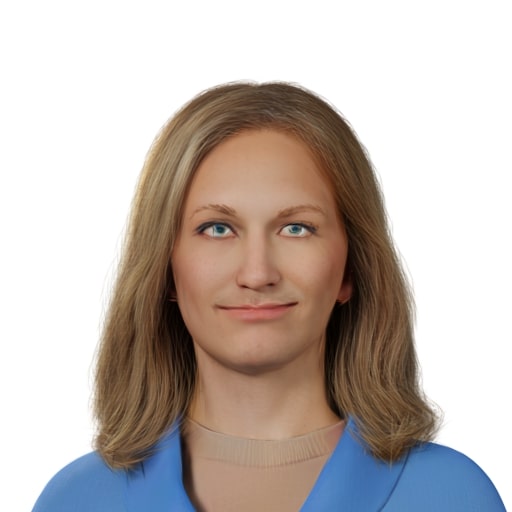} &
    \includegraphics[width=0.18\linewidth]{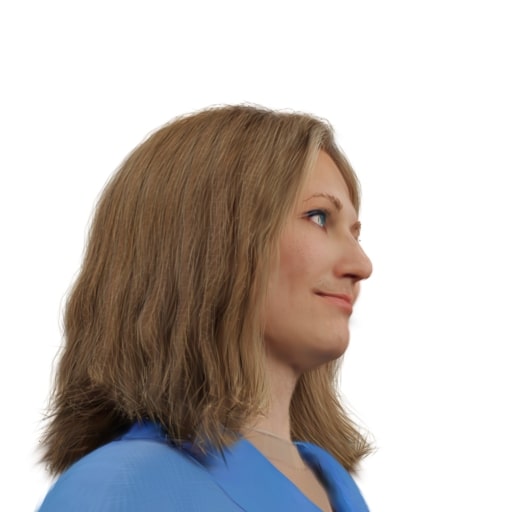} \\
    \includegraphics[width=0.18\linewidth]{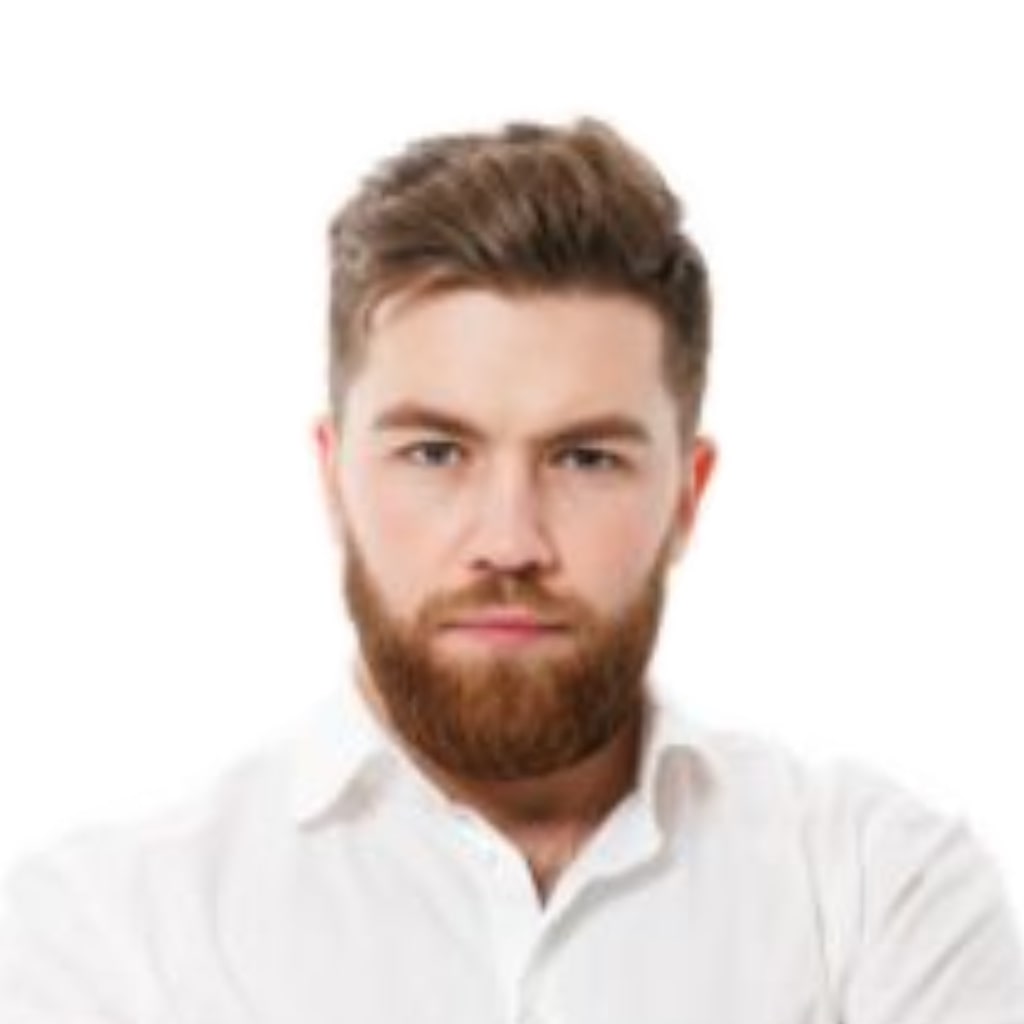}  &
    \includegraphics[width=0.18\linewidth]{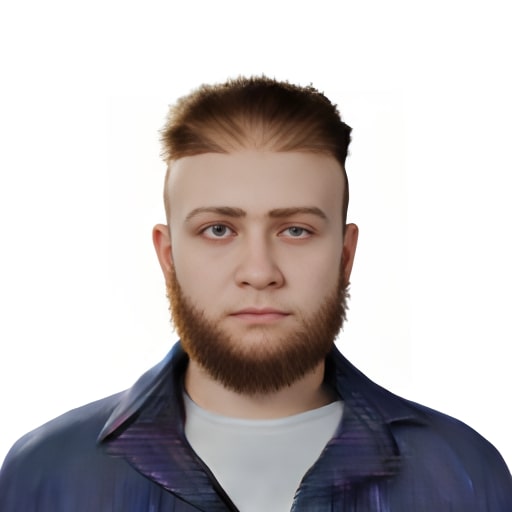} &
    \includegraphics[width=0.18\linewidth]{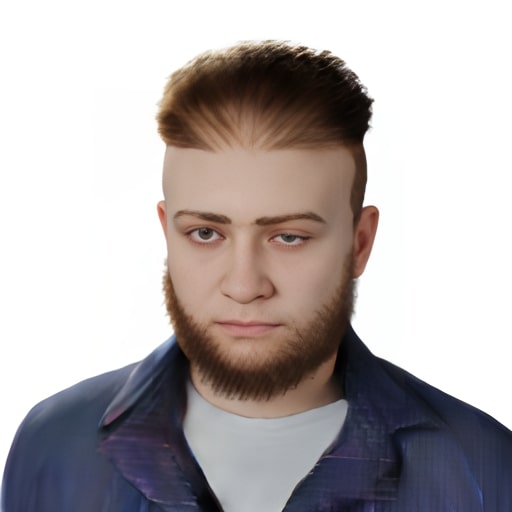} &
    \includegraphics[width=0.18\linewidth]{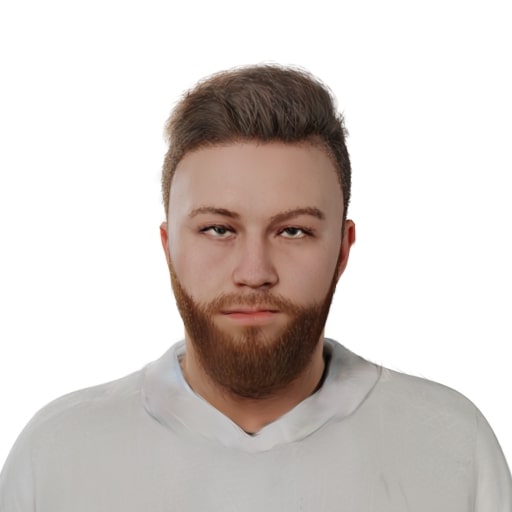} &
    \includegraphics[width=0.18\linewidth]{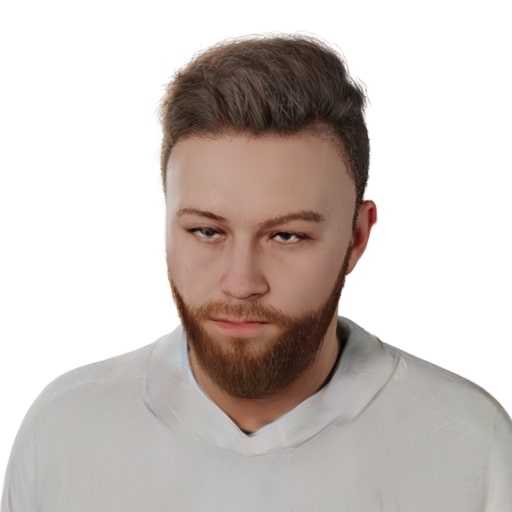} \\
    \includegraphics[width=0.18\linewidth]{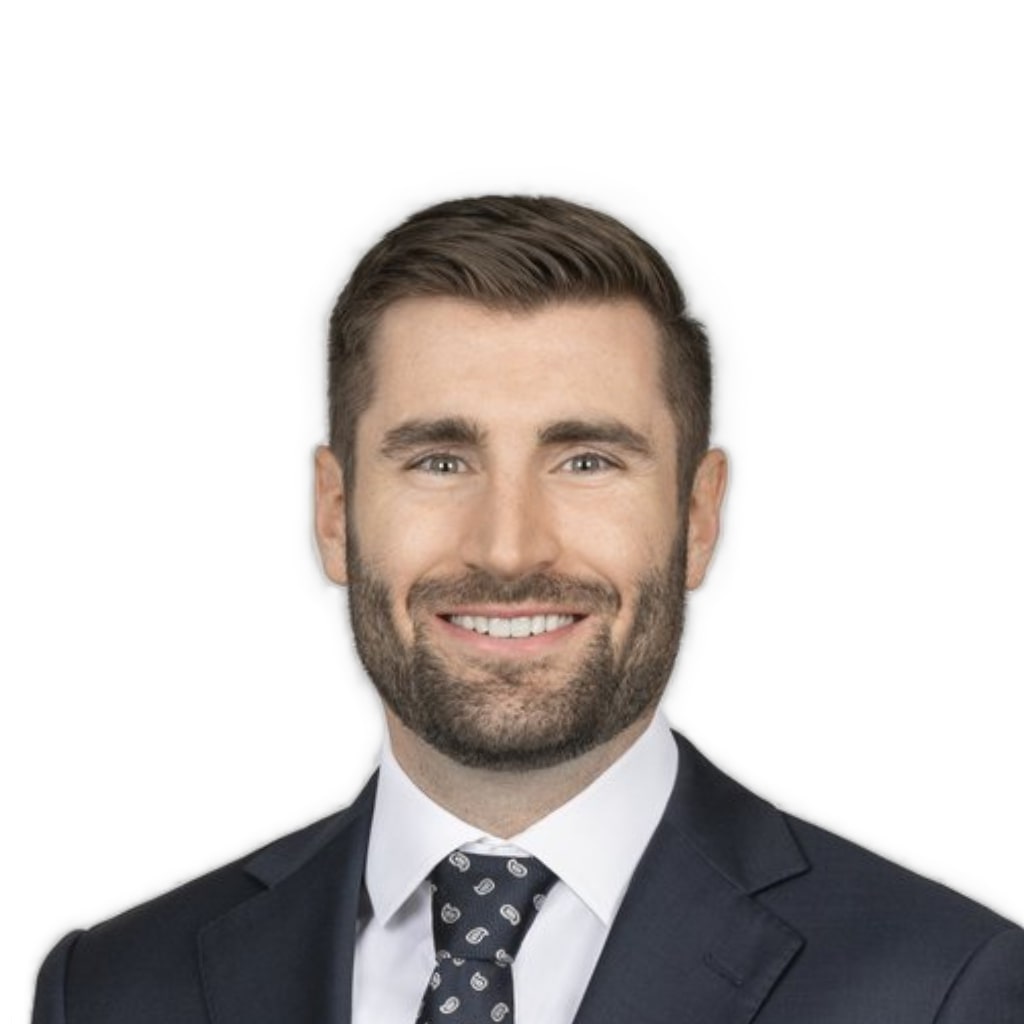}  &
    \includegraphics[width=0.18\linewidth]{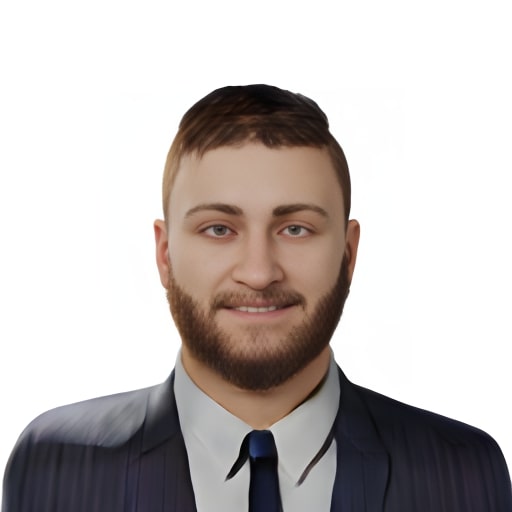} &
    \includegraphics[width=0.18\linewidth]{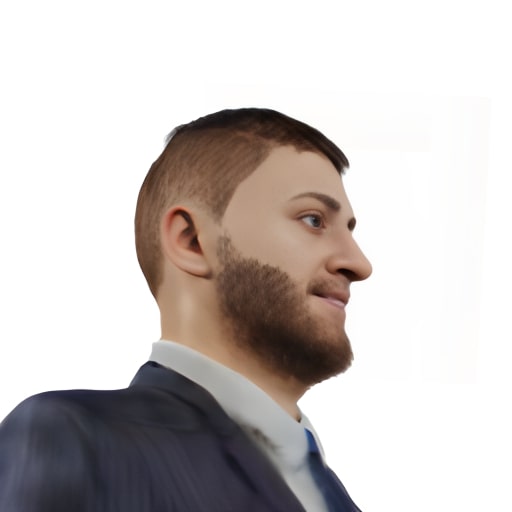} &
    \includegraphics[width=0.18\linewidth]{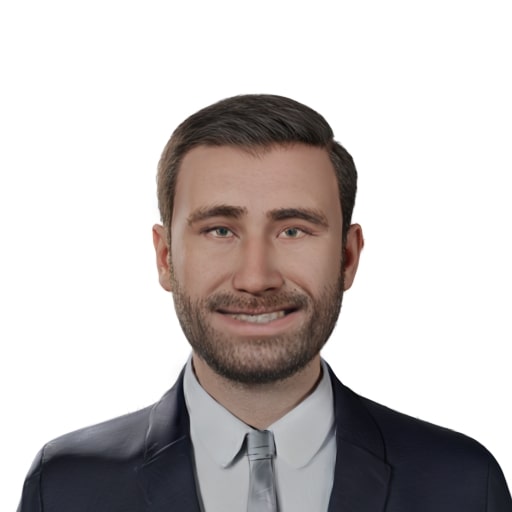} &
    \includegraphics[width=0.18\linewidth]{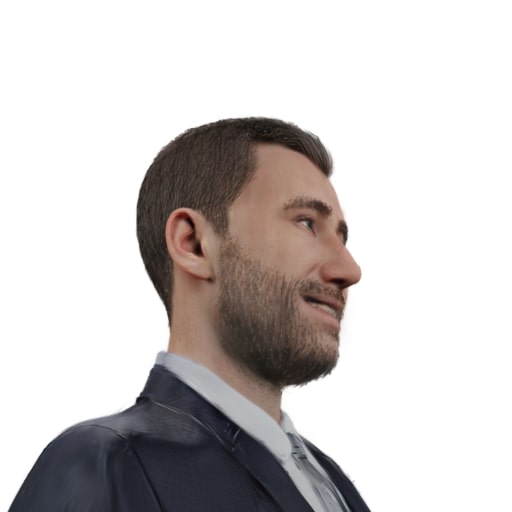} \\
    \includegraphics[width=0.18\linewidth]{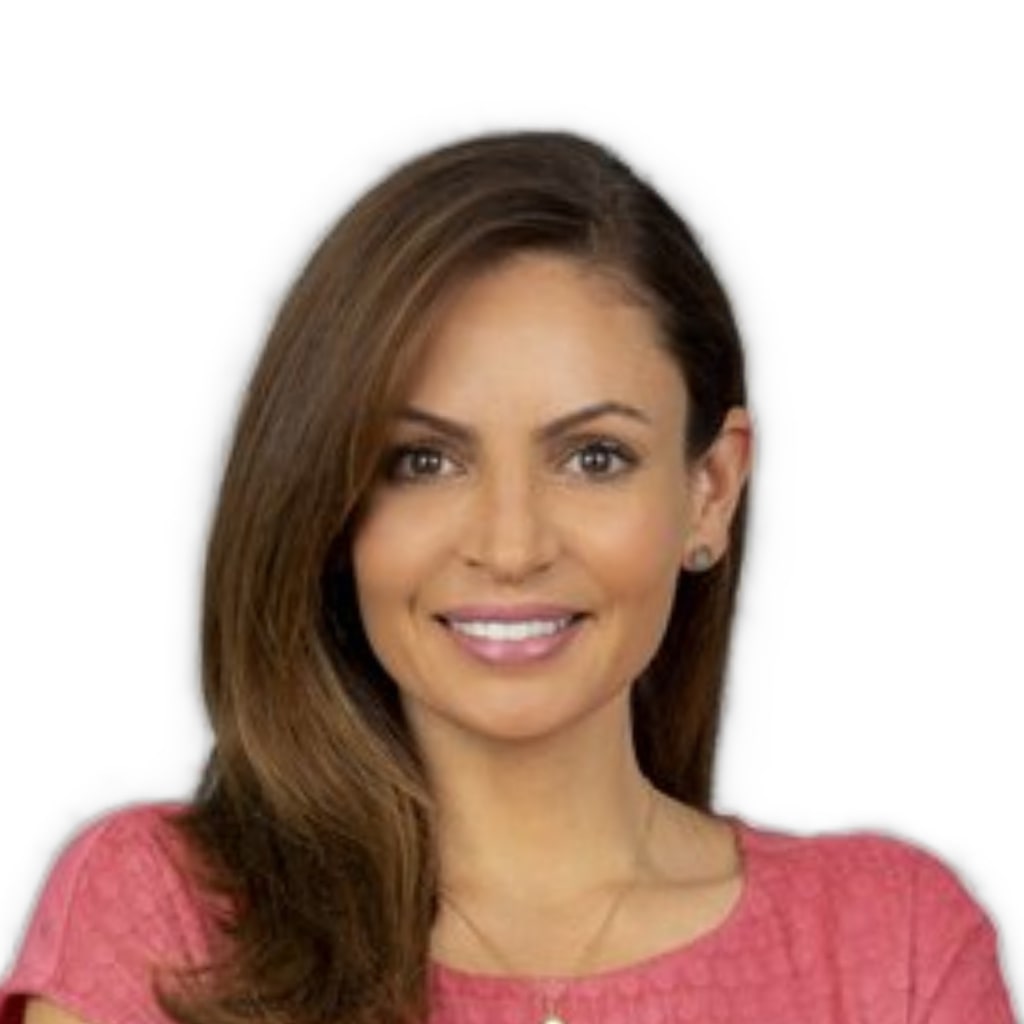}  &
    \includegraphics[width=0.18\linewidth]{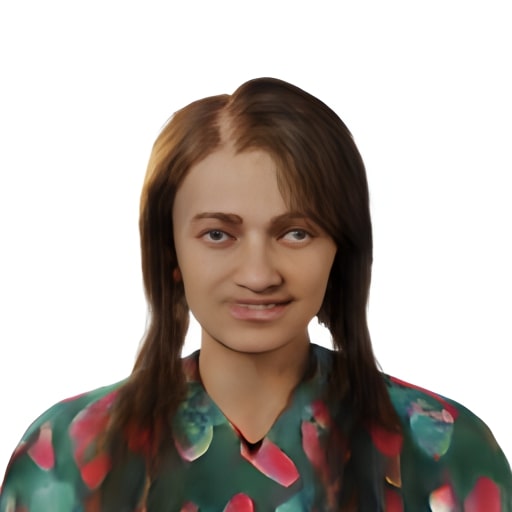} &
    \includegraphics[width=0.18\linewidth]{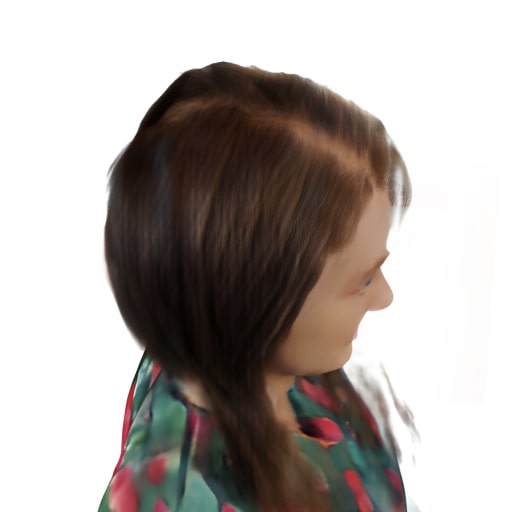} &
    \includegraphics[width=0.18\linewidth]{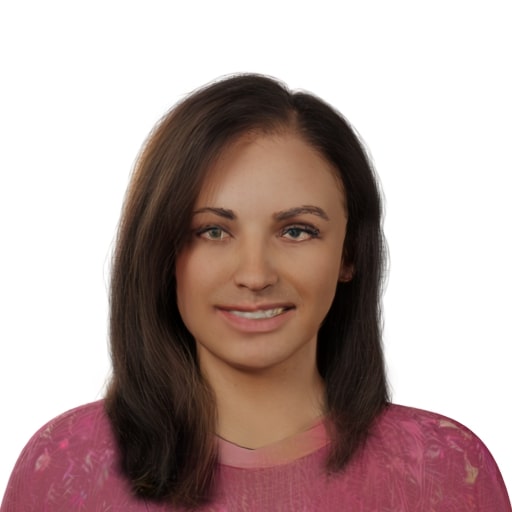} &
    \includegraphics[width=0.18\linewidth]{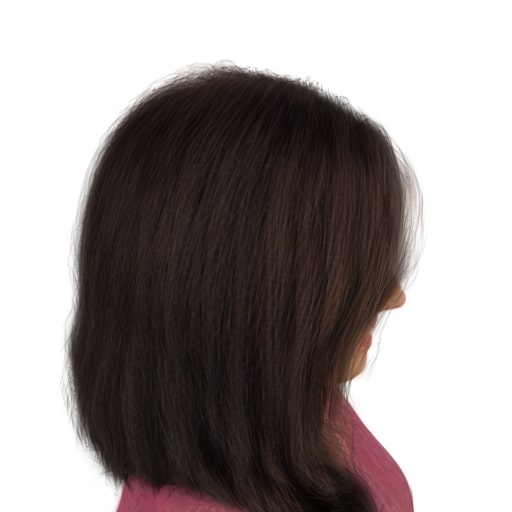} \\
    Reference & \multicolumn{2}{c}{Rodin~\cite{wang2023rodin}} & \multicolumn{2}{c}{\textbf{Ours}}\\
  \end{tabular}
  \vspace{-1mm}
  \caption{Additional qualitative comparison of 3D avatars creation conditioned on single in-the-wild portraits.}
  \label{supp/fig:real_world_avatars}
\end{figure*}

\begin{figure*}[t]
	\small
	\centering
         \begin{overpic}
            [width=0.997\linewidth]{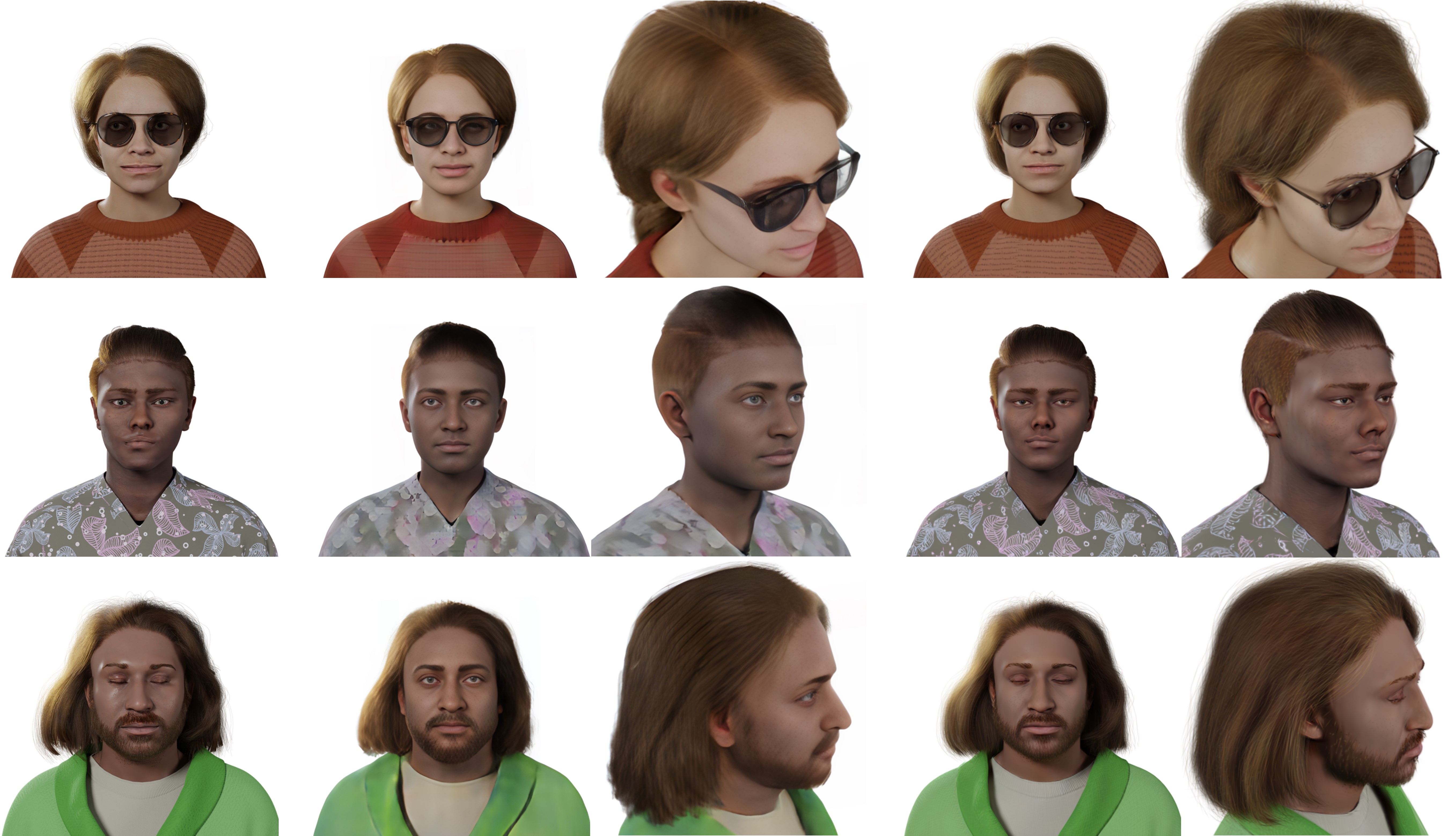}
            \put(5,-3){Reference}
            \put(37,-3){Rodin~\cite{wang2023rodin}}
    	\put(79,-3){\textbf{Ours}}
          \end{overpic}
	\caption{Qualitative comparison generated digital avatars conditioned on synthetic portraits.}
	\vspace{-1mm}
	\label{supp/fig:synthetic_avatars}
\end{figure*}

For 3D avatar generation, while trained on synthetic dataset, our model is capable of generalizing to in-the-wild portrait input. We provide more visual comparison of 3D avatar creation conditioned on in-the-wild portraits with Rodin~\cite{wang2023rodin} in~\Cref{supp/fig:real_world_avatars}. We also include additional comparison conditioned on synthetic input from our test in~\Cref{supp/fig:synthetic_avatars}. Our model can faithfully retain the identity of the reference portrait and is able to provide high-fidelity results with rich details, \textit{e}.\textit{g}. hair, glasses and clothing. Although utilizing a pretrained 2D super-resolution module which significantly compromises 3D consistency, Rodin struggles to follow the conditional images and fails to produce detailed textures in non-facial areas \textit{e}.\textit{g}. clothing and hair.

\begin{figure*}[t]
	\centering
	\small
	\begin{tabular}{c}
            \includegraphics[width=0.85\linewidth]{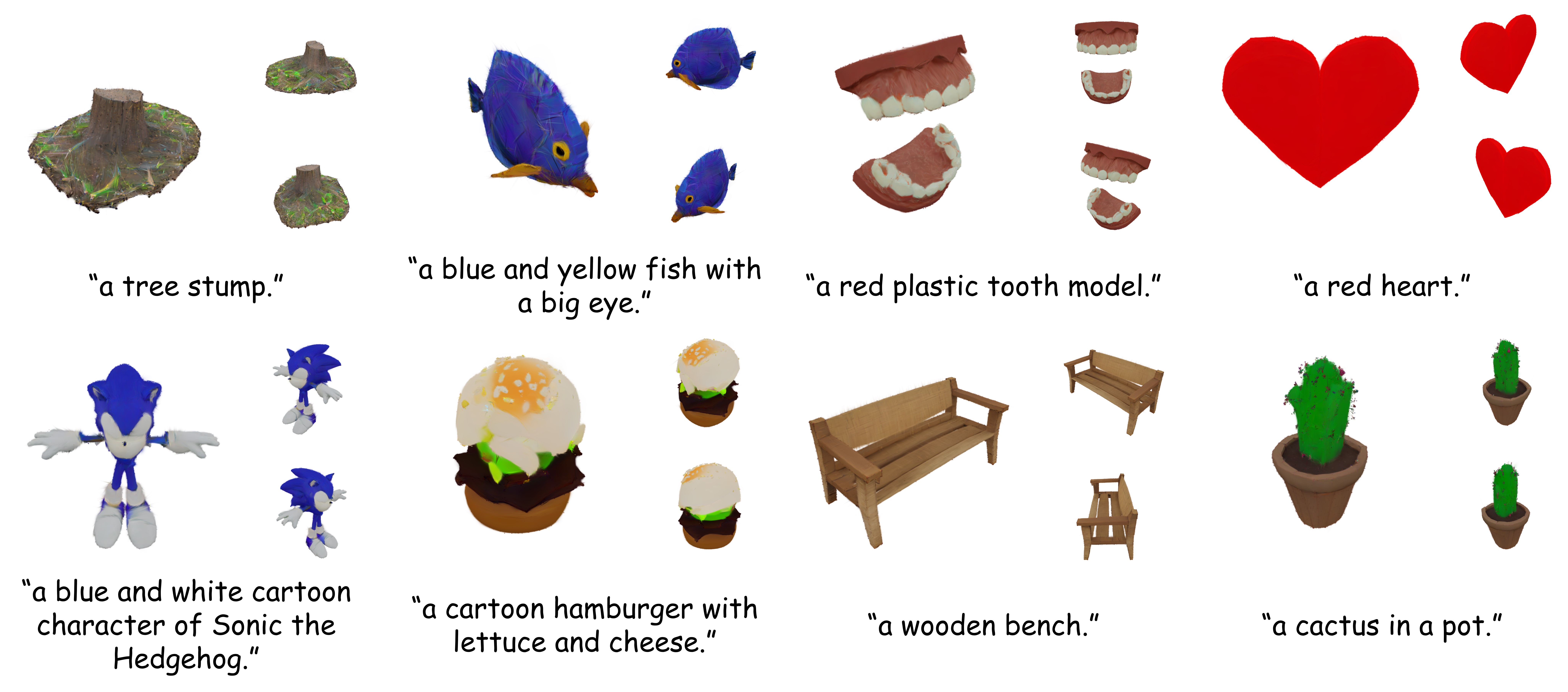}\\
	\end{tabular}
	\vspace{-3mm}
	\caption{Additional results of text-to-3D generation.}
	\vspace{-5mm}
	\label{supp/fig:textcond_gen_additional}
\end{figure*}

\begin{figure*}[t]
	\centering
	\small
	\begin{overpic}[width=0.85\linewidth]{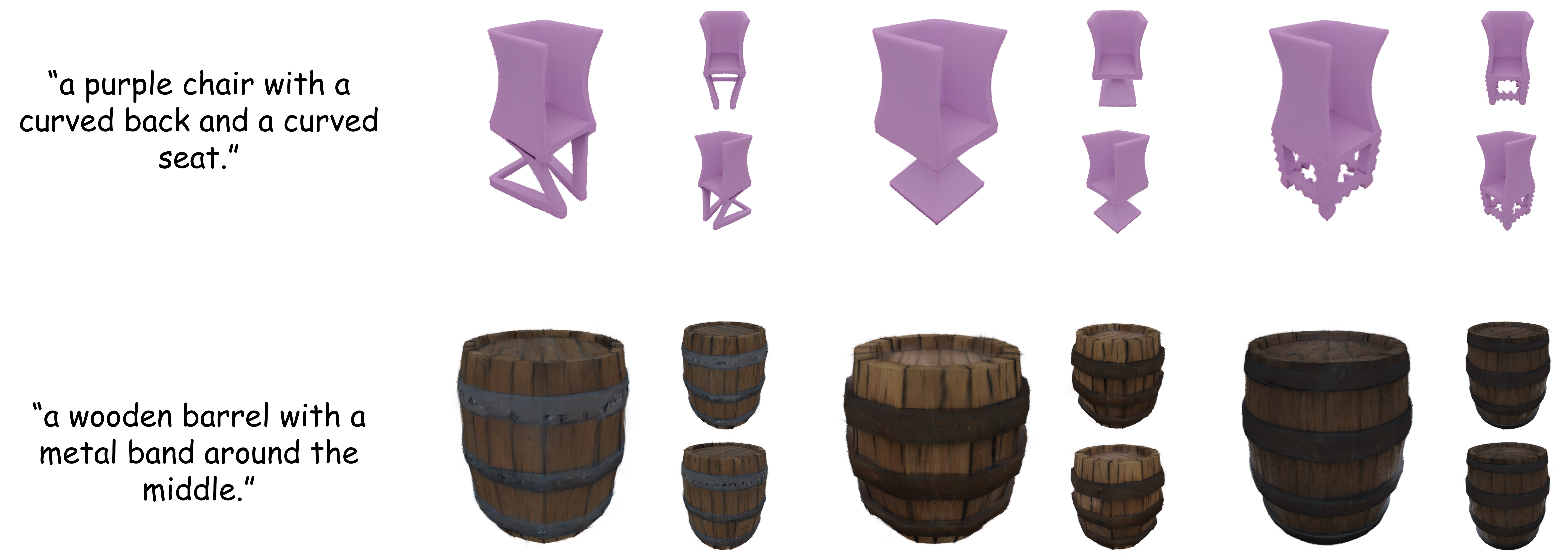}
		\put(5,-2.5){Text Condition}
            \put(37, -2.5){Sample 1}
            \put(60, -2.5){Sample 2}
		\put(85, -2.5){Sample 3}
	\end{overpic}
	\vspace{1mm}
	\caption{Variation of text-to-3D generation. Our model is able to generate diverse results conditioned on the same text prompt.}
	\vspace{-5mm}
	\label{supp/fig:textcond_gen_var}
\end{figure*}

\begin{figure*}[t]
	\centering
	\small
	\begin{overpic}[width=0.85\linewidth]{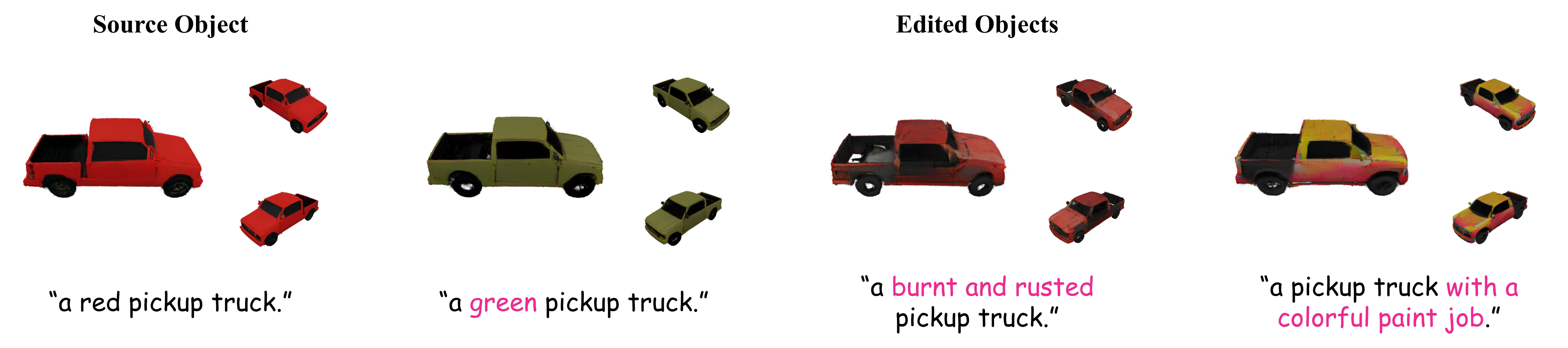}
	\end{overpic}
	\caption{Example of text-guided 3D editing.}
	\vspace{-5mm}
	\label{supp/fig:textcond_gen_edit}
\end{figure*}

We include additional qualitative comparison and generated samples of text-to-3D generation in~\Cref{supp/fig:textcond_gen} and~\Cref{supp/fig:textcond_gen_additional} respectively. Our model yields samples with better visual quality, and is capable of handling challenging prompts. The results in~\Cref{supp/fig:textcond_gen_var} show the generation diversity of our results given the same text prompt. Our model is also capable of performing text-guided editing of generated objects by leveraging SDEdit~\cite{meng2021sdedit} as depicted in~\Cref{supp/fig:textcond_gen_edit}, demonstrating the promise of achieving controllable 3D generation.

\begin{figure*}[t]
	\small
	\centering
	\begin{tabular}{c}
		\includegraphics[width=0.997\linewidth]{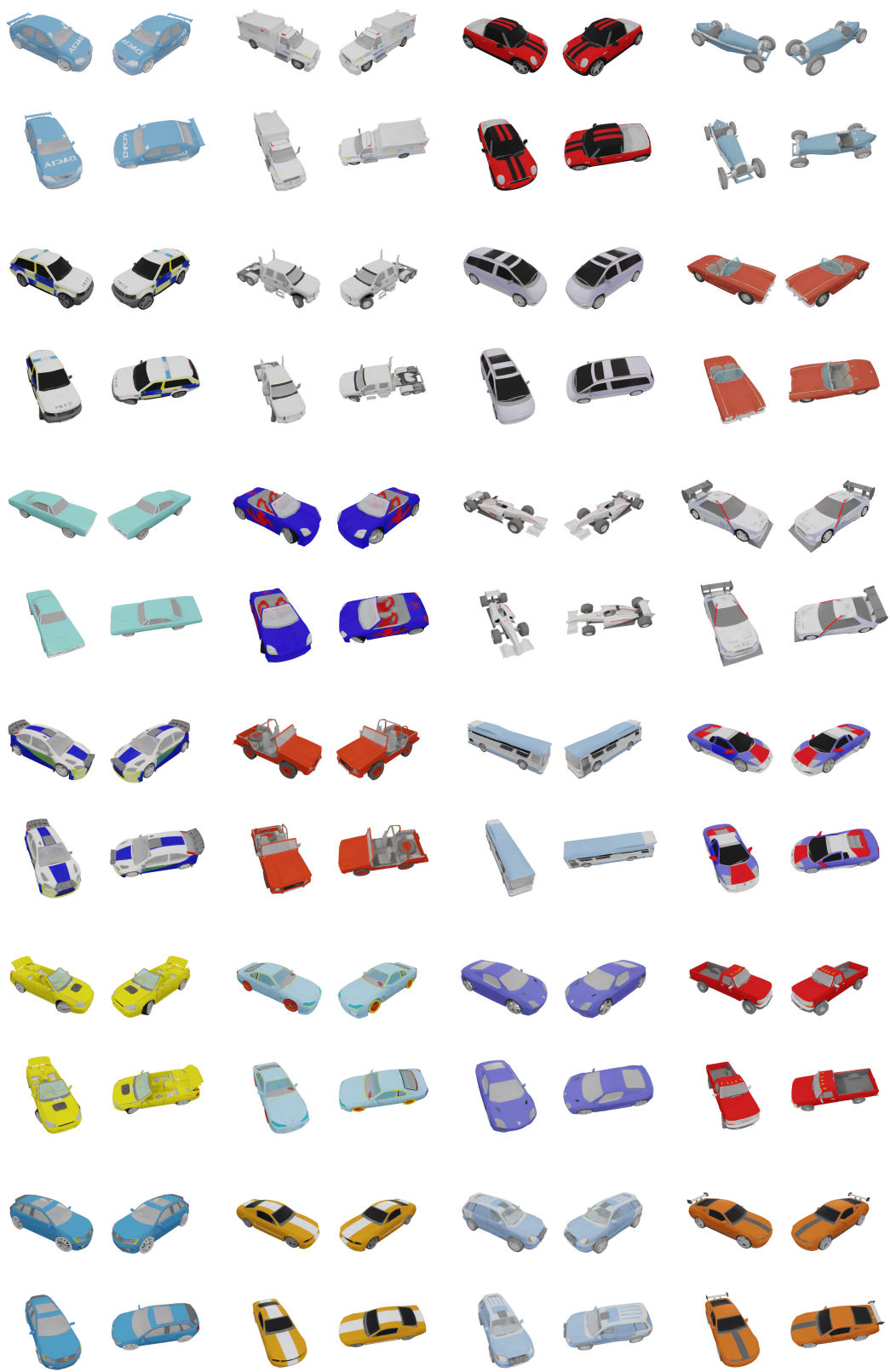}
	\end{tabular}
	\caption{Additional generated samples on ShapeNet Car.}
	\vspace{-3mm}
	\label{fig/supp:car_supp}
\end{figure*}

\begin{figure*}[t]
	\small
	\centering
	\begin{tabular}{c}
		\includegraphics[width=0.997\linewidth]{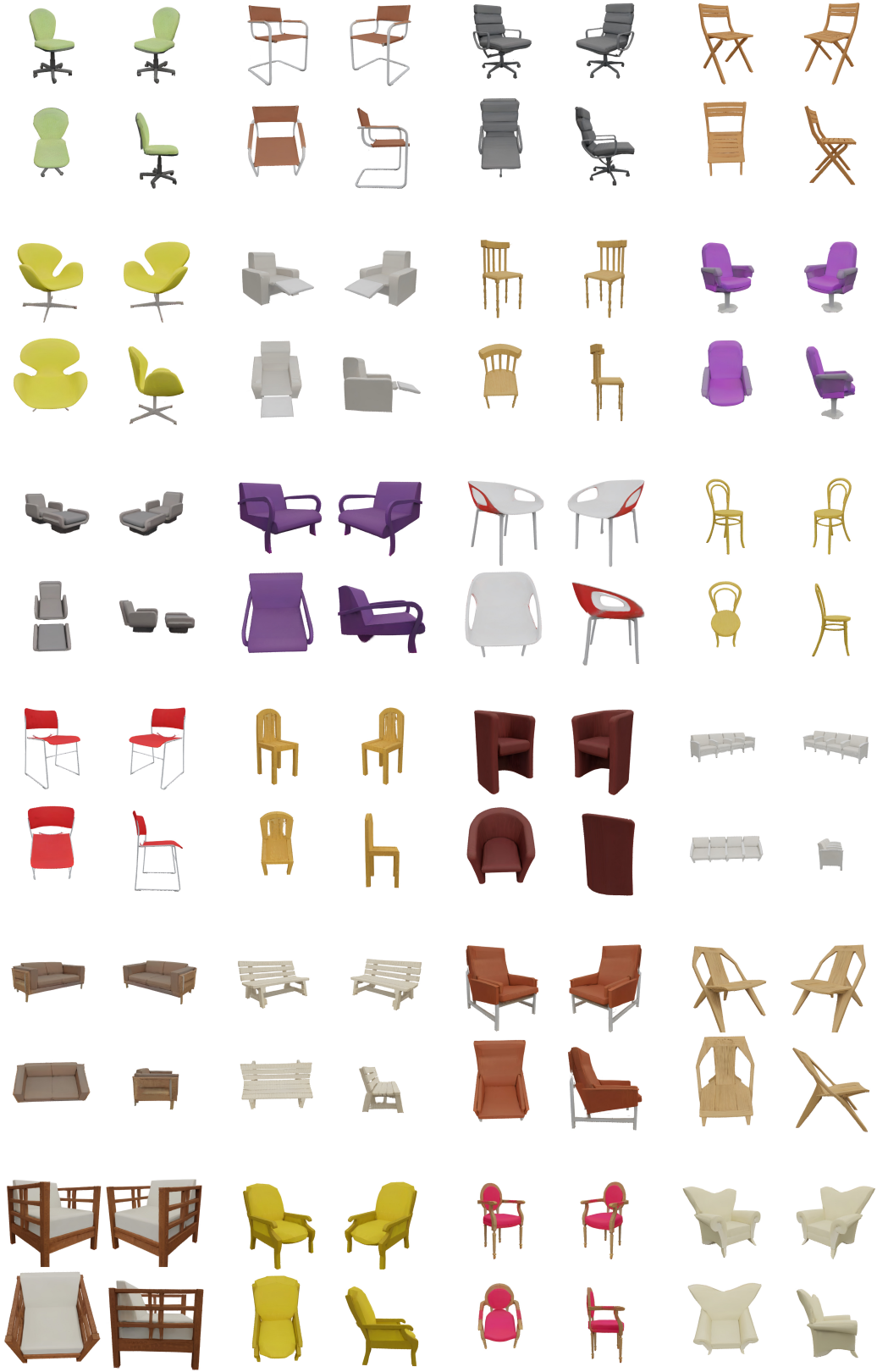}
	\end{tabular}
	\caption{Additional generated samples on ShapeNet Chair.}
	\vspace{-3mm}
	\label{fig/supp:chair_supp}
\end{figure*}

\begin{figure*}[t]
	\small
	\centering
	\begin{tabular}{c}
		\includegraphics[width=0.997\linewidth]{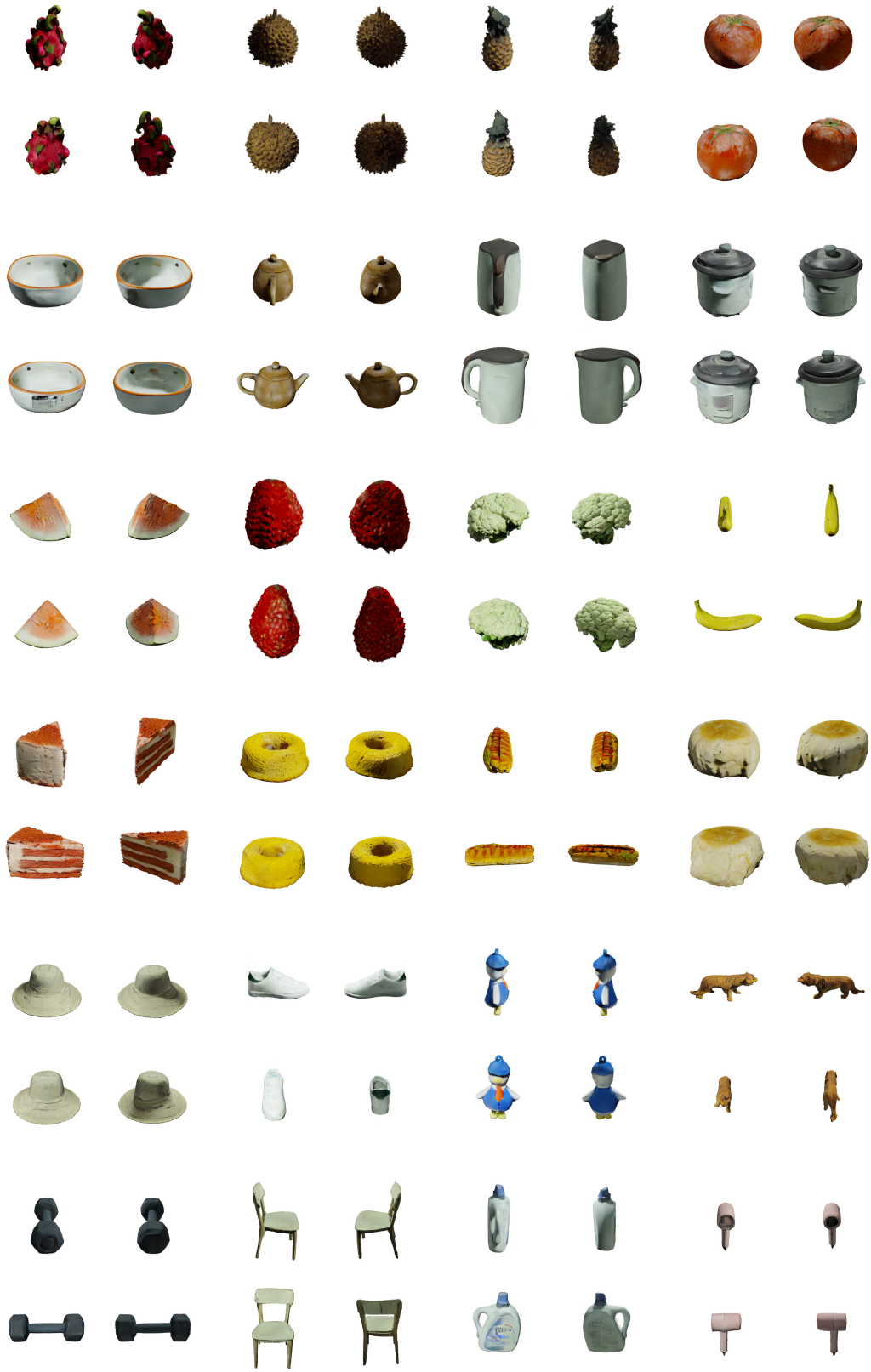}
	\end{tabular}
	\caption{Additional generated samples on OmniObject3D.}
	\vspace{-3mm}
	\label{fig/supp:omni_supp}
\end{figure*}

We provide more generated samples of unconditional and class-conditioned generation in~\Cref{fig/supp:car_supp},~\Cref{fig/supp:chair_supp} and~\Cref{fig/supp:omni_supp}. The additional results demonstrate the strong capability of our model to create high-quality 3D assets with complex geometry and intricate textures.

Furthermore, we also provide an additional video in supplementary material, which intuitively illustrates our approach and visualizes the generated results.

\subsection{Limitations}
\label{supp:limitations}
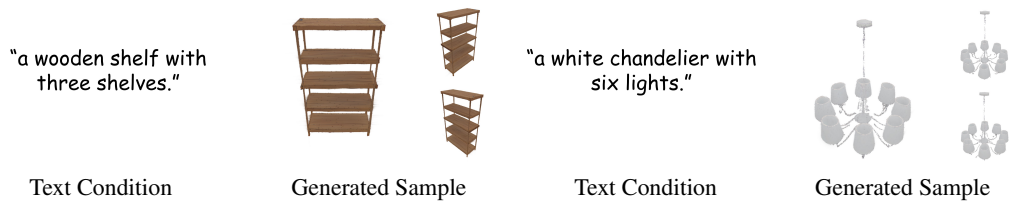
\begin{figure*}[h!]
	\small
	\centering
	\begin{overpic}[width=0.997\linewidth]{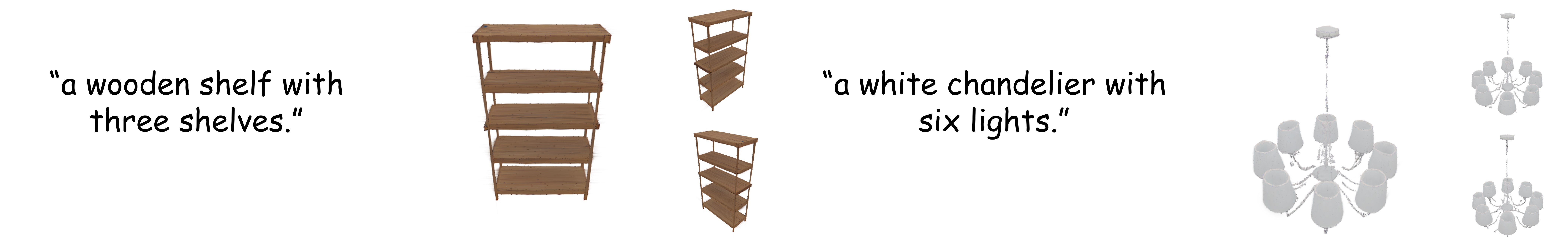}
            \put(5, -3){Text Condition}
            \put(30, -3){Generated Sample}
            \put(57, -3){Text Condition}
    	\put(80, -3){Generated Sample}
	\end{overpic}
	\caption{Failure cases.}
	\vspace{-3mm}
	\label{fig/supp:failure_cases}
\end{figure*}

While GaussianCube represents a substantial step forward in developing an ideal representation for 3D content generation, it still has some limitations. Specifically, although the GaussianCube construction procedure is considerably more rapid than that of NeRF-based methods and can be executed in parallel, it still requires approximately 5 minutes to construct each object. This presents a challenge for scaling up training on extensive 3D datasets. In future work, we plan to investigate more time-efficient methods for GaussianCube construction. Additionally, akin to prior 2D diffusion models, our text-to-3D diffusion model encounters difficulties in presenting the specified number of objects within prompts as shown in~\Cref{fig/supp:failure_cases}. To address this, we will look into enhancing the precision and controllability of 3D generation in the future.

\subsection{Broader Impacts}
\label{supp:broader_impacts}
The proposed GaussianCube enables high-quality 3D asset fitting with few parameters, which significantly simplifies the challenges of 3D generative modeling. Our diffusion model is capable of generating high-quality 3D assets of complex geometry and intricate textures  while also accommodating a variety of conditional signals to steer  the creating procedure. The strong capability of GaussianCube suggests its potential to serve as a versatile 3D representation for a variety of applications in future 3D research endeavors.

Like all generative models, particular caution is required when dealing with sensitive tasks involving human representations. 
Our avatar creation model is trained exclusively on a synthetic dataset~\cite{wood2021fake} composed of large-scale 3D digital avatars which are generated through a graphics pipeline. We conceptualize digital avatars as analogous to those created by specialized 3D artists, rather than photorealistic human images. This strategy in selecting training data mitigates privacy and copyright issues that might arise from utilizing real human photo collections. Nevertheless, it is crucial to acknowledge that avatars generated by our model from real-world imagery could still be misused for spreading disinformation. As such, we advocate implementing rigorous safeguards and promoting responsible use of our technology other related ones to mitigate such risks. 

\begin{figure*}[t]
	\centering
	\scriptsize
	\begin{overpic}[width=0.9\linewidth]{imgs/supp/text_cond_supp_small.jpg}
		\put(2,-2){DreamGaussian~\cite{tang2023dreamgaussian}}
        \put(22, -2){VolumeDiffusion~\cite{tang2023volumediffusion}}
        \put(47, -2){Shap-E~\cite{jun2023shap}}
        \put(67, -2){LGM~\cite{tang2024lgm}}
		\put(88, -2){\textbf{Ours}}
	\end{overpic}
	\vspace{3mm}
	\caption{Additional qualitative comparison of text-to-3D generation on Objaverse~\cite{deitke2023objaverse}. Our model is capable of creating high-quality samples following input text prompts.}
	\vspace{-5mm}
	\label{supp/fig:textcond_gen}
\end{figure*}

\end{document}